\documentclass[opre,nonblindrev]{informs4}

\OneAndAHalfSpacedXI

\let\theoremstyle\relax

\usepackage{amsmath,amssymb,amsthm,mathtools}
\RequirePackage{tgtermes}
\RequirePackage{newtxtext}
\RequirePackage{newtxmath}
\usepackage{natbib}
 \bibpunct[, ]{(}{)}{,}{a}{}{,}%
 \def\bibfont{\small}%
\usepackage{xcolor}
\usepackage{booktabs}
\usepackage{multirow}
\usepackage{array}
\usepackage{longtable}
\newcolumntype{L}[1]{>{\raggedright\arraybackslash}p{#1}}

\usepackage{enumitem}
\usepackage{graphicx}
\usepackage{float}
\usepackage[colorlinks=true,linkcolor=blue,citecolor=blue,hypertexnames=false]{hyperref}

\providecommand{\ECHowTheorems}{}
\providecommand{\ECHowEquations}{}
\providecommand{\ECHowSections}{}
\graphicspath{{}}

\newtheorem{theorem}{Theorem}
\newtheorem{proposition}{Proposition}
\newtheorem{lemma}{Lemma}
\newtheorem{corollary}{Corollary}
\theoremstyle{definition}
\newtheorem{assumption}{Assumption}

\newtheorem{remark}{Remark}

\newcommand{\ECnumbering}{%
  \setcounter{theorem}{0}\renewcommand{\thetheorem}{EC.\arabic{theorem}}%
  \setcounter{proposition}{0}\renewcommand{\theproposition}{EC.\arabic{proposition}}%
  \setcounter{lemma}{0}\renewcommand{\thelemma}{EC.\arabic{lemma}}%
  \setcounter{corollary}{0}\renewcommand{\thecorollary}{EC.\arabic{corollary}}%
  \setcounter{assumption}{0}\renewcommand{\theassumption}{EC.\arabic{assumption}}%
  \setcounter{definition}{0}\renewcommand{\thedefinition}{EC.\arabic{definition}}%
  \setcounter{remark}{0}\renewcommand{\theremark}{EC.\arabic{remark}}%
  \setcounter{example}{0}\renewcommand{\theexample}{EC.\arabic{example}}}

\newcommand{\R}{\mathbb{R}}
\newcommand{\E}{\mathbb{E}}

\newcommand{\Cov}{\operatorname{Cov}}
\newcommand{\op}{\mathrm{op}}
\newcommand{\dto}{\stackrel{d}{\longrightarrow}}
\newcommand{\pto}{\stackrel{p}{\longrightarrow}}
\newcommand{\weakto}{\Rightarrow}

\newcommand{\Diag}{\operatorname{Diag}}
\newcommand{\Sym}{\operatorname{Sym}}
\newcommand{\thetast}{\theta^{\star}}
\newcommand{\Qly}{\mathsf{Q}}
\newcommand{\bth}{\bar{\theta}}
\newcommand{\Pcal}{\mathcal{P}}
\newcommand{\Scal}{\mathcal{S}}
\newcommand{\Acal}{\mathcal{A}}
\newcommand{\Fcal}{\mathcal{F}}
\newcommand{\Bcal}{\mathcal{B}}
\newcommand{\Qcal}{\mathcal{Q}}
\newcommand{\ind}{\mathbf{1}}

\makeatletter
\newcommand{\flushcaption}{\long\def\@makecaption##1##2{\EGT\TableCaptionFontStyle
  \noindent\HD{9}{0}{\TableNameFontStyle ##1\kern16pt}\ignorespaces ##2\HD{0}{5}\endgraf}}
\makeatother
\makeatletter
\def\theARTICLETOPLEFT{}
\def\theARTICLETOPRIGHT{}
\def\theARTICLEABSTRACT{\HOOKb\vspace*{18pt}%
  \begin{minipage}[t]{\textwidth}\parindent1em\ABSfont\noindent\theABSTRACT\endgraf\vskip5pt
  \theFUNDING\theKEYWORDS\theSUBJECTCLASS\theAREAOFREVIEW\theMSCCLASS\theORMSCLASS
  \noindent\hrulefill\end{minipage}\vspace*{0pt}}
\RRHSecondLine{}\LRHSecondLine{}
\makeatother
\hypersetup{pdftitle={Adaptive Optimizers Can Get a Policy Ranking Wrong: Momentum, Dependent Data, and a Two-Run Remedy},pdfauthor={Sunyoung An and Xiaoming Huo}}
\begin{document}

\RUNAUTHOR{An and Huo}

\RUNTITLE{Adaptive Optimizers Can Get a Policy Ranking Wrong}

\TITLE{Adaptive Optimizers Can Get a Policy Ranking Wrong:
Momentum, Dependent Data, and a Two-Run Remedy}

\ARTICLEAUTHORS{%
\AUTHOR{Sunyoung An}
\AFF{H. Milton Stewart School of Industrial and Systems Engineering, Georgia Institute of Technology, \EMAIL{san49@gatech.edu}}

\AUTHOR{Xiaoming Huo}
\AFF{H. Milton Stewart School of Industrial and Systems Engineering, Georgia Institute of Technology, \EMAIL{huo@gatech.edu}}
}

\ABSTRACT{%
Operators choosing among decision rules, such as when a battery charges or where an idle taxi
goes, rank them by estimated long-run value. A standard estimator, linear temporal-difference (TD)
learning, fits values from one stream of correlated observations. Large fits often use an
adaptive optimizer, such as adaptive moment estimation (Adam), with fixed averaging rates. We show that these rates
can change what is estimated, not just how fast: the same observations drive the updates and the
optimizer's step-size scaling, moving the fit away from the intended TD value. The shift
differs across rules, so it need not cancel in a comparison, and more data only makes a
wrong ranking look more certain. Our remedy refits the same data at a second averaging rate that halves the effective gain, the
number to which the leading shift is proportional. Twice the new fit minus the original removes
that shift. In New York taxi and battery models calibrated to real data, common optimizer
settings leave the selected rule unchanged but order close rules wrongly and misstate its
value. Our remedy corrects nearly every wrong ordering, with one extra fit when the shift is
moderate and two when it is larger.
}%

\KEYWORDS{adaptive optimizers; policy ranking; momentum; Markovian stochastic approximation; Richardson--Romberg extrapolation; temporal-difference learning; policy evaluation}

\maketitle

\section{Introduction}
\label{sec:intro}

Operators choose among decision rules by comparing the rules' estimated long-run values. A
battery operator facing five-minute electricity prices picks two price thresholds: charge below
one and discharge above the other. A taxi dispatcher decides what a newly idle cab should do: wait
for a street hail, drive empty to the airport, or return empty to the Manhattan core. Each
dispatch rule assigns these actions differently by location. In both settings, each rule's
estimate comes from observations collected under it.

In reinforcement learning, estimating a fixed rule's value is called policy evaluation. Under
each rule, the system evolves as a Markov chain, the rule's value is its expected discounted
sum of rewards, and the value function is approximated by a weighted combination of features.
Temporal-difference (TD) learning fits the weights incrementally from one trajectory of
correlated observations. Least-squares temporal differences (LSTD) estimates, in one linear
solve, the weights that standard TD converges to.

With hundreds of thousands of features, LSTD's linear system can be impractical to form
and solve, and LSTD does not apply directly to a neural-network value model. The weights are then often fitted by an adaptive optimizer whose averaging rates, the share of each running average carried into the next, are left fixed: root mean square propagation (RMSProp)
\citep{tieleman2012} or adaptive moment estimation (Adam) \citep{kingma2015}. Both divide each
update by the square root of a running average of squared gradients. Adam also averages the
gradients themselves, and that average is the momentum. In the online setting studied here,
each observation supplies one update. The two averaging rates, for the gradient and squared-gradient averages, often stay at library defaults,
as the squared-gradient rate does in \citet{andrychowicz2021}, and are treated as optimizer settings rather than modeling
choices. We call either optimizer,
run this way, the deployed optimizer.

Practitioners keep the adaptive optimizer because it reduces tuning effort. TD learning with a single step size needs that step size tuned to each application, and a poor choice slows convergence or destabilizes the fit
\citep{georgepowell2006,sun2020adatd}. Scaling each coordinate's update by the root mean square of
its recent gradients can reduce sensitivity to the step size.
Reinforcement-learning studies adopt Adam because it is less sensitive to the step size than
RMSProp \citep{hessel2018}, because one step size, 0.0003, performed well on all five tasks of a large study \citep{andrychowicz2021}, or because Adam with a squared-error loss outperformed RMSProp with a Huber loss on Atari games \citep{obandoceron2021}. A ride-hailing platform used Adam to fit the value network it tested in production \citep{tang2019}.

The deployed optimizer's averaging rates can change what the fit estimates, not just how fast it
converges. The same observations drive both the parameter updates and the running average of
squared gradients. With the squared-gradient averaging rate fixed, this average need not settle, and because it moves together with
the observations that drive the updates, it can shift the fit's limit away from the intended LSTD
value. We call this shift the displacement. To first order, it is proportional to the weight the running average gives each new squared gradient, which we call the gain. Adding momentum does not generally remove it.

The displacement undermines policy comparisons in three ways. First, convergence and a correct
standard error can describe a fit centered on the wrong value, because sampling uncertainty does
not measure displacement. Second, different rules generate different data, so their
displacements can differ and need not cancel in a comparison. Third, with the rates
fixed, more observations shrink the standard error but leave the displacement intact, increasing
confidence in a wrong ranking. These statements hold whether the running average used in an update already includes the
current observation, as in the \emph{current} update of RMSProp and Adam, or excludes it, as in the
\emph{lagged} update of some newer optimizers. With a decaying step size, our convergence results hold on the event that the fitted weights stay
near the displaced target (Section~\ref{sec:tddisplacement}).

On two models calibrated to operational data, the deployed optimizer orders close rules wrongly
and misstates the value of the rule it selects, although the selected rule does not change. In a
taxi model fitted to New York trip records, at a squared-gradient averaging rate of 0.95, the rate
used by \citet{mnih2015}, it orders four pairs of repositioning rules wrongly
(Table~\ref{tab:taxirev}). For one of them, two rules \$0.55 apart in value at the core, the
wrong-signed difference without momentum exceeds two LSTD standard errors after about six weekdays of the fleet's daytime trips
per rule. In a battery-storage model calibrated to California market prices, close pairs are
ordered wrongly already at the library defaults, an averaging rate of 0.9 for the momentum and
0.999 for the squared-gradient average (Tables~\ref{tab:caisorev} and~\ref{tab:caisorun}). Both
models are small so that each rule's LSTD value, the reference for a wrong ordering, is exactly
computable. The wrong orderings are not an artifact of crude features: the storage model refitted with one feature
per state, which makes LSTD exact, has more wrong orderings, smaller on average. Our theory puts the divisor's stabilizing constant inside the
square root, and it can require gains small relative to how often each feature is active. With Adam
exactly as libraries implement it, on a taxi model with 264 indicator features, the shift is
larger, a fifth of the selected rule's value. Two runs remove every wrong ordering there but
only about half of the shift.

In the storage model, the displacement matters more for the reported value than for the ranking.
Under the current update without momentum, the rule the optimizer selects is reported about
four percent below its LSTD value at the default squared-gradient averaging rate, however long the price record. Its wrongly ordered pairs are close: for the pair we
report in full, the wrong-signed difference exceeds two single-run standard errors only after about five million
five-minute steps, some 50 years of prices, when both rules are fitted on one price path. Over
an eight-month record like ours, sampling noise dominates that comparison.

\subsection{Our proposal and the theory behind it}
\label{sec:introproposal}

We correct the fit with a second run on the same data, leaving the deployed run unchanged, so the operator keeps the adaptive optimizer, momentum included, and its savings in tuning effort. A single run at a smaller gain would shrink the displacement without showing how much remains, and in the storage model, such a run still orders rules wrongly when the deployed run uses the defaults. The leading
displacement is proportional to one number, the \emph{effective gain}, computed from the two
averaging rates. Keeping the momentum rate fixed, we change the
squared-gradient averaging rate so that the effective gain halves, and we refit on the same trajectory
from the same initialization. Twice the second fit minus the original cancels the leading
displacement at any momentum rate, including Adam's default of 0.9, under the current or the lagged update. This combination is
Richardson--Romberg (RR) extrapolation, a standard device in numerical analysis and simulation:
estimates at two values of a small parameter are combined so that the error term linear in that
parameter cancels. Here, the parameter is the effective gain.

When the second-order displacement matters, a third fit at one quarter of the original effective
gain removes the next term. We add it when the two-run correction changes the gap between the competing rules' values
by more than about eight times the corrected gap, a rule of
thumb calibrated on one storage pair. At large gains, the rule can fail to call for it, and a third fit can also leave more error than
two. The analyst does not need to know where the displacement comes from. Under the current update, the two-run combination's standard error in our numerical studies is at most 1.51 times a single run's.

Our first result gives the value the deployed optimizer converges to (Theorems~\ref{thm:momentum} and~\ref{thm:deployedcons}).
To first order, the displacement is proportional to the effective gain, with a coefficient set by how residuals,
the LSTD fit's feature-weighted one-step prediction errors, covary with squared residuals at the same and other times.

Our second result bounds what the displacement costs the operator (Theorem~\ref{thm:regret}).
Selection regret, the LSTD value lost by the optimizer's choice in the limit, is at most about the effective
gain times the spread of the displacement coefficients across rules. When that spread is positive, no value is lost if the effective gain is below roughly the gap between the two best rules' values divided by the spread.

When the operator tunes a continuous parameter, such as a price threshold, rather than choosing
among a few rules, the displacement
can matter more for the reported value than for the decision. Near a smooth interior optimum with negative curvature, the tuned parameter
moves slightly and the LSTD value lost is second order, but the tuned rule's reported value is off
by the full first-order displacement (Proposition~\ref{prop:knob}).

Our third result justifies the remedy. Two runs remove the displacement's first-order term and
three runs its second-order term (Theorem~\ref{thm:expansion} and Corollary~\ref{cor:rr}), so
the selection regret after correction has a bound of second order in the effective gain
rather than first order (Corollary~\ref{cor:rrregret}). Because the runs share one
trajectory, a confidence interval for the combination's limit can be computed from the runs alone
(Theorem~\ref{thm:rr}). The interval that covers reliably is wide, about four times as wide as a
narrower one that undercovers at our run lengths. At four times our standard run length, a typical interval excludes zero for the taxi pair above, but
for the storage pair, it still contains zero at sixteen times that length.

The mechanism reaches beyond TD learning, but our proofs, which rely on the structure of linear TD, do not. The same shift can arise
whenever a model is fitted online to correlated data with updates rescaled by statistics of
those data, and where it is first order in the gain, a second run can cancel its leading term.

\subsection{Relation to prior work}
\label{sec:intropriorwork}

The displacement that exists only under dependent data and its leading-order removal by extrapolation in the effective gain
are new, but they build on two recent lines of work, one on Adam with independent data and one
on fixed-step methods with Markov data. Neither combines dependent data with an adaptive
optimizer.

Recent work locates Adam's limit under independent sampling, and the part of our displacement
that remains under independent sampling is already known.
\citet{dereich2024,dereich2025ode,dereich2025sym,dereich2026} study Adam with momentum, fixed
averaging rates and a diminishing step size under independent sampling. They show that it
can converge to a point that does not minimize the loss. We claim no priority
for that part.

Recent work also uses extrapolation to reduce the bias of fixed-step stochastic approximation on Markov
data, but that bias has a different source. \citet{huo2026} study linear stochastic
approximation whose updates are not rescaled by the data, at a fixed step size on Markov data. They link its bias to mixing time for reversible chains and reduce it by RR extrapolation in the step size. \citet{zeng2026} develop inference for such RR combinations. Their bias comes from
the fixed step size and disappears as the step shrinks. Ours comes from the fixed gain of the
squared-gradient average used to rescale each update, and it remains even as the step shrinks.

The lagged update, which three optimizers build on, need not remove the displacement without momentum, and at Adam's default momentum, it can keep most of it (Section~\ref{sec:tddisp}). AdaShift \citep{zhou2019}, ACProp \citep{zhuang2021} and ADOPT
\citep{taniguchi2024} keep the current observation out of the running average in different
ways and make other changes, so our results concern the lag alone. Without momentum, it leaves the displacement carried by earlier
observations, which vanishes when the features represent the value exactly. With
momentum, it also keeps, to first order, a share of the current observation's contribution equal to the momentum rate, and at Adam's default
of 0.9, it still orders taxi rules wrongly.

Letting the gain shrink over time on a suitable schedule can remove the displacement, but it changes the deployed configuration, whereas our remedy keeps the deployed run. AdamNC \citep{reddi2018}
and Adafactor \citep{shazeer2018} let the gain decay. With independent noise, related
stochastic-approximation schemes recover the minimizer in stochastic optimization
\citep{anhuo2025a,anhuo2026b}. The corresponding result for TD learning is not proved here.

Comparing policies using values estimated from temporally dependent data also
arises in mobile health \citep{luckett2020} and in the estimation of dynamic treatment regimes \citep{zhang2018}.

Adaptive-gradient methods with a fixed gain are widely used in
reinforcement learning. These methods developed from AdaGrad \citep{duchi2011} to RMSProp
\citep{tieleman2012} and Adam \citep{kingma2015}. \citet{mnih2015} used a squared-gradient averaging rate of
0.95 for action-value fitting, a value \citet{castro2018} retained. Other studies keep that rate at
the library default \citep{andrychowicz2021} or lower it to 0.9 to preserve networks'
ability to adapt to changing targets \citep{lyle2023}. The evidence for keeping adaptive
optimizers is empirical. In offline action-value fitting on 60 Atari games, replacing RMSProp with Adam
raised the median normalized score from 83 to 112 percent \citep{agarwal2020}. In an eight-problem
deep-learning benchmark, Adam at its default settings is competitive with tuned alternatives on several problems
\citep{schmidt2021}. The advantage is not universal: adaptive optimizers remain sensitive to the step
size \citep{hendersonoptim2018}, and a well-tuned TD update with one global step size matches them in most settings
tested \citep{romoff2021}.
For reinforcement learning,
\citet[Thms.~3--4]{xiong2021} establish rates for Adam-type methods. Their projected AMSGrad
also uses a fixed gain, but adds a running maximum. Under the conditions of their diminishing-step theorem,
its output converges to the weights LSTD computes, so a fixed gain alone need not displace the target.

That reweighting the state-sampling distribution can change the TD fixed point is well established in the
off-policy literature \citep{tsitsiklis1997,kolter2011}. Here, the optimizer's divisor
reweights the updates instead, at a gain chosen
to speed convergence rather than to control the target. Because the update weights depend on the residuals, the target can move even when the features represent the value exactly.

Prior work studies configuration-sensitive rankings and overestimation caused by tuning. Algorithm rankings vary across seeds, implementations and
hyperparameters \citep{henderson2018,patterson2024}. Tuning against an unbiased estimator can
select settings whose performance is overstated \citep{saito2024}. Here, the target can move at fixed optimizer settings, with no hyperparameter search.

Ranking and selection, like inference for convergent estimators, assumes each estimate converges
to the value of interest, which fails here. In simulation, under regularity conditions, the
probability of ordering two systems wrongly decays exponentially in the run length, far faster
than either standard error \citep{dai1996,chang2004}. Ordinal comparison exploits this. When the
displacements invert the limiting ranking, consistency instead yields the wrong order
asymptotically, which no valid standard error detects. Inference for stochastic-gradient estimates
and for policy evaluation \citep{chen2020,ramprasad2023,luckett2020} and the self-normalized
methods \citep{shao2010,kiefer2000,lee2022,glynniglehart1990} likewise target the limit of a
convergent estimator, which here need not be the LSTD value.

Our new contributions are the displacement that exists only under dependent data, the effective gain as the common first-order scale of every part of the displacement, and the displacement's consequences for decisions: a regret bound, wrong rankings and first-order error in the reported value. With momentum, extrapolating in the effective gain rather than the gain alone also avoids an extra second-order error term.

The remedy extends established tools rather than introducing a new one. Pairing fits on
one trajectory is the common-random-numbers design of simulation \citep{glasserman1992}. We use RR
extrapolation \citep{talay1990,dieuleveut2020} as adapted to Markov data by
\citet{huo2026}. Our extension extrapolates in the effective gain rather than in the step size or the gain. It shows that the remedy remains valid with momentum, and it provides
the combination's limit law and confidence interval on one trajectory.

\subsection{Organization}
\label{sec:introorg}

Section~\ref{sec:summary} specifies the two applications, the estimating equation and the
deployed optimizer. Section~\ref{sec:remedies} describes the remedy, and
Section~\ref{sec:tddisplacement} proves the results supporting its steps.
Section~\ref{sec:numerics} quantifies displacement and its effect on rankings in the calibrated
storage and taxi models: close pairs are ordered wrongly, but the selected rule does not change.
Section~\ref{sec:discussion} discusses practical implications and limitations.

\section{The problem, the model, and the deployed optimizer}
\label{sec:summary}

This section sets up the problem, the estimating equation and the deployed optimizer. Proofs
omitted from the text appear in the electronic companion, Sections EC.1 to EC.3.

\emph{Notation.}\label{sec:notation} The spectral norm of a matrix $M$ is $\|M\|_{\op}$. For
square $M$, its symmetric part is $\Sym(M)=\tfrac12(M+M^\top)$, and $M\succ0$ means that $M$ is
symmetric positive definite. The identity matrix is $I$, and $\Bcal(q,r)$ is the closed ball of radius $r$ centered at $q$. For signed measures, $\|\cdot\|_{\mathrm{TV}}$ denotes the
total-variation norm. The Hadamard product is $\odot$, $\Diag(q)$ is the diagonal matrix with diagonal $q$, $\Diag(M_1,M_2)$ is block diagonal with blocks $M_1$ and $M_2$, and $\ind$ is the vector of ones. The arrows $\weakto$, $\dto$ and $\pto$ denote
weak convergence in the specified Skorokhod space, convergence in distribution and convergence
in probability. For a standard Brownian motion $W$, $\bar W(\xi)=W(\xi)-\xi W(1)$ is its bridge on
$[0,1]$. Dimensions follow from context.

For the optimizer, $\beta_2$ is the averaging rate of the squared-gradient average,
$\rho=1-\beta_2$ its gain, and $\lambda=\beta_1$ the momentum rate. We set $0^0:=1$.

\subsection{The operator's problem and the estimating equation}
\label{sec:problem}
\label{sec:tdsetup}

In both of our applications, an operator selects
from a small set of rules the one with the highest estimated long-run value. We ask whether the
fitted values rank the rules correctly and how to correct the ranking when they do not.

In merchant storage arbitrage, the rules are double-threshold rules: a battery charges
at low electricity prices and discharges at high ones. Prior studies estimate the arbitrage value of
storage \citep{sioshansi2009,zhou2016} and optimize its trading and dispatch
\citep{lohndorf2013,jiangpowell2015}. For a capacitated asset with Markovian prices,
\citet{secomandi2010} shows that optimal trading follows two price-dependent basestock targets
(target inventory levels), which motivates double-threshold rules.

In idle-vehicle repositioning, the rules decide whether a newly idle cab waits
locally for a street hail or drives empty to an area with better earning opportunities. Empty-car routing models
address this choice across a ridesharing fleet \citep{braverman2019}.

\emph{The operator's problem.} The operator chooses between two decision rules, or
\emph{policies}, $\pi_1$ and $\pi_2$. The system is a Markov decision process with finite state
and action sets $\Scal$ and $\Acal$, of sizes $n_S$ and $n_A$, and discount factor
$\gamma\in(0,1)$. At time $t=0,1,\dots$, its state is $x_t\in\Scal$ and its one-step reward
is $r_t$. A policy $\pi$ maps each state to an action distribution. Its state kernel
$\Pcal^\pi$ on $\Scal$ has stationary law $\mu$, and its discounted values are
\begin{equation}
\label{eq:estimand}
  V^\pi(x)=\E^\pi\Big[\sum_{k\ge0}\gamma^k r_{t+k}\,\Big|\,x_t=x\Big],
  \qquad
  \mathcal V^\pi(\nu_0):=\E_{x\sim\nu_0}\!\big[V^\pi(x)\big],
\end{equation}
where $\E^\pi$ denotes expectation under the law induced by $\pi$ and $\nu_0$ is a reference
law over starting states. The correct choice is $\pi_1$ when
$\mathcal V^{\pi_1}(\nu_0)>\mathcal V^{\pi_2}(\nu_0)$ and $\pi_2$ otherwise, and the operator
decides from the sign of the estimated difference of the two values. The analysis extends to any finite set of rules
compared pairwise in this way, with the LSTD values defined below as the reference (Theorem~\ref{thm:regret}).

\emph{Policy evaluation supplies the estimates.} The values are estimated when the
kernel $\Pcal^\pi$ is unknown or too large to compute with. Each fit uses one stationary
trajectory of transition tuples $\zeta_t=(x_t,a_t,r_t,x_{t+1})$, with action $a_t\sim\pi(\cdot\mid x_t)$, generated by running or
simulating its policy. With $d$ features $\varphi:\Scal\to\R^d$ and a parameter
$\theta\in\R^d$, the value function is approximated by $V_\theta(x)=\varphi(x)^\top\theta$, and
the reported value is $\ell^\top\theta$ with $\ell:=\E_{\nu_0}[\varphi(x)]$. With
$\theta^{(i)}$ fitted to policy $\pi_i$'s trajectory, the decision uses the sign of
$\ell^\top(\theta^{(1)}-\theta^{(2)})$.

\emph{TD learning fits the parameter.} The TD estimating equation
\citep{sutton1988,tsitsiklis1997} makes the one-step Bellman residual orthogonal to the features
in expectation. In descent form, its estimating function is
$g(\theta,\zeta_t):=-\varphi(x_t)\{r_t+\gamma\varphi(x_{t+1})^{\top}\theta-\varphi(x_t)^{\top}\theta\}
=A(\zeta_t)\theta-b(\zeta_t)$, where
$A(\zeta_t):=\varphi(x_t)\{\varphi(x_t)-\gamma\varphi(x_{t+1})\}^{\top}$ and
$b(\zeta_t):=\varphi(x_t)r_t$. Its population root, the \emph{LSTD target}, is
$\thetast=A^{-1}b$, where
\begin{equation}
\label{eq:Adef}
  A=\E_\mu\!\left[\varphi(x)\{\varphi(x)-\gamma\varphi(x')\}^{\top}\right],
  \qquad b=\E_\mu[\varphi(x)r],
  \qquad (x,a,r,x')\sim\mu_\zeta ,
\end{equation}
and $\mu_\zeta$ is the stationary tuple law induced by $\mu$. Here, $\E_\mu$ denotes expectation under
either law, as the integrand indicates.

\emph{What the fit is compared with.} The reference is LSTD \citep{bradtke1996,boyan2002},
which solves the sample version of $A\theta=b$ using $O(d^{2})$ memory and an $O(d^{3})$ dense factorization, and
converges to $\thetast$ with no optimizer and hence no displacement. Each rule's
\emph{LSTD value}, the value the analyst intends to estimate, is therefore $\ell^\top\thetast$. Its gap to the exact value
\eqref{eq:estimand} is the linear approximation error, which vanishes when the features
represent $V^\pi$ exactly, does not depend on the optimizer, and is reported separately for each model. The storage and taxi models are kept small so that LSTD is computable. The same
displacement mechanism operates in larger models, although its first-order description can require
the gain to be small relative to the smallest state-visit or feature-activation frequency
(Section~\ref{app:zoneperiod}).

\subsection{The deployed optimizer}
\label{sec:deployed}

Without momentum, RMSProp and Adam fit the parameter online, one transition per update, by
\begin{equation}
\label{eq:update}
  \theta_{t+1}=\theta_t-\eta_tP_tg(\theta_t,\zeta_t),
  \qquad
  \bth_n=\frac1n\sum_{t=1}^n\theta_t,
\end{equation}
where the \emph{preconditioner} $P_t$, the matrix that rescales each update, is symmetric and
positive definite and $\eta_t=\eta_0t^{-\alpha}$, with fixed $\eta_0>0$
and $\alpha\in(1/2,1)$. The average $\bth_n$ of the first $n$ iterates is the Polyak--Ruppert average \citep{polyak1992,ruppert1988}. The numerical studies average over the second half of each run. The data and initial state generate the natural filtration
$\Fcal_t:=\sigma(\theta_1,v_{\min(0,1-m)},\dots,v_0,u_0,\zeta_1,\dots,\zeta_t)$, with $m$ the delay defined in \eqref{eq:momentum},
where $v_{\min(0,1-m)},\dots,v_0\succeq0$ are initial values of the running average of squared
gradients and $u_0$ is an initial value of the running average of gradients (the momentum).
That squared-gradient average, the \emph{accumulator} $v_t$, determines the lagged and
current preconditioners:
\begin{equation}
\label{eq:sarmsprop}
  v_t=(1-\rho_t)v_{t-1}+\rho_tg(\theta_t,\zeta_t)^{\odot2},
  \qquad
  P^{\mathrm{lag}}_t=\phi(v_{t-1}),\quad P^{\mathrm{cur}}_t=\phi(v_t),
\end{equation}
with $\phi(v):=\Diag(v+\epsilon\ind)^{-1/2}$ and $\rho_t\in[0,1]$. RMSProp and Adam use the fixed
gain $\rho_t\equiv\rho=1-\beta_2\in(0,1)$.

Our damping $\epsilon>0$ is inside the square root, as in TensorFlow's RMSProp
\citep{tfrmsprop}. Adam's denominator $\sqrt v+\epsilon$ \citep{kingma2015} is not covered by our
theory because its reciprocal has an unbounded derivative near $v=0$. The storage and taxi
models use $\epsilon=0.25$, with coordinatewise residual second moments ranging from $0.009$ to $4.6$.

The update's divisor can use the accumulator before or after it includes the current
observation. Deployed RMSProp and Adam use
$P^{\mathrm{cur}}_t$ \citep{tieleman2012,kingma2015}. The lagged choice
$P^{\mathrm{lag}}_t$ is $\Fcal_{t-1}$-measurable. The parameter and accumulator paths
of the two updates can differ after the first update, even on the same trajectory from the
same $(\theta_1,v_0)$.

Adam updates the parameter using a first-moment average of the gradients, with its own
averaging rate $\lambda=\beta_1\in[0,1)$:
\begin{equation}
\label{eq:momentum}
  u_t=\lambda u_{t-1}+(1-\lambda)\,g(\theta_t,\zeta_t),
  \qquad
  \theta_{t+1}=\theta_t-\eta_t\phi(v_{t-m})\,u_t ,
\end{equation}
where the integer \emph{delay} $m\ge0$ is the number of steps by which the divisor's accumulator trails
the current observation, so $m=0$ gives the current update and $m=1$ the lagged one. This recursion is used throughout. At $\lambda=0$, it
reduces to \eqref{eq:update} with $P_t=\phi(v_{t-m})$. Adam's library defaults are the current update, momentum rate $0.9$ and $\beta_2=0.999$, so $(m,\lambda)=(0,0.9)$. Our recursion changes the placement of $\epsilon$ and omits Adam's
bias corrections, whose transient effects do not enter the limits below (Remark~\ref{rem:biascorr}).

``Deployed'' refers to the optimizer's configuration, its two averaging rates
$(\beta_1,\beta_2)$ and whether its update is current or lagged, not its step sizes $\eta_t$. A smaller gain $\rho=1-\beta_2$ means a larger averaging rate $\beta_2$. We use a decaying parameter step size $\eta_t$ with Polyak--Ruppert averaging
so that the gain is the only source of displacement.

Constant step sizes, common in practice, can add a separate bias. When updates are not rescaled by the data, as in the
linear stochastic approximation on Markov data of \citet{huo2026}, this bias is of the order of
the step. Our remedy does
not address it, but in our runs at two constant step sizes, this bias caused no wrong ordering of the pair we report in full for each model, and the two-run combination still removed every wrong ordering of these pairs
(Section~\ref{app:tier4}).

\section{The remedy: a second run at a second averaging rate}
\label{sec:remedies}
\label{sec:recipe}

The remedy keeps the deployed configuration and measures what that configuration costs. Lowering
the gain shrinks the first-order displacement but does not reveal how much remains, and no gain is small
enough for every problem (Remark~\ref{rem:n1se}). Twice the difference between the deployed run and a
second run at half the effective gain estimates the first-order displacement, and their combination
removes that term without the analyst knowing its source. The steps are stated in terms of the hyperparameters
practitioners set.

The remedy is organized around a single scalar, the \emph{effective gain},
\begin{equation}
\label{eq:seff}
  s=s(\rho,\lambda):=\frac{\rho(1-\lambda)}{1-\lambda(1-\rho)}=\frac{(1-\beta_1)(1-\beta_2)}{1-\beta_1\beta_2},
\end{equation}
to which every contribution to the first-order displacement is proportional, under either
update (Theorem~\ref{thm:momentum}). Without momentum, $s=\rho$. With it, $s$ is not linear in
$\rho$, markedly so when $\beta_1$ is near one: $s(2\rho)/s(\rho)\approx1.85$ at $\beta_1=0.9$, $\rho=0.01$.

The remedy has four steps, none of which changes the deployed run.

\noindent\fbox{\begin{minipage}{\dimexpr\textwidth-2\fboxsep-2\fboxrule\relax}
\textbf{The remedy.} \emph{Input:} each rule's trajectory, the deployed rates
$(\beta_1,\beta_2)$ and the contrast $\ell$ fixed in advance. \emph{Output:} a corrected
contrast, its interval and a decision.
\begin{enumerate}[label=(\arabic*),leftmargin=2.2em,itemsep=2pt]
\item Keep the deployed $\beta_1$ and $\beta_2$ and compute the effective gain
  $s=(1-\beta_1)(1-\beta_2)/(1-\beta_1\beta_2)$.
\item Run the same optimizer once more on the same trajectory, from the same initialization,
  at the averaging rate that halves the effective gain,
  \begin{equation}
  \label{eq:beta2prime}
    \beta_2'=1-\frac{\tfrac12 s(1-\beta_1)}{(1-\beta_1)-\tfrac12 s\beta_1} ,
  \end{equation}
  the exact inverse of \eqref{eq:seff}, well defined because $s<1-\beta_1$.
  At Adam's defaults $(\beta_1,\beta_2)=(0.9,0.999)$, this is $\beta_2'\approx0.9995$, and at
  $(0.9,0.99)$, $\beta_2'\approx0.9952$ rather than the $0.995$ that halving $1-\beta_2$ would give.
\item For a contrast $\ell$ fixed in advance, report $\ell^\top\{2\bth_n(\beta_2')-\bth_n(\beta_2)\}$
  with an interval computed from the combined iterates $Y_t=\ell^\top\{2\theta_t(\beta_2')-\theta_t(\beta_2)\}$
  alone. Split the iterates entering the average, in time order, into a first and a second half, with means $\bar Y_{\mathrm I}$ and $\bar Y_{\mathrm{II}}$, and report $\pm\,12.706\,|\bar Y_{\mathrm I}-\bar Y_{\mathrm{II}}|/2$, the two-block batch-means interval.
  The constant is the $0.975$ quantile of Student's $t$ with one degree of freedom. A narrower
  self-normalized interval, scaled by the fluctuation of the partial sums of $Y_t$, is also
  asymptotically valid but undercovers at the run length of our numerical studies. At a fixed gain, either
  interval is for the corrected target, the same two-run combination of the two fits' limits,
  which need not be the LSTD value. To compare rules~1 and~2 on independent trajectories, fit each rule at both rates, four runs in all, and use $Y_t=\ell^\top\{2\theta^{(1)}_t(\beta_2')-\theta^{(1)}_t(\beta_2)\}-\ell^\top\{2\theta^{(2)}_t(\beta_2')-\theta^{(2)}_t(\beta_2)\}$. Prefer the rule the corrected contrast favors. When the interval contains zero, the runs do not settle the comparison.
\item If the two-run correction changes the gap between the rules being ranked by more than about eight times the size of the corrected gap (a rule of thumb, calibrated on one storage pair, that can fail to trigger at large gains, where three runs can also leave more error than two), add a third run at the rate $\beta_2''$ at which the effective gain is $s/4$ and repeat step (3) with
    $\ell^\top\{\tfrac13\bth_n(\beta_2)-2\bth_n(\beta_2')+\tfrac83\bth_n(\beta_2'')\}$.
\end{enumerate}
\end{minipage}}
\medskip

Let $\bth_n(s)$ be the averaged iterate at effective gain $s$, so $\bth_n(\beta_2)$ above is $\bth_n(s)$ and $\bth_n(\beta_2')$ is $\bth_n(s/2)$. More generally, let $\omega\ne1$ be the second run's effective gain divided by the deployed run's. The two-run estimator is
\begin{equation}
\label{eq:rrhat}
  \hat\theta^{\mathrm{RR}}_n:=\frac{\omega\,\bth_n(s)-\bth_n(\omega s)}{\omega-1},
  \qquad\text{so that }\hat\theta^{\mathrm{RR}}_n=2\bth_n(s/2)-\bth_n(s)\text{ at }\omega=\tfrac12 ,
\end{equation}
and step (4) is the same extrapolation on three nodes, the effective gains $s$, $s/2$ and $s/4$
at which the fits are run.

Proved results support every step except the rule of thumb in step (4) and the interval it selects. Step~(2)'s rate halves the leading term
of the displacement, so for every $\beta_1$ and either update, step (3)'s weights eliminate
that term without identifying where the displacement comes from (Corollary~\ref{cor:rr}). The remaining second-order
term has an explicit coefficient, and the three nodes in step (4) cancel it (Theorem~\ref{thm:expansion} and Corollary~\ref{cor:rr}). The interval of step (3) is asymptotically valid because the two fits have a joint functional limit on their shared
trajectory (Theorem~\ref{thm:rr}). Selection by the
corrected targets has regret $O(s^2)$, and when the rules' displacement coefficients differ at both orders, the effective gain below which selection is guaranteed to lose no value scales with the square root of the value gap rather than the gap (Corollary~\ref{cor:rrregret}).

\section{What the theory establishes}
\label{sec:tddisplacement}

This section establishes what the deployed optimizer
estimates and what two or three runs recover, following the order of the remedy. The deployed optimizer's target can be displaced at
first order in the effective gain \eqref{eq:seff}, through separable channels. The iterates
converge to the displaced target, and the displacement can cause decision regret. Two runs cancel
the first-order term, and three runs also cancel the second-order term. For the two-run combination, we establish an interval and a bound on
the remaining regret.

\subsection{Conditions}
\label{sec:assumptions}

Unless otherwise stated, the results in this section assume the data conditions below. The limit theory also
assumes that the iterates stay near the displaced target with positive probability
(Assumption~\ref{a:deployedstable}). No identification result (a result that locates the target rather than describing the iterates) uses this assumption.

\begin{assumption}[Conditions on the data]
\label{a:data}
\begin{enumerate}[label=(\alph*),ref=\theassumption(\alph*),leftmargin=2.4em,itemsep=4pt]
\item\label{a:ergodic}\emph{$V$-uniform geometric ergodicity.} In the sense of
  \citet[Ch.~16]{meyntweedie2009}, there are a measurable $V:\Scal\to[1,\infty)$ and constants
  $C_{\mathrm{mix}}<\infty$, $\kappa\in(0,1)$ with
  $\|(\Pcal^{\pi})^k(x,\cdot)-\mu\|_V\le C_{\mathrm{mix}}V(x)\kappa^k$ for all $x$ and $k\ge0$,
  and $V$ is bounded,
  \begin{equation}
  \label{eq:Vbounded}
    V_{\infty}:=\sup_{x}V(x)<\infty ,
  \end{equation}
  which is automatic when $\Scal$ is finite. Here, $\|\nu\|_V:=\sup_{|f|\le V}|\int f\,d\nu|$ for a signed measure $\nu$.
\item\label{a:bounded}\emph{Bounded features and rewards.} $\|\varphi(x)\|\le C_\varphi$ for all
  $x$, and $|r_t|\le C_r$ almost surely.
\item\label{a:nondegen}\emph{Non-degenerate features.}
  $\Sigma_\varphi:=\E_\mu[\varphi(x)\varphi(x)^\top]\succeq\lambda_\varphi I$,
  $\lambda_\varphi>0$.
\end{enumerate}
\end{assumption}

All three are standard assumptions in the finite-time linear-TD literature
(\citealt[Assumptions~1--3 and Cor.~11]{srikant2019}; \citealt{bhandari2018}, whose
feature-regularity condition, ${\lambda_{\min}(\Sigma_\varphi)>0}$, is Assumption~\ref{a:nondegen}). Assumptions~\ref{a:ergodic}
and~\ref{a:nondegen} hold automatically on a finite state space with an irreducible aperiodic chain and
linearly independent features \citep[Assumptions~\mbox{1--2}]{tsitsiklis1997}. Assumption~\ref{a:bounded} adds
only the reward bound. Since $1\le V\le V_\infty$, \eqref{eq:Vbounded} reduces
Assumption~\ref{a:ergodic} to \emph{uniform} (Doeblin) ergodicity, the form the first-order
expansion uses (Theorem~\ref{thm:momentum}). The $V$-norm formulation is kept for the Poisson bounds
(Section~\ref{sec:setup}).

Stationarity and Assumption~\ref{a:nondegen} give the one-sided TD bound
$\theta^\top A\theta\ge(1-\gamma)\theta^\top\Sigma_\varphi\theta\ge\lambda_A\|\theta\|^2$ with
$\lambda_A:=(1-\gamma)\lambda_\varphi>0$, so $A+A^\top\succeq2\lambda_AI$ and $A$ is invertible
and positive stable (all eigenvalues have positive real parts). This is Lemma~3 of \citet{bhandari2018}, after
\citet[Lemma~9]{tsitsiklis1997}. It plays the role of strong convexity for the nonsymmetric TD estimating equation,
and $\lambda_A\to0$ as $\gamma\to1$.

\begin{remark}[No almost-sure bound on the iterates]
\label{rem:nobound}
No result assumes that the deployed optimizer's iterates are bounded almost surely, a standard stochastic-approximation
condition \citep[Ch.~5]{kushner2003}, which Adam's own analysis imposes in deterministic form
\citep[Thm.~4.1]{kingma2015}. The identification results place no condition on the iterates, and the
limit theory holds on an event of positive probability on which the iterates stay near the
displaced target (Assumption~\ref{a:deployedstable}).
\end{remark}

\subsection{The first-order displacement and its channels}
\label{sec:tddisp}

The deployed optimizer converges to a zero of its \emph{mean field}, its average update
direction when the parameter is held fixed and the accumulator, momentum, and observation follow
their stationary joint law. Because the residual is centered at the LSTD target
($\E_\mu[g(\thetast,\zeta)]=0$), the mean field at $\thetast$ is the covariance $\E[(P-\E P)\,u]$
between the preconditioner $P$ and the momentum $u$ it multiplies ($u=g(\thetast,\zeta)$ without
momentum), under that law. At any gain, the LSTD target
is therefore a zero of the mean field if and only if this covariance vanishes. Momentum does not generally eliminate the displacement. At first order, it reweights the residuals'
cross-cumulants across lags with a kernel fixed by the two rates and the delay.

Write $\varepsilon(\zeta):=g(\thetast,\zeta)=A(\zeta)\thetast-b(\zeta)$ for the residual at the
LSTD target, $\varepsilon_t:=\varepsilon(\zeta_t)$, $F(\zeta):=\varepsilon(\zeta)^{\odot2}$, and
$\bar v:=\E_\mu[F(\zeta)]$, and let $\Qcal$ be the transition operator of the tuple chain, the Markov chain
$(\zeta_t)$ of transition tuples. Let
$D\phi(v)[a]:=-\tfrac12\Diag\big((v+\epsilon\ind)^{-3/2}\odot a\big)$ be the derivative of the
preconditioner map $\phi$ of \eqref{eq:sarmsprop}, and set $J_0:=\phi(\bar v)A$. Fix a radius
$R>\|\thetast\|$. By Assumption~\ref{a:bounded}, $g$ is Lipschitz in $\theta$ with constant
$L_g:=C_\varphi^2(1+\gamma)$, and $\|g(\theta,\zeta)\|\le G:=L_gR+C_\varphi C_r$ whenever $\|\theta\|\le R$, as in
\citet[Lemma~6]{bhandari2018}.

The one object needed is the residual's scaled third-order cross-cumulant function: for
$\tau\in\mathbb Z$,
\begin{equation}
\label{eq:cumulant}
  c(\tau):=\E_\mu\big[D\phi(\bar v)[F(\zeta_0)-\bar v]\,\varepsilon(\zeta_\tau)\big],
  \qquad
  c(\tau)_j=-\tfrac12(\bar v_j+\epsilon)^{-3/2}\,\E_\mu\big[\varepsilon_j(\zeta_0)^2\,\varepsilon_j(\zeta_\tau)\big].
\end{equation}
A negative $\tau$ means that the squared residual comes \emph{after} the residual. Because both
factors are centered and bounded and the tuple chain is uniformly ergodic,
$\|c(\tau)\|\le C\kappa^{|\tau|}$ in both directions. For independent and identically distributed (i.i.d.) tuples, $c(\tau)=0$
for every $\tau\ne0$. The deployed optimizer weights this function by a kernel fixed by its two
rates and its delay, with $(\cdot)_+:=\max\{\cdot,0\}$ and $\ind\{\cdot\}$ the indicator,
\begin{equation}
\label{eq:kernel}
  w_{\rho,\lambda,m}(\tau)
  :=(1-\lambda)\sum_{k\ge0}\lambda^{k}(1-\rho)^{\tau-m+k}\ind\{\tau-m+k\ge0\}
  =\frac{(1-\lambda)\,\lambda^{(m-\tau)_+}\,(1-\rho)^{(\tau-m)_+}}{1-\lambda(1-\rho)} ,
\end{equation}
so that $\rho\,w_{\rho,\lambda,m}(\tau)$ is the weight $\rho(1-\rho)^{i}$ that the accumulator
$v_{t-m}$ puts on $F(\zeta_{t-m-i})$ times the weight $(1-\lambda)\lambda^{k}$ that the momentum
$u_t$ puts on $\varepsilon_{t-k}$, summed over the pairs $i,k\ge0$ with lag $m+i-k=\tau$.

\begin{theorem}[First-order displacement under momentum, on dependent data]
\label{thm:momentum}
Let Assumption~\ref{a:data} hold and fix an integer $m\ge0$. There are
$\rho_0>0$ and $C<\infty$, depending only on $\epsilon$, $G$, $d$, $L_g$, $\lambda_A$, $R-\|\thetast\|$, the mixing constants of
Assumption~\ref{a:ergodic}, and $m$---not on $\lambda$---such that for every $0<\rho<\rho_0$ and every
$\lambda\in[0,1)$, the mean field of \eqref{eq:momentum} at delay $m$ has a locally unique
zero $\thetast_{\rho,m,\lambda}$ near $\thetast$, and
\begin{equation}
\label{eq:momentumfirst}
  \thetast_{\rho,m,\lambda}-\thetast\;=\;-\rho\,J_0^{-1}B_{\rho,\lambda,m}+r,
  \qquad
  B_{\rho,\lambda,m}:=\sum_{\tau\in\mathbb Z}w_{\rho,\lambda,m}(\tau)\,c(\tau),
  \qquad \|r\|\le C\rho^{3/2}.
\end{equation}
At fixed $\lambda$, as $\rho\downarrow0$, the kernel tends pointwise to $\lambda^{(m-\tau)_+}$. Consequently,
$B_{\rho,\lambda,m}\to B^{\lambda}_m:=\sum_{\tau}\lambda^{(m-\tau)_+}c(\tau)$, and
$\|B^\lambda_m\|\le C_B(1+m)\max(\kappa,\lambda)^{m}$, with $C_B$ independent of $m$ and $\lambda$.
\end{theorem}

The proof is in Section~\ref{app:thmmomentum}.

\emph{Decaying gain.} The fixed-gain displacement vanishes as $\rho\downarrow0$, though the updates
remain preconditioned. Letting the gain decay, as AdamNC
\citep{reddi2018} and Adafactor \citep{shazeer2018} do, can remove it: under the cited conditions for stochastic optimization with martingale-difference
noise, the averaged iterate has the limit law of a fit whose updates are not
rescaled, with $\rho_t=c/t$ \citep[Thm.~5.4]{anhuo2025a} and, under momentum, a suitably decaying momentum gain
\citep[Thm.~3]{anhuo2026b}. No such result is proved here for TD learning. A
decaying-gain refit also replaces the deployed run's value, which may already have been used in a
contract or comparison, whereas our remedy keeps that run.

\emph{Without momentum.} At $\lambda=0$, the limit $B^\lambda_m$ is the sum of the cumulants at
lags $\tau\ge m$. In this case, \eqref{eq:momentum} reduces to \eqref{eq:update} with $P_t=\phi(v_{t-m})$, and the kernel
\eqref{eq:kernel} is $(1-\rho)^{\tau-m}\ind\{\tau\ge m\}$. Replacing $(1-\rho)^{\tau-m}$ by one
changes $B_{\rho,0,m}$ by $O(\rho)$, since $\|c(\tau)\|\le C\kappa^{|\tau|}$, so for fixed $m$,
Theorem~\ref{thm:momentum} gives, for the zero $\thetast_{\rho,m}:=\thetast_{\rho,m,0}$,
\begin{equation*}
  \thetast_{\rho,m}-\thetast=-\rho\,J_0^{-1}B_m+O(\rho^{3/2}),
  \qquad
  B_m:=\sum_{\tau\ge m}c(\tau)=\E_\mu\big[D\phi(\bar v)[F(\zeta)-\bar v]\,(\Qcal^{m}\hat\varepsilon)(\zeta)\big],
\end{equation*}
where $\hat\varepsilon:=\sum_{k\ge0}\Qcal^{k}\varepsilon$ solves the Poisson equation
$\hat\varepsilon-\Qcal\hat\varepsilon=\varepsilon$ and is bounded \eqref{eq:poissonbounds}.

The coefficient $B_0$ splits into a \emph{same-observation channel}, created because the
divisor includes the current observation, and a \emph{memory channel}, carried by earlier
observations still in the accumulator,
\begin{equation*}
  B_{\mathrm{same}}:=B_0-B_1=c(0),
  \qquad
  (B_{\mathrm{same}})_j=-\tfrac12(\bar v_j+\epsilon)^{-3/2}\,\E_\mu[\varepsilon_j(\zeta)^3],
  \qquad
  B_{\mathrm{mem}}:=B_1=\sum_{\tau\ge1}c(\tau).
\end{equation*}
The same-observation channel $B_{\mathrm{same}}$ is a weighted third moment of the residuals:
it vanishes for symmetric residuals, and without momentum, the lagged update gives it weight zero
at every gain.

The memory channel $B_{\mathrm{mem}}$, which a lagged update without momentum does not remove, requires both dependence and
approximation error. It vanishes for i.i.d.\ tuples, and $\|B_m\|\le C\kappa^{m}$ for $m\ge1$
with $C$ free of $m$, so it decays with the delay at least as fast as the trajectory mixes. It also
vanishes when the features represent the value exactly: $(\Qcal\varepsilon)(\zeta)$ is
$-\varphi(x')$ times the expected one-step Bellman residual of $V_{\thetast}$ at the successor
state $x'$, which is zero when $V_{\thetast}$ is the rule's value. AdaShift \citep{zhou2019}, ACProp
\citep{zhuang2021}, and ADOPT \citep{taniguchi2024} use versions of the lagged update to
converge. By itself, the lag leaves this channel, which persists in the storage model, whose price
chain mixes slowly (Section~\ref{sec:numerics}).

\begin{corollary}[First-order displacement for independent tuples]
\label{cor:iid}
Let Assumptions~\ref{a:bounded} and~\ref{a:nondegen} hold and the tuples be i.i.d.\ draws from $\mu_\zeta$. Then, for $0<\rho<\rho_0$,
$\thetast_{\rho,m,\lambda}-\thetast=-\rho\,\frac{(1-\lambda)\lambda^{m}}{1-\lambda(1-\rho)}\,J_0^{-1}c(0)+r$ with $\|r\|\le C\rho^{3/2}$,
so the first-order displacement vanishes if and only if $\E_\mu[\varepsilon_j^{3}]=0$ for
every $j$, or $m\ge1$ and $\lambda=0$.
\end{corollary}

On one trajectory, consecutive tuples share a state, since the successor state of one is the
starting state of the next, so the adjacent-lag cumulants $c(1)$ and $c(-1)$ need not vanish.

\emph{For i.i.d.\ tuples} (Corollary~\ref{cor:iid}), momentum leaves the current update's
first-order displacement unchanged. The lagged update removes the same-observation channel
without momentum but keeps the fraction $\lambda=\beta_1$ of it with momentum, nine tenths at
Adam's default. The dependent taxi model shows about the same ratio: with $\beta_1=0.9$, its best rule's lagged displacement is about $0.90$ times its displacement under the current update (Table~\ref{tab:taxirev}).
The corollary is proved in Section~\ref{app:thmmomentum}.

\emph{Under dependence and the current update}, momentum also adds a \emph{forward} channel,
$B^\lambda_0=B_0+\sum_{i\ge1}\lambda^{i}c(-i)$. This channel sums the scaled covariances between a
residual and the squared residuals that enter the accumulator after it, damped by
$\beta_1^{\,i}$. It is identically zero under independence, so no analysis on
i.i.d.\ data, including the symmetry theorem of \citet{dereich2025sym}, can produce it. In the storage model, momentum
makes the deployed run's wrong-sign contrast about six to twelve percent larger in magnitude (Table~\ref{tab:caisorun}). In the taxi
model, it shrinks the best rule's displacement at all but the smallest gain (Table~\ref{tab:taxirev}).

\emph{A delay longer than both the momentum and mixing time scales} makes the first-order
coefficient small: by Theorem~\ref{thm:momentum}, $\|B^\lambda_m\|\le\mathrm{tol}$ for a tolerance
$\mathrm{tol}>0$ once $C_B(1+m)\max(\kappa,\lambda)^{m}\le\mathrm{tol}$. AdaShift's
temporal shift \citep{zhou2019} lags the divisor in this way, and Theorem~\ref{thm:momentum}
gives the size of the displacement the lag by itself leaves. How small the gain must be for the first-order
term to dominate depends on $m$, as the theorem's constants do.

\subsection{Convergence to the displaced target}
\label{sec:limit}

Theorem~\ref{thm:momentum} locates the zero of the field. Proving that the recursion converges to it requires
one more condition. Fix the gain, the delay $m\ge0$, and the momentum rate $\lambda$, and write
$h_\rho$ for the field of \eqref{eq:momentum} and $\thetast_\rho$ for its zero near $\thetast$.
There exist a threshold $\rho_0>0$ free of $\lambda$ and, for every $\rho<\rho_0$, a
radius $\varrho_\rho>0$ and a symmetric positive-definite matrix $\Qly_\rho$ such that $h_\rho$ is coercive in
the metric $\Qly_\rho$, that is, $(\theta-\thetast_\rho)^\top\Qly_\rho h_\rho(\theta)\ge\tfrac14\|\theta-\thetast_\rho\|^2$, on the
ball $\Bcal(\thetast_\rho,\varrho_\rho)$, which we call the \emph{basin} (Lemma~\ref{lem:basin}).

\begin{assumption}[Basin confinement]
\label{a:deployedstable}
For the gain in use, $\rho<\rho_0$, there is a closed convex set $\Theta\subseteq\Bcal(\thetast_\rho,\varrho_\rho)$, with
$\thetast_\rho$ in its interior, such that the \emph{confinement event}
$\mathcal E_\Theta:=\{\theta_t\in\Theta\text{ for every }t\}$ has positive probability.
\end{assumption}

The convergence results that use this assumption hold almost surely \emph{on} $\mathcal E_\Theta$, as is
standard for local convergence in stochastic approximation \citep[Thms.~5.2.1 and~6.1.1]{kushner2003},
and the limit laws hold conditionally on $\mathcal E_\Theta$. The assumption
restricts attention to the paths on which the \emph{unprojected} recursion stays in $\Theta$, where the coercivity holds.

We neither construct $\Theta$ from the primitives nor bound $\Pr(\mathcal E_\Theta)$ below in
terms of them, and we use none of the usual devices for boundedness: projection onto a known
set \citep{bhandari2018}, an ordinary differential equation argument \citep{borkarmeyn2000}, or
the verifiable conditions of \citet{andrieu2005}. The recursion projected onto $\Theta$, which the proofs use,
satisfies every statement unconditionally
(Theorem~\ref{thm:deployedclt}).

\begin{theorem}[Consistency at the displaced target]
\label{thm:deployedcons}
Let Assumptions~\ref{a:data} and~\ref{a:deployedstable} hold, let
$0<\rho<\rho_0$ be constant, and run \eqref{eq:momentum} at a delay $m\ge0$ from initial
accumulators in $[0,G^{2}]^{d}$ and an initial momentum of norm at most $G$. Then
\begin{equation}
\label{eq:deployedcons}
  \theta_t\;\longrightarrow\;\thetast_{\rho}
  \qquad\text{almost surely on }\mathcal E_\Theta,
\end{equation}
and the Jacobian $J:=\nabla h_{\rho}(\thetast_{\rho})$ is positive stable.
\end{theorem}

The deployed run's estimate is asymptotically normal, but around the displaced target.
Under the same assumptions, the Polyak--Ruppert average satisfies moment bounds and, conditionally on
$\mathcal E_\Theta$, a functional central limit theorem at $\thetast_\rho$
(Theorem~\ref{thm:deployedclt}, proved with Theorem~\ref{thm:deployedcons} in
Section~\ref{app:deployedcons}). The limit covariance is the sandwich $J^{-1}\Gamma_PJ^{-\top}$, where $\Gamma_P$ is the
long-run variance of $\phi(v_{t-m})u_t$ frozen at $\thetast_\rho$.

Both results hold at a
fixed gain, not jointly in $(n,\rho)$, with constants not uniform as $\rho\downarrow0$ or
$\lambda\uparrow1$. They hold for the current update ($m=0$), the one software libraries use, and for the
lagged one. The arguments behind the lagged update require the divisor to be fixed before the current observation
arrives, which need not hold at $m=0$. The proof instead treats the tuple, accumulator, and momentum
as one Markov chain controlled by the parameter, so only the parameter must be fixed before that observation arrives.

The convergence and limit-law results use Assumption~\ref{a:deployedstable}, as does the
large-sample conclusion of the regret theorem (Theorems~\ref{thm:deployedcons}, \ref{thm:deployedclt},
\ref{thm:rr}, and~\ref{thm:regret}(iv)).

\begin{remark}[The sample size at which the displacement is one standard error]
\label{rem:n1se}
For a contrast $\ell$ fixed in advance, put $\sigma_\rho^2(\ell):=\ell^\top J^{-1}\Gamma_PJ^{-\top}\ell$, the limit variance of $\sqrt n\,\ell^\top(\bth_n-\thetast_\rho)$
given $\mathcal E_\Theta$ (Theorem~\ref{thm:deployedclt}). Assume $\sigma_\rho^2(\ell)\to\sigma_0^2(\ell)>0$. The displacement is one asymptotic standard error in magnitude at
\[
  n_{\mathrm{1se}}(\rho,\ell):=\frac{\sigma_\rho^2(\ell)}{\{\ell^\top(\thetast_\rho-\thetast)\}^2}
  \;\sim\;\frac{\sigma_0^2(\ell)}{\chi(\ell)^2}\,\rho^{-2},
  \qquad \chi(\ell):=-\ell^\top J_0^{-1}B^{\lambda}_m,
\]
if $\chi(\ell)\ne0$, since $\ell^\top(\thetast_\rho-\thetast)=\rho\,\chi(\ell)+O(\rho^{3/2})$ at fixed $\lambda$ (Theorem~\ref{thm:momentum}), and $n_{\mathrm{1se}}=\infty$ if $\ell^\top(\thetast_\rho-\thetast)=0$. When $\chi(\ell)\ne0$, the exponent is the same on every problem, but $\chi(\ell)$
depends on the transition law, features, rewards, delay, momentum, and contrast. (If $\chi(\ell)=0$ and $\ell^\top b_2\ne0$, with $b_2$ from Theorem~\ref{thm:expansion}, $n_{\mathrm{1se}}$ is of order $\rho^{-4}$.) So a gain small enough for one problem need not be small enough for another.
\end{remark}

\subsection{What the displacement costs a decision}
\label{sec:contrast}

A comparison of two rules need not cancel the displacement, because each rule has its own. A comparison cancels only bias common to both sides, and two rules evaluated with the same
optimizer configuration generate their own data, so each has its own
chain, stationary law, cumulant function \eqref{eq:cumulant}, and coefficient $B^{\lambda}_{m}$. For a
rule $i$ with LSTD value $V_i=\ell^\top{\thetast}^{(i)}$, write
\begin{equation}
\label{eq:deltai}
  \delta_i:=\ell^\top J_{0,i}^{-1}B^{\lambda}_{m,i},
  \qquad
  V_i(\rho):=\ell^\top{\thetast}^{(i)}_{\rho,m,\lambda}=V_i-\rho\,\delta_i+r_i(\rho),
  \qquad |r_i(\rho)|\le C_i\rho^{3/2},
\end{equation}
so that $\delta_i$, of either sign, is the first-order downward displacement of rule $i$'s reported value per unit of gain.

\begin{theorem}[Selection regret and wrong pairwise orderings]
\label{thm:regret}
Let $I$ be a finite set of two or more rules, each satisfying the hypotheses of
Theorem~\ref{thm:momentum} at a common $(m,\lambda)$ and for a common functional $\ell$, with $\rho<\rho_0$, where $\rho_0$ is the smallest of the rules' thresholds. Let
$i^\star\in\arg\max_iV_i$ be an LSTD-best rule, $\hat\imath(\rho)\in\arg\max_iV_i(\rho)$ the
rule the deployed optimizer selects, $\Delta:=V_{i^\star}-\max_{i\ne i^\star}V_i\ge0$ the LSTD
margin, $\mathrm{spread}(\delta):=\max_i\delta_i-\min_i\delta_i$, and
$C:=\max_iC_i$, and let $R(\rho):=V_{i^\star}-V_{\hat\imath(\rho)}$ be the regret. Then:
\begin{enumerate}[label=(\roman*),leftmargin=2.2em,itemsep=2pt]
\item $0\le R(\rho)\le\rho\,(\delta_{i^\star}-\delta_{\hat\imath(\rho)})+2C\rho^{3/2}
      \le\rho\,\mathrm{spread}(\delta)+2C\rho^{3/2}$.
\item $R(\rho)=0$ whenever $\rho\,\mathrm{spread}(\delta)+2C\rho^{3/2}<\Delta$. The left side
      is nondecreasing in $\rho$, so this holds for every $\rho<\rho_{\mathrm{crit}}$, where
      $\rho_{\mathrm{crit}}:=\sup\{\rho'\in(0,\rho_0):\rho'\,\mathrm{spread}(\delta)+2C\rho'^{3/2}<\Delta\}$,
      with $\rho_{\mathrm{crit}}:=0$ when $\Delta=0$. When $\mathrm{spread}(\delta)>0$,
      $\rho_{\mathrm{crit}}=\{\Delta/\mathrm{spread}(\delta)\}\{1+O(\Delta^{1/2})\}$ as $\Delta\downarrow0$
      with $\mathrm{spread}(\delta)$, $C$, and $\rho_0$ held fixed.
\item A close pair can be ordered wrongly. For two rules, put
      $\Delta_{12}:=V_1-V_2$ and $\mathcal D:=\delta_1-\delta_2$, so that
      \begin{equation}
      \label{eq:reversal}
        V_1(\rho)-V_2(\rho)\;=\;\Delta_{12}-\rho\,\mathcal D+r_1(\rho)-r_2(\rho).
      \end{equation}
      Fix $\eta\in(0,1)$. There is $\rho_\eta>0$, depending on $\eta$, $\rho_0$, and $C/\min|\mathcal D|$, with the
      minimum taken over the pairs of $I$ with $\mathcal D\ne0$ (if none, (iii) is vacuous), such that for every $\rho<\rho_\eta$ and
      every such pair with $\operatorname{sign}\Delta_{12}=\operatorname{sign}\mathcal D$ and
      $0<|\Delta_{12}|\le(1-\eta)\rho|\mathcal D|$, the deployed ordering of the pair is the
      opposite of the LSTD ordering, and on the two-rule set $\{1,2\}$, the regret is $|\Delta_{12}|$.
\item If each recursion satisfies the hypotheses of Theorem~\ref{thm:deployedcons} at its
      displaced target and the event that all stay in their basins
      (Assumption~\ref{a:deployedstable}) has positive probability, then, almost surely on it, the estimated ordering of any two rules agrees with the
      deployed ordering for all large $n$, whenever the deployed ordering is strict. Under the conditions of (iii), the estimates therefore report the wrong ordering.
\end{enumerate}
\end{theorem}

\begin{proof}
Theorem~\ref{thm:momentum} applied to each rule on its own data and contracted with $\ell$ gives
\eqref{eq:deltai}, and differencing gives \eqref{eq:reversal}. Here, $\mathcal D$ is a difference of
rule-specific objects, so no identity forces it to vanish.

(i) $R(\rho)\ge0$ as $i^\star$ maximizes $V_i$. By the definition of $\hat\imath$ and \eqref{eq:deltai},
$V_{\hat\imath}-\rho\delta_{\hat\imath}+r_{\hat\imath}\ge V_{i^\star}-\rho\delta_{i^\star}+r_{i^\star}$, so
$R(\rho)\le\rho(\delta_{i^\star}-\delta_{\hat\imath})+|r_{\hat\imath}|+|r_{i^\star}|$, and $\delta_{i^\star}-\delta_{\hat\imath}\le\mathrm{spread}(\delta)$.

(ii) If $\hat\imath\ne i^\star$, then $V_{\hat\imath}\le\max_{i\ne i^\star}V_i$, so $R(\rho)\ge\Delta$, which
contradicts (i) when its right side is below $\Delta$. The map $f(\rho'):=\rho'\,\mathrm{spread}(\delta)+2C\rho'^{3/2}$ is
continuous and nondecreasing, so $\{\rho'\in(0,\rho_0):f(\rho')<\Delta\}=(0,\rho_{\mathrm{crit}})$. If $\mathrm{spread}(\delta)>0$ and
$f(\rho_0)>\Delta$, then $f(\rho_{\mathrm{crit}})=\Delta$, so $\rho_{\mathrm{crit}}\le\Delta/\mathrm{spread}(\delta)$, and inserting this in
$\Delta=\rho_{\mathrm{crit}}\{\mathrm{spread}(\delta)+2C\rho_{\mathrm{crit}}^{1/2}\}$ gives
$\rho_{\mathrm{crit}}\ge\{\Delta/\mathrm{spread}(\delta)\}\{1-2C\,\mathrm{spread}(\delta)^{-3/2}\Delta^{1/2}\}$.

(iii) Put $\rho_\eta:=\min\{\rho_0,(\eta\min|\mathcal D|/(2C))^{2}\}$ ($\rho_\eta:=\rho_0$ if $C=0$ or no pair has $\mathcal D\ne0$), so that
\mbox{$|r_1(\rho)-r_2(\rho)|$}${}\le2C\rho^{3/2}<\eta\rho|\mathcal D|$ for $\rho<\rho_\eta$. With $\varsigma:=\operatorname{sign}(\mathcal D)=\operatorname{sign}(\Delta_{12})$, $\varsigma(\Delta_{12}-\rho\mathcal D)=|\Delta_{12}|-\rho|\mathcal D|\le-\eta\rho|\mathcal D|$, so
$\Delta_{12}-\rho\mathcal D$ has the sign opposite to $\Delta_{12}$ and magnitude at least
$\eta\rho|\mathcal D|$, which exceeds the remainder in magnitude. The contrast \eqref{eq:reversal} therefore has
the sign opposite to $\Delta_{12}$. On the two-rule set, the worse rule is selected, with
regret $|\Delta_{12}|$. In a larger set of rules, the pair is ordered wrongly but the set's regret need not change.
The condition $\eta>0$ cannot be dropped: at $|\Delta_{12}|=\rho|\mathcal D|$, the leading term vanishes, and the
remainder determines the ordering. The theorem does not claim that wrongly ordered pairs exist, but the storage and taxi models contain them.

(iv) Theorem~\ref{thm:deployedcons} gives
$\theta_{t,i}\to{\thetast}^{(i)}_{\rho,m,\lambda}$ almost surely on rule $i$'s confinement event, and
on the intersection of those events, the Polyak--Ruppert averages converge to the same limits by
Ces\`aro's lemma, so \mbox{$\ell^\top(\bth_{n,1}-\bth_{n,2})$}${}\to V_1(\rho)-V_2(\rho)$ almost surely. When that
limit is nonzero, the sign is continuous there, so almost every path in that intersection has an index beyond which
the estimated contrast has the limit's sign. Under the conditions of (iii), that sign is the wrong one.
\end{proof}

Part (iii) identifies which \emph{pairs} must be ordered wrongly at a given gain, not the full set of
gains at which a given pair is ordered wrongly: $\rho_\eta$ is fixed first, and part (iii)'s condition is then checked at each $\rho<\rho_\eta$.

With momentum, lagging keeps the regret bound first order in the gain:
taking $m=1$ in Theorem~\ref{thm:momentum} puts $B^{\lambda}_{1}=\sum_{\tau}\lambda^{(1-\tau)_+}c(\tau)$
in place of $B^{\lambda}_{0}$, so \eqref{eq:deltai} holds with $\delta_i$ formed from the lagged
kernel and Theorem~\ref{thm:regret} applies unchanged.

The regret has a bound of first order in the gain, is zero below a threshold, and is positive on
each wrongly ordered pair's two-rule set. \emph{In the worst case}, the bound's leading coefficient is
the spread of the displacement coefficients across the set of rules, which can be large when a rule's $J_{0,i}$ is nearly singular, that is, when its mean field is weak in some direction
(Section~\ref{sec:revcaiso}). \emph{Below a threshold, the regret is zero}: for a fixed set of rules with a unique LSTD-best rule ($\Delta>0$), the deployed optimizer selects that rule at every gain below $\rho_{\mathrm{crit}}$. \emph{Counts of wrong
orderings count two-rule sets with positive regret}: each wrongly ordered pair is such a set
(Section~\ref{sec:reversal}).

More data does not reduce this ordering error, as part (iv) shows. It raises confidence in the
wrong ordering. The displacement term $\rho\mathcal D$ is fixed once the rates are chosen, while the
sampling error of the estimated contrast shrinks at rate $n^{-1/2}$ (Theorem~\ref{thm:deployedclt}), so a
longer run narrows the interval around the wrong-signed contrast without revealing the
error. Here, consistency recovers the ordering of the displaced targets, not necessarily the LSTD ordering (Section~\ref{sec:intropriorwork}). When the two
are reversed, the estimates report the wrong order for all large $n$, almost surely on the intersection of the confinement events.

The average of the TD estimating function over one fit, $n^{-1}\sum_{t\le n}g(\theta_t,\zeta_t)$, converges almost surely
on the confinement event to $A(\thetast_\rho-\thetast)$, of order at most $\rho$. This signal can show \emph{whether} the fit is
displaced but not, by itself, which ordering is right, so the remedy of Section~\ref{sec:remedies} uses two runs.

\begin{proposition}[A continuous decision parameter: decision displacement and LSTD-value loss]
\label{prop:knob}
Let the rules be indexed by $z$ in an open set $T\subset\R^{p}$, with LSTD value $V(z)$,
displacement coefficient $\delta(z)$, and deployed value ${V_\rho(z):=V(z)-\rho\,\delta(z)+r(z,\rho)}$
as in \eqref{eq:deltai}. Assume that $V$ and $\delta$ are $C^{2}$ on $T$ and that
$r(\cdot,\rho)=O(\rho^{3/2})$ in $C^{1}$ on compacts. Let $z^\star$ be an interior maximizer of $V$
with $H:=\nabla^{2}V(z^\star)\prec0$. Then there is a neighborhood $\mathcal O$ of $z^\star$
such that for all small $\rho$, the deployed value $V_\rho$ has a local maximizer in $\mathcal O$, and
every critical point $\hat z_\rho$ of $V_\rho$ in $\mathcal O$, in particular every such
maximizer, satisfies
\[
  \hat z_\rho-z^\star=\rho\,H^{-1}\nabla\delta(z^\star)+o(\rho),\qquad
  V(z^\star)-V(\hat z_\rho)=\tfrac12\rho^{2}\,\nabla\delta(z^\star)^\top(-H)^{-1}\nabla\delta(z^\star)+o(\rho^{2}),
\]
while the tuned rule's reported value is
$V_\rho(\hat z_\rho)=V(z^\star)-\rho\,\delta(z^\star)+O(\rho^{3/2})$.
\end{proposition}

The proof, a classical perturbation argument, is in Section~\ref{app:knob}.

For a rule with a continuous decision parameter, such as a base-stock target, the displacement thus costs LSTD value
only at second order in the gain (cf.\ \citealt[Thm.~3.4]{tan2022}, an upper bound of this order in estimate-then-optimize)
but can misstate the tuned rule's reported value, the number that enters a contract, at first order by $\rho\,|\delta(z^\star)|$
(Theorem~\ref{thm:regret} treats finite sets of rules). The regularity in
$z$ is assumed, not derived: Theorem~\ref{thm:momentum} is proved for a fixed rule and does not
control how its remainder varies with the rule.

\subsection{The second-order expansion, and what two runs cancel}
\label{sec:expansion}

The expansion is taken in the effective gain because the kernel
\eqref{eq:kernel} has a common scalar factor across all channels and lags,
\begin{equation}
\label{eq:kernelfactor}
  \rho\,w_{\rho,\lambda,m}(\tau)=s\,\lambda^{(m-\tau)_+}(1-\rho)^{(\tau-m)_+},
\end{equation}
with $s$ the effective gain \eqref{eq:seff}. At fixed $\beta_1$, an expansion in $\rho$ is
equally valid, but $s=\rho-\{\beta_1/(1-\beta_1)\}\rho^{2}+O(\rho^{3})$, so the two expansions can differ at second order.
We extend Theorem~\ref{thm:momentum} through second order in $s$.

\begin{theorem}[Second-order expansion of the deployed target]
\label{thm:expansion}
Under the hypotheses of Theorem~\ref{thm:momentum}, for fixed $m$ and $\lambda$ and every
$\rho<\rho_0$,
\begin{equation}
\label{eq:expansion}
  \thetast_{\rho,m,\lambda}-\thetast\;=\;s\,b_1+s^{2}\,b_2+O(s^{3}),
  \qquad
  b_1=-J_0^{-1}B^{\lambda}_{m},
\end{equation}
where $b_2$ \eqref{eq:b2} is built from expectations, under $\mu$, of bounded
functions of the tuple chain summed over lags against polynomial or geometric weights. It combines the second
derivative of the $\rho\to0$ field at $\thetast$ in the direction $b_1$, the first-order correction
to its Jacobian, the derivative in $s$ of the lag weights of \eqref{eq:kernelfactor}, the third cumulant of two squared-residual deviations and the residual, and the pairing of the
accumulator's variance with the kernel. Neither $b_1$ nor $b_2$ depends on $\rho$, and
the constant in $O(s^{3})$ depends on $m$, $\epsilon$, $G$, $d$, $L_g$, $\lambda_A$,
$R-\|\thetast\|$, the mixing constants, and $\lambda$.
\end{theorem}

The proof, by cumulant counting, is in Section~\ref{app:expansion}.

\begin{corollary}[Richardson--Romberg extrapolation in the effective gain]
\label{cor:rr}
Under the hypotheses of Theorem~\ref{thm:expansion}, fix $(m,\lambda)$ and $\omega>0$, $\omega\ne1$, and take two gains below
$\rho_0$ with effective gains $s$ and $\omega s$. Then
\begin{equation}
\label{eq:rrtarget}
  \frac{\omega\,\thetast(s)-\thetast(\omega s)}{\omega-1}\;=\;\thetast-\omega\,s^{2}b_2+O(s^{3}),
\end{equation}
writing $\thetast(s)$ for the deployed target at effective gain $s$. In particular, the
downward pair $\omega=\tfrac12$ gives $2\thetast(s/2)-\thetast(s)=\thetast-\tfrac12s^{2}b_2+O(s^{3})$,
and the three nodes $(s,s/2,s/4)$ with weights $(\tfrac13,-2,\tfrac83)$ give $\thetast+O(s^{3})$.
\end{corollary}

\begin{proof}
Weight \eqref{eq:expansion} at $s$ and $\omega s$ by $\omega/(\omega-1)$ and $-1/(\omega-1)$: the weights sum to one
and cancel $sb_1$, leaving $(\omega-\omega^{2})s^{2}b_2/(\omega-1)=-\omega s^{2}b_2$. At the nodes $s_i=s,s/2,s/4$, the weights
solve $\sum_i\nu_i=1$, $\sum_i\nu_is_i=\sum_i\nu_is_i^{2}=0$, the Vandermonde system of
\citet[Cor.~4 and (26)]{huo2026}.
\end{proof}

The two-run combination is step (3) of the remedy, and the three-run combination is step (4). Neither combination needs an
estimate of the displacement's sign, and both include the deployed run. Pairing \emph{downward} leaves a quarter of the
$s^{2}$ term in the residual of pairing upward, keeps both nodes in the expansion's regime whenever the deployed one is,
and is always feasible, unlike doubling $s$, which can violate $s<1-\beta_1$.

Extrapolating in $\rho$ instead of $s$ also cancels the first-order term but, under momentum,
leaves the second-order term $-\omega\rho^{2}\{b_2-\beta_1b_1/(1-\beta_1)\}$ in place of
$-\omega s^{2}b_2$. The extra term involves the unknown first-order coefficient $b_1$, which the design in $s$ never needs. On two
computable instances
(Section~\ref{app:rrcheck}, Table~\ref{tab:rrcheck}), the design in $\rho$
leaves 13--15\% of the displacement at $\rho=0.01$ and about 40\% at $0.04$ with $\beta_1=0.9$,
where the design in $s$ on the same two runs leaves under 3\% and under 7\%. On the i.i.d.\ instance with momentum,
the residual of the design in $s$ has log-log slopes of $1.9$ to $2.0$ against $\rho$, consistent with second order.

\begin{table}[!htp]
\centering
\small
\flushcaption
\begin{tabular*}{0.9\textwidth}{@{\extracolsep{\fill}}clrrr@{\extracolsep{0pt}\ }r@{\extracolsep{\fill}\hspace{2\tabcolsep}}r@{\extracolsep{0pt}\ }r@{\extracolsep{\fill}\hspace{2\tabcolsep}}r@{\extracolsep{0pt}\ }r@{}}
\toprule
\multicolumn{1}{@{}c}{instance} & \multicolumn{1}{c}{$(m,\beta_1)$} & \multicolumn{1}{c}{$\rho$} & \multicolumn{1}{c}{disp.\ at $\rho$} & \multicolumn{2}{c}{RR2 in $\rho$} & \multicolumn{2}{c}{RRs2 in $s$} & \multicolumn{2}{c@{}}{RRs3 in $s$} \\
\midrule
\multirow{6}{*}{Markov} & $(0,\,0)$ & $0.01$ & $-0.0299$ & 0.6\% & [0.4$\sigma$] & same &  & 0.4\% & [0.2$\sigma$] \\
 & $(0,\,0)$ & $0.04$ & $-0.1204$ & 0.5\% & [1.1$\sigma$] & same &  & 3.3\% & [7.6$\sigma$] \\
 & $(0,\,0.9)$ & $0.01$ & $-0.0410$ & 13.0\% & [12.0$\sigma$] & 2.6\% & [2.4$\sigma$] & 0.8\% & [0.8$\sigma$] \\
 & $(0,\,0.9)$ & $0.04$ & $-0.1349$ & 40.4\% & [115.7$\sigma$] & 2.6\% & [7.6$\sigma$] & 17.8\% & [54.7$\sigma$] \\
 & $(1,\,0.9)$ & $0.01$ & $-0.0394$ & 14.2\% & [11.8$\sigma$] & 1.2\% & [1.0$\sigma$] & 1.4\% & [1.2$\sigma$] \\
 & $(1,\,0.9)$ & $0.04$ & $-0.1290$ & 39.0\% & [101.0$\sigma$] & 4.9\% & [13.1$\sigma$] & 13.4\% & [36.2$\sigma$] \\
\midrule
\multirow{6}{*}{i.i.d.} & $(0,\,0)$ & $0.01$ & $-0.0115$ & 2.1\% & [1.1$\sigma$] & same &  & 0.3\% & [0.1$\sigma$] \\
 & $(0,\,0)$ & $0.04$ & $-0.0474$ & 7.0\% & [15.1$\sigma$] & same &  & 0.5\% & [1.2$\sigma$] \\
 & $(0,\,0.9)$ & $0.01$ & $-0.0107$ & 14.5\% & [7.1$\sigma$] & 0.9\% & [0.4$\sigma$] & 1.3\% & [0.7$\sigma$] \\
 & $(0,\,0.9)$ & $0.04$ & $-0.0348$ & 37.8\% & [59.2$\sigma$] & 6.9\% & [11.0$\sigma$] & 5.8\% & [9.3$\sigma$] \\
 & $(1,\,0.9)$ & $0.01$ & $-0.0095$ & 13.9\% & [7.2$\sigma$] & 1.6\% & [0.8$\sigma$] & 0.4\% & [0.2$\sigma$] \\
 & $(1,\,0.9)$ & $0.04$ & $-0.0312$ & 38.0\% & [63.4$\sigma$] & 6.6\% & [11.1$\sigma$] & 5.2\% & [8.7$\sigma$] \\
\bottomrule
\end{tabular*}
\caption{Extrapolation in $\rho$ and in the effective gain $s$ on two computable instances. Absolute residual displacement of the extrapolated estimator as a share of the absolute single-run
displacement at the deployed gain, and in the estimator's own Monte Carlo standard errors ($\sigma$)
(Section~\ref{app:rrcheck}): RR2 extrapolates in $\rho$ on the nodes $(\rho,2\rho)$, RRs2 in $s$ on the same two runs, and RRs3 in $s$ on the runs at $(\rho,2\rho,4\rho)$, both with the weights for those runs' effective gains. Rows marked ``same'' have
$s=\rho$, so the two designs coincide. Without momentum, two runs remove the displacement to
within Monte Carlo error at $\rho=0.01$. With momentum, at $\rho=0.01$, the design in $\rho$ leaves 13--15\% of it
and the design in $s$ under 3\%, on both instances and both updates. On the Markov instance,
RRs3 has a larger absolute residual than RRs2 at $\rho=0.04$.}
\label{tab:rrcheck}
\end{table}

\subsection{The joint limit law on one trajectory, and the regret after the remedy}
\label{sec:rrlaw}

Let $\hat\theta^{\mathrm{RR}}_n$ be the combination \eqref{eq:rrhat}, with the two recursions run on
one trajectory from the same initialization and differing only in $\beta_2$. For the run at effective
gain $s$, let $J_s$ and $\Gamma_s$ be the Jacobian $J$ of Theorem~\ref{thm:deployedcons} and the long-run variance $\Gamma_P$
defined after it, at the displaced target $\thetast(s)$, and let $\Gamma_{\times}$ be the
long-run cross-covariance of $\phi(v_{t-m})u_t$ in the two recursions under the joint law frozen at their two
targets.

\begin{theorem}[Extrapolation on one trajectory: limit law and variance]
\label{thm:rr}
Let the hypotheses of Theorem~\ref{thm:deployedcons} hold for both gains at a common $(m,\lambda)$,
both recursions being driven by one trajectory, and let the event $\mathcal E$ that
both recursions stay in their basins (Assumption~\ref{a:deployedstable}) have positive probability.
The limits below hold conditionally on $\mathcal E$.
\begin{enumerate}[label=(\roman*),leftmargin=2.2em,itemsep=2pt]
\item Jointly,
  $\sqrt n\big(\bth_n(s)-\thetast(s),\ \bth_n(\omega s)-\thetast(\omega s)\big)
   \dto\mathcal N\big(0,\ \mathbf J^{-1}\boldsymbol\Gamma\,\mathbf J^{-\top}\big)$
  with $\mathbf J=\Diag(J_s,J_{\omega s})$ and $\boldsymbol\Gamma$ the block matrix with diagonal
  blocks $\Gamma_s$ and $\Gamma_{\omega s}$, upper-right block $\Gamma_\times$, and lower-left block $\Gamma_\times^\top$. The functional
  version holds in $D([0,1],\R^{2d})$ for the partial-sum processes
  $n^{-1/2}\sum_{t\le\lfloor n\xi\rfloor}(\theta_t(s)-\thetast(s))$ and
  $n^{-1/2}\sum_{t\le\lfloor n\xi\rfloor}(\theta_t(\omega s)-\thetast(\omega s))$.
\item Consequently, with the weights $\nu=(\omega,-1)/(\omega-1)$ and the corrected target
  $\thetast_{\mathrm{RR}}(s):=(\omega\thetast(s)-\thetast(\omega s))/(\omega-1)$,
  \begin{equation}
  \label{eq:rrclt}
    \sqrt n\big(\hat\theta^{\mathrm{RR}}_n-\thetast_{\mathrm{RR}}(s)\big)\dto\mathcal N(0,\Sigma_{\mathrm{RR}}),
    \qquad
    \Sigma_{\mathrm{RR}}=\sum_{a,b\in\{1,2\}}\nu_a\nu_b\,J_{s_a}^{-1}\Gamma_{ab}J_{s_b}^{-\top},
  \end{equation}
  $s_1=s$, $s_2=\omega s$, $\Gamma_{11}=\Gamma_s$, $\Gamma_{22}=\Gamma_{\omega s}$,
  $\Gamma_{12}=\Gamma_{21}^\top=\Gamma_\times$. By Corollary~\ref{cor:rr},
  $\thetast_{\mathrm{RR}}(s)=\thetast-\omega s^{2}b_2+O(s^{3})$.
\item Fix a contrast $\ell$ chosen before the data are examined. Put
  $Y_t:=\ell^\top\{\nu_1\theta_t(s)+\nu_2\theta_t(\omega s)\}$, $\bar Y_n:=n^{-1}\sum_{t\le n}Y_t=\ell^\top\hat\theta^{\mathrm{RR}}_n$, and
  $\widehat\sigma_{\mathrm{RR},n}^{2}:=n^{-2}\sum_{k=1}^{n}\big[\sum_{t\le k}(Y_t-\bar Y_n)\big]^{2}$.
  If $\ell^\top\Sigma_{\mathrm{RR}}\ell>0$, then
  $\sqrt n\{\bar Y_n-\ell^\top\thetast_{\mathrm{RR}}(s)\}/\widehat\sigma_{\mathrm{RR},n}$ converges
  in distribution to $W(1)/\{\int_0^1\bar W(\xi)^{2}d\xi\}^{1/2}$, with $W$ and $\bar W$ scalar, so that
  \begin{equation}
  \label{eq:rrinterval}
    \ell^\top\hat\theta^{\mathrm{RR}}_n\;\pm\;n^{-1/2}\,q_{1-\alpha_{\mathrm{sig}}/2}\,\widehat\sigma_{\mathrm{RR},n}
  \end{equation}
  is an asymptotic $1-\alpha_{\mathrm{sig}}$ interval for $\ell^\top\thetast_{\mathrm{RR}}(s)$ at significance level $\alpha_{\mathrm{sig}}\in(0,1)$,
  where $q_p$ is the $p$-quantile of that limit ($q_{0.975}=6.747$).
\end{enumerate}
\end{theorem}

The proof is in Section~\ref{app:rr}. Part (i) applies the functional central limit theorem at
the displaced target (Theorem~\ref{thm:deployedclt}) to the stacked recursion on the product state. Its mean-field Jacobian is block diagonal and positive stable.
The stacking device is taken from \citet[Thm.~5]{zeng2026}, whose design stacks recursions at several
constant \emph{parameter} step sizes. Part (ii) centers the combination at the corrected target, which the
second-order expansion then places within $O(s^{2})$ of $\thetast$. The $O(s^{3})$ term is a
property of that target and never enters the limit.

Part (iii)'s interval uses only the combined iterates and estimates no long-run variance. It is
the self-normalized interval of \citet[Cor.~6 and (26)]{zeng2026} for a scalar contrast, applied to
the combined path. Its pivot is that of \citet{kiefer2000} \citep[see also][]{shao2010,lee2022}.

The same functional limit gives the two-block batch-means interval of \citet{glynniglehart1990},
which step (3) of the remedy reports. If
$\bar Y_{\mathrm I}$ and $\bar Y_{\mathrm{II}}$ are the means of $Y_t$ over the first and second halves of the iterates entering the average,
the functional limit of part (i) makes the components of $\sqrt{n}\,(\bar Y_{\mathrm I}-\ell^\top\thetast_{\mathrm{RR}}(s),\,\bar Y_{\mathrm{II}}-\ell^\top\thetast_{\mathrm{RR}}(s))$
asymptotically independent Gaussians with a common variance. By continuous mapping,
\begin{equation}
\label{eq:rrbm}
  \ell^\top\hat\theta^{\mathrm{RR}}_n\;\pm\;t_{1,1-\alpha_{\mathrm{sig}}/2}\,\tfrac12|\bar Y_{\mathrm I}-\bar Y_{\mathrm{II}}|
\end{equation}
is an asymptotic $1-\alpha_{\mathrm{sig}}$ interval for $\ell^\top\thetast_{\mathrm{RR}}(s)$, with
$t_{1,0.975}=12.706$ the quantile of Student's $t$ with one degree of freedom, under the same
condition $\ell^\top\Sigma_{\mathrm{RR}}\ell>0$.

Both pivots also apply to the four-run combination that compares two rules run on independent
trajectories, if its asymptotic variance is positive. There, part (i) holds for each rule's pair, the two pairs are independent, and the
pivots apply conditionally on the intersection of the two confinement events.

On one trajectory, the combination's measured variance cost is well below that of independent
runs, although the theorem does not bound it. With independent trajectories and equal asymptotic
variances, $\Gamma_\times$ would vanish and the combination's variance would be
$(\omega^{2}+1)/(\omega-1)^{2}$ times a single run's. At $\omega=2$ or $\tfrac12$, this is a factor of five in variance and
$\sqrt5\approx2.2$ in standard error. On one trajectory, the standard error of the two-run estimate is within a tenth of a single run's on the computable instances of
Table~\ref{tab:rrcheck} and, under the current update, at most $1.51$ times it in the storage and taxi models (Section~\ref{sec:reversal}).

\begin{corollary}[Selection regret after extrapolation]
\label{cor:rrregret}
Let every rule in the set satisfy the hypotheses of Theorem~\ref{thm:expansion} at a common
$(m,\lambda)$, fix $\omega>0$ with $\omega\ne1$, write $s_0:=s(\rho_0,\lambda)$ with $\rho_0$ the smallest of the rules' thresholds, and let each rule's
corrected target be $V^{\mathrm{RR}}_i:=\ell^\top\thetast_{\mathrm{RR},i}(s)$ with $s,\omega s<s_0$. Then
$V^{\mathrm{RR}}_i=V_i+O(s^{2})$, and Theorem~\ref{thm:regret}(i)--(ii) holds for selection by the corrected
targets with $(\rho,\delta_i,C\rho^{3/2})$
replaced by $(\omega s^{2},\delta^{(2)}_i,C's^{3})$, $\delta^{(2)}_i:=\ell^\top b_{2,i}$, and
$\rho<\rho_{\mathrm{crit}}$ by $s<s_{\mathrm{crit}}:=\sup\{s'\in(0,\min(s_0,s_0/\omega)):\omega s'^{2}\,\mathrm{spread}(\delta^{(2)})+2C's'^{3}<\Delta\}$,
with $C'$ as in the proof and $s_{\mathrm{crit}}:=0$ when $\Delta=0$, except that
$s_{\mathrm{crit}}=\{\Delta/(\omega\,\mathrm{spread}(\delta^{(2)}))\}^{1/2}\{1+O(\Delta^{1/2})\}$
as $\Delta\downarrow0$ if $\mathrm{spread}(\delta^{(2)})>0$, for fixed $\omega$, $\lambda$, spread, $C'$, and $s_0$.
\end{corollary}

\begin{proof}
Apply Theorem~\ref{thm:expansion} to each rule on its own data and contract with $\ell$.
Since $b_{1,i}=-J_{0,i}^{-1}B^{\lambda}_{m,i}$, the first-order term is $-s\,\delta_i$ with
$\delta_i$ as in \eqref{eq:deltai}, so the value at effective gain $s$ is
\[
  V_i(s)=V_i-s\,\delta_i+s^{2}\delta^{(2)}_i+O(s^{3}) .
\]
Both runs evaluate the same rule at the same $(m,\lambda)$, so both carry the \emph{same}
$\delta_i$. Combining them with the weights of \eqref{eq:rrhat}, therefore, cancels that term,
\begin{equation}
\label{eq:VRR}
  V^{\mathrm{RR}}_i=\frac{\omega V_i(s)-V_i(\omega s)}{\omega-1}
  =V_i-\omega\,s^{2}\delta^{(2)}_i+r^{\mathrm{RR}}_i(s),
  \qquad |r^{\mathrm{RR}}_i(s)|\le C'_is^{3},
\end{equation}
where $C'_i$ depends on $\omega$ and on the constants of Theorem~\ref{thm:expansion}, and so
carries the factor $1/(1-\lambda)$ that the second-order expansion has and the first-order one
does not (Section~\ref{app:expansion}). At Adam's $\beta_1=0.9$, this factor is ten. This is
\eqref{eq:rrtarget} contracted with $\ell$, and it gives $V^{\mathrm{RR}}_i=V_i+O(s^{2})$.

The regret bound is now Theorem~\ref{thm:regret} applied to \eqref{eq:VRR} in place of
\eqref{eq:deltai}, with $(\rho,\delta_i,C\rho^{3/2})$ replaced by
$(\omega s^{2},\delta^{(2)}_i,C's^{3})$ and $C':=\max_iC'_i$. Writing
$\hat\imath^{\mathrm{RR}}$ for a maximizer of $V^{\mathrm{RR}}$ and inserting \eqref{eq:VRR} into both sides of
$V^{\mathrm{RR}}_{\hat\imath^{\mathrm{RR}}}\ge V^{\mathrm{RR}}_{i^\star}$,
\[
  R^{\mathrm{RR}}(s)=V_{i^\star}-V_{\hat\imath^{\mathrm{RR}}}
  \le\omega s^{2}\big(\delta^{(2)}_{i^\star}-\delta^{(2)}_{\hat\imath^{\mathrm{RR}}}\big)+2C's^{3}
  \le\omega s^{2}\,\mathrm{spread}(\delta^{(2)})+2C's^{3},
\]
which is part~(i) with the stated replacement. Part~(ii) follows likewise: the regret is zero
once the right side falls below $\Delta$, and since that side is nondecreasing in $s$, this holds
for every $s<s_{\mathrm{crit}}$, the expansion being valid there. The
bound concerns the corrected targets, not a finite-run estimate of them.

With $\varsigma:=\mathrm{spread}(\delta^{(2)})>0$ and $\bar s:=\{\Delta/(\omega\varsigma)\}^{1/2}$, for $\Delta$ small enough that $s_{\mathrm{crit}}<\min(s_0,s_0/\omega)$, $s_{\mathrm{crit}}$ solves $\omega\varsigma s^{2}+2C's^{3}=\Delta$. So $s_{\mathrm{crit}}\le \bar s$, $\Delta\le s_{\mathrm{crit}}^{2}(\omega\varsigma+2C'\bar s)$, and $s_{\mathrm{crit}}\ge \bar s(1+y)^{-1/2}\ge \bar s(1-y/2)$, where $y:=2C'(\omega\varsigma)^{-3/2}\Delta^{1/2}$. If also $\mathrm{spread}(\delta)>0$, $\rho_{\mathrm{crit}}\sim\Delta/\mathrm{spread}(\delta)$ and $s_{\mathrm{crit}}/\rho_{\mathrm{crit}}$ grows as $\Delta^{-1/2}$.
\end{proof}

\section{Numerical evidence}
\label{sec:numerics}

We first measure the displacement in models fitted to electricity-price and taxi-trip records,
then ask, for each, whether the deployed optimizer ranks rules in the wrong order and whether the remedy
of Section~\ref{sec:recipe} restores the correct order.

\subsection{How large the displacement is in models fitted to operational data}
\label{sec:calib}

In fits to real price and trip data, the displacement can be large. The fitted memory coefficient varies widely
across parts of the price record, and higher-order terms matter at the larger tested gains. This is why the remedy
extrapolates instead of estimating first-order coefficients.

We use models fitted to real data, not raw trajectories, as environments, so that the LSTD
target $\thetast$ is known and coverage can be checked against it. The quantities of interest are population quantities of
the fitted model, evaluated by exact enumeration or, when the recursion is run, estimated by simulation
to a stated precision. Discretization error remains, but sampling uncertainty from the record is not propagated.

We use two data sources that differ in the properties the theory in Section~\ref{sec:tddisp} says
drive displacement. Skewed residuals matter because the same-observation channel is a
weighted third moment of the residuals. Slow mixing and approximation error matter because the
memory channel fades with the lag at least as fast as the chain mixes, and it vanishes when the
features represent the value exactly.

We compare a slow-mixing price series with faster-mixing taxi records: eight months
of five-minute real-time locational marginal prices from the California Independent System
Operator (CAISO) at its SP15 trading hub for southern California ($69{,}696$ intervals in 2024) and January 2024
New York City Taxi and Limousine Commission (TLC) yellow-taxi records, analyzed both in five-minute bins (\emph{fares} below) and, for the first $400{,}000$ trips, trip by trip (\emph{trips}). All three series are fitted the same way
(replication package).

\begin{table}[!htp]
\centering
\small
\flushcaption
\begin{tabular}{p{3.4cm}lrrr}
\toprule
category & quantity & CAISO $5$-min & TLC $5$-min & TLC trip \\
\midrule
\multirow{6}{=}{\raggedright fitted primitives} & observations & $69{,}696$ & $8{,}910$ & $400{,}000$ \\
& $\max_j|\text{skew}(\varepsilon_j)|$ & $3.13$ & $3.11$ & $2.54$ \\
& $\kappa$ & $0.9880$ & $0.8227$ & $0.0282$ \\
& spectral time scale $(1-\kappa)^{-1}$ & $83.6$ & $5.6$ & $1.0$ \\
& $\|B_{\mathrm{same}}\|$ & $1.72$ & $1.82$ & $1.53$ \\
& $\|B_{\mathrm{mem}}\|$ & $0.118^{\ast}$ & $0.114$ & $0.185$ \\
\midrule
\multirow{3}{=}{\raggedright with Adam's $\beta_1=0.9$ (Theorem~\ref{thm:momentum})} & $\|B^{\lambda}_{0}\|$, current & $0.423$ & $1.51$ & $2.01$ \\
& $\|B^{\lambda}_{1}\|$, lagged & $0.371$ & $1.36$ & $1.82$ \\
& forward channel $\|\sum_{i\ge1}\lambda^{i}c(-i)\|$ & $1.47$ & $1.69$ & $0.335$ \\
\midrule
\multirow{4}{=}{\raggedright run displacement\newline at $\beta_2=0.95$\newline in s.e.\ at $n=4\times10^{5}$} & current update, $\beta_1=0$ & $1.85$ & $0.61$ & $23.7$ \\
& lagged update, $\beta_1=0$ & $1.85^{\ast}$ & $1.51$ & $1.93$ \\
& current update, $\beta_1=0.9$ & $14.3$ & $18.1$ & $18.8$ \\
& lagged update, $\beta_1=0.9$ & $12.7$ & $16.4$ & $17.1$ \\
\bottomrule
\end{tabular}
\caption{Fitted primitives and implied displacement for the CAISO and TLC series. Each series is fitted to a finite chain on quantile bins with an
empirical within-bin reward law and $\gamma=0.9$, so every entry of the first two blocks is exact for the fitted model.
$\kappa$ is the second-largest eigenvalue modulus of the fitted kernel. The middle block
evaluates the kernel of Theorem~\ref{thm:momentum} at Adam's $\beta_1=0.9$: the $\rho\to0$
coefficients $B^{\lambda}_{m}$ under the current update ($m=0$, the one RMSProp and Adam
use) and the lagged one ($m=1$), and the forward channel $\sum_{i\ge1}\lambda^{i}c(-i)$ they
contain, which is absent without momentum. The last block measures the displacement at $\beta_2=0.95$
in standard errors: each entry is $|\ell^\top(\bar\theta-\thetast)|$ for the contrast
$\ell=d^{-1/2}\ind$, the displacement of the recursion \eqref{eq:momentum} run on the fitted model
(signed mean of $64$ replications, Monte Carlo $95$ percent interval within $\pm0.07$), divided by the asymptotic standard
error (s.e.) of LSTD after $n=4\times10^{5}$ transitions, $(\ell^\top A^{-1}\Gamma A^{-\top}\ell/n)^{1/2}$
with $\Gamma$ the long-run variance of the residual. An entry above two is a displacement larger
than two standard errors. $^{\ast}$Depends at first order on $B_{\mathrm{mem}}$, which varies widely across parts of the price record (third observation below).}
\label{tab:calib}
\end{table}

At $\beta_2=0.95$ and Adam's $\beta_1=0.9$, the displacement is substantial in all three fits (Table~\ref{tab:calib}):
the current update gives displacements of $14$, $18$, and $19$ LSTD
standard errors for the price fit and the two taxi fits at $n=4\times10^{5}$.

First, \emph{skewness does not separate the three fits}: each has a residual coordinate whose skewness
exceeds $2.5$ in absolute value, so strongly asymmetric residuals are not particular to electricity.

Second, \emph{mixing does separate the three fits}. The price fit has $\kappa=0.988$, compared
with $0.028$ for trips, giving spectral time scales of about $84$ five-minute steps and one trip.
Yet trips have the largest $\|B_{\mathrm{mem}}\|$ of the three: consecutive tuples share a state, since
the successor state of one is the starting state of the next, so they stay dependent even when the state
chain mixes quickly, and the memory channel need not vanish, whereas it vanishes for independent tuples (Corollary~\ref{cor:iid}).

Third, $\|B_{\mathrm{mem}}\|$ \emph{varies widely across parts of this slow-mixing
record}. Refitted on each of eight disjoint equal parts of the record, the model keeps $\kappa$ in
$[0.984,0.991]$ and $\|B_{\mathrm{same}}\|$ within a factor of two, but
$\|B_{\mathrm{mem}}\|$ ranges from $0.045$ to $1.89$, a factor of $42$.

The inverse in the Poisson equation can amplify variation across refits. The memory
coefficient depends on the Poisson solution through $(I-\Qcal)^{-1}$ on centered functions, which
is ill-conditioned when an eigenvalue on that space approaches one. The mixing heuristic
$n(1-\kappa)$ gives a rough effective sample size of about $830$, not $69{,}696$. Thus, report
$B_{\mathrm{mem}}$ with its spread across different pricing nodes or separately fitted
environments as a sensitivity check, not as a point estimate from one trajectory alone.

Fourth, \emph{momentum alters magnitude and sign, and the primitives, not the optimizer alone,
determine the direction} (Theorem~\ref{thm:momentum} and the middle block of Table~\ref{tab:calib}).
For the five-minute series, the forward channel, absent for independent
tuples, has norm $90\%$ and $97\%$ of $\|B_{\mathrm{same}}+B_{\mathrm{mem}}\|$, respectively,
versus a fifth on trips. It opposes this sum on prices and fares, reducing the current
coefficient's norm to about a quarter and seven eighths of its momentum-free value. On trips, it
reinforces the sum, increasing the norm by a fifth. In the storage model, however,
momentum enlarges the featured pair's displacement (Section~\ref{sec:revcaiso}).

With momentum at $\beta_1=0.9$, lagging leaves most of the coefficient's norm: the ratios
$\|B^\lambda_1\|/\|B^\lambda_0\|$ are $0.88$, $0.90$, and $0.91$ in column order, near the fraction $\beta_1$ that Corollary~\ref{cor:iid} gives for independent tuples.

At $\beta_2=0.999$, runs of \eqref{eq:momentum} on the three fits reproduce the two current-update sign changes that first order predicts: on fares, the displacement goes from $+0.006$ without momentum to $-0.041$ with it,
and on prices, from $+0.22$ to $-0.03$. Among the eleven cells passing the run-length check (a cell passes if its estimates at full and quarter length
differ by at most three Monte Carlo standard errors of their difference), eight agree with the first-order prediction within two Monte Carlo standard errors (Section~\ref{app:expt}). In the other three, the first-order prediction is eight percent too large for prices under the current update without
momentum and about half the simulated value in two lagged cells.

At $\beta_2=0.99$ and $0.95$, higher-order terms (Theorem~\ref{thm:expansion})
become important, particularly in the slow-mixing price fit. Simulated values differ from the first-order prediction by at most ten percent on trips ($\kappa=0.028$) and at most a factor of $1.61$ on fares ($\kappa=0.82$), except
for the current update at $0.95$ without momentum, where the signs differ. For prices
($\kappa=0.988$) at $0.95$, simulation gives about a twelfth of the first-order prediction under the current update
without momentum and six to eleven times it in the other three cells. Thus, the last block
reports simulated displacement and motivates applying the remedy of Section~\ref{sec:remedies}.

\subsection{Consequences for decisions, and what the remedy restores}
\label{sec:reversal}

In both models, the deployed optimizer orders close rules wrongly and misstates the selected
rule's value, but it does not change which rule is selected. Among the simulated pairs, the remedy of
Section~\ref{sec:remedies} removes every wrong ordering in the two models as fitted and nearly
every one when the storage model is refitted with exact features. A storage or taxi pair is counted only if its exact and
LSTD values agree on the better rule, and a wrong ordering only if the simulated contrast has the
wrong sign by more than two Monte Carlo standard errors. Unless stated otherwise, runs simulate \eqref{eq:momentum} for $8\times10^{6}$ steps with $128$ replications per rule, under both updates, without momentum and with $\beta_1=0.9$, and average the iterates over the second half of each run. A \emph{case} is a choice of model, update, and momentum rate. The second run uses the same trajectory at $\beta_2'$ from \eqref{eq:beta2prime}. Section~\ref{app:expt} records the construction of both models, each with $\gamma=0.9$.

\emph{The selected rule does not change.} At every tested gain, the rule with the largest LSTD
value also has the largest deployed value, in simulation among the taxi model's twenty-seven rules and at first order among the
storage model's eighty-four, so the selection regret of Theorem~\ref{thm:regret} is zero in both models.

\emph{The reported value moves at first order.} The selected rule's reported value, which enters
a contract, a bid, or a capacity plan, falls under the current update without momentum by $4.1$ and
$15.3$ percent of its LSTD value at $\beta_2=0.999$ and $0.996$ in the storage model, and by $0.3$,
$3.3$, and $17$ percent at $0.999$, $0.99$, and $0.9$ in the taxi model. Exact features need not
eliminate this shift: in the taxi model refitted at zone resolution, where LSTD is exact, the current
update without momentum still displaces the best rule's value by $1.45$ percent at $\beta_2=0.999$ (Section~\ref{app:zoneperiod}).

\emph{Close rules are ordered wrongly, and the remedy restores nearly all of them.} This is not an artifact
of calibration: on eight random instances of
the Garnet benchmark of finite Markov decision processes, with all $14{,}160$ pairs measured against their LSTD ordering, first order
predicts $90$ wrong orderings at $\beta_2=0.95$ and $389$ at $0.8$, and the recursion confirms
$81$ of $89$ and $275$ of $383$ among the pairs whose policies pass the run-length check. Counting the
$11$ further pairs ordered wrongly at $0.8$, two runs reduce the wrongly ordered pairs from $81$
to $2$ and from $286$ to $42$, and three runs leave $0$ and
$5$ (Section~\ref{app:expt}).

\subsubsection{Storage dispatch on a persistent price process}
\label{sec:revcaiso}

The storage model represents battery arbitrage on the CAISO prices of Section~\ref{sec:calib},
refitted with eight price bins. It has three charge levels and the actions charge, hold, and discharge.
Its eighty-four randomized double-threshold rules give $3{,}486$ pairs, of which $2{,}411$ are
counted. The price chain is persistent ($\kappa=0.984$), and the mean field has a weak direction, with
$\|J_0^{-1}\|_{\op}$ from $496$ to $4{,}405$ across rules. This amplifies displacement, widens its spread
across rules, and makes simulating every rule too slow. We therefore scan all pairs at first order
(Table~\ref{tab:caisorev}), simulate the rules of every pair the scan orders wrongly at
$\beta_2\le0.999$ without momentum, and report one pair in full (Table~\ref{tab:caisorun}).

\begin{table}[!htp]
\centering
\small
\flushcaption
\setlength{\tabcolsep}{10pt}
\begin{tabular}{lrccrccc}
\toprule
\multicolumn{1}{c}{$\beta_2$} & \multicolumn{1}{c}{shift} & \multicolumn{1}{c}{current} & \multicolumn{1}{c}{lagged} & \multicolumn{1}{c}{rate} & \multicolumn{1}{c}{confirmed} & \multicolumn{1}{c}{strongest gap} & \multicolumn{1}{c}{$n$ for $2$ s.e.} \\
\midrule
$0.9995$ & $0.3\%$ & $\phantom{0}6$ & $\phantom{0}7$ & $0.25\%$ & --- & $0.16$ s.e. & $5.9\times10^{7}$ \\
$0.999$  & $0.6\%$ & $10$ & $10$ & $0.41\%$ & $7/10$ & $0.45$ s.e. & $7.8\times10^{6}$ \\
$0.998$  & $1.1\%$ & $14$ & $15$ & $0.58\%$ & $12/14$ & $1.00$ s.e. & $1.6\times10^{6}$ \\
$0.996$  & $2.2\%$ & $26$ & $28$ & $1.08\%$ & $19/25$ & $1.96$ s.e. & $4.2\times10^{5}$ \\
\bottomrule
\end{tabular}
\caption{Wrong dispatch-rule orderings on CAISO-calibrated primitives. A pair is counted
only when its exact discounted values and LSTD targets order it the same way ($2{,}411$ of
the $3{,}486$ pairs), so that a counted wrong ordering inverts the true ordering. Deployed targets are
computed from \eqref{eq:momentumfirst} at $\beta_1=0$ without its remainder. ``Shift'' is the norm of one reference rule's
implied current-update displacement as a fraction of its $\|\thetast\|$ (about a third of the median fraction across rules), a
rough indicator of whether the first-order expansion still applies. ``Current''
and ``lagged'' count the pairs ordered wrongly under the two updates, and ``rate'' is the
current-update count as a fraction of the counted pairs. ``Confirmed'' is the number of
current-update wrong orderings the recursion confirms, among those whose rules pass the
run-length check (Section~\ref{app:expt}). The recursion was not run at $0.9995$. ``Strongest gap'' is the size of the deployed
contrast of the most strongly misordered pair under the current update, in s.e.\ of the LSTD contrast from independent runs of
$n=4\times10^{5}$ per rule. The last column is the $n$ at which that contrast reaches two such s.e.}
\label{tab:caisorev}
\end{table}

Wrong orderings in the scan grow with the gain, from $0.25$ percent of counted pairs at $\beta_2=0.9995$ to
$1.08$ percent at $0.996$, and at $\beta_2\le0.999$, each costs about $0.65$ model units (about $\$0.82$) of exact value on average if acted on. Lagging does not reduce them: the lagged scan finds at least as many, all from the memory
channel, which persists on this slow-mixing chain.

The featured pair has a clear gap: its exact values are $52.56$ and $50.33$, and its LSTD values, $8.011$ and $7.929$, order it the same way. Storage values are in model units of about $\$1.27$ per charge level. The gap between exact and LSTD values is the approximation error of four features on twenty-four states. At the library defaults, the simulated contrast has the wrong sign, about $30$ Monte Carlo standard errors from zero, and momentum enlarges its magnitude by about six percent at the defaults and about twelve percent at $0.996$. Adam's update, with its bias corrections and $\epsilon=10^{-8}$
added after the square root, also orders the pair wrongly ($-0.274$, standard error $0.040$, from a warm start with $48$ independent replications of $2\times10^{6}$ steps per rule). The
control at $\beta_2=0.99999$ recovers the LSTD contrast $+0.082$ within Monte Carlo error.

\begin{table}[!htp]
\centering
\small
\flushcaption
\setlength{\tabcolsep}{8pt}
\begin{tabular}{llrrrrrrr}
\toprule
& & \multicolumn{2}{c}{deployed} & & \multicolumn{2}{c}{two runs} & \multicolumn{2}{c}{three runs} \\
\cmidrule(lr){3-4}\cmidrule(lr){6-7}\cmidrule(lr){8-9}
\multicolumn{1}{c}{$\beta_1$} & \multicolumn{1}{c}{$\beta_2$} & \multicolumn{1}{c}{gap} & \multicolumn{1}{c}{(s.e.)} & \multicolumn{1}{c}{at $\beta_2'$} & \multicolumn{1}{c}{gap} & \multicolumn{1}{c}{(s.e.)} & \multicolumn{1}{c}{gap} & \multicolumn{1}{c}{(s.e.)} \\
\midrule
$0$ & $0.99999$ & $+0.077$ & $(0.008)$ & $+0.078$ & $+0.079$ & $(0.008)$ & --- &  \\
 & $0.999$ & $-0.219$ & $(0.008)$ & $-0.078$ & $+0.064$ & $(0.008)$ & --- &  \\
 & $0.998$ & $-0.460$ & $(0.008)$ & $-0.219$ & $+0.022$ & $(0.008)$ & --- &  \\
 & $0.997$ & $-0.659$ & $(0.008)$ & $-0.346$ & $-0.033$ & $(0.008)$ & $+0.071$ & $(0.008)$ \\
 & $0.9965$ & $-0.745$ & $(0.008)$ & $-0.404$ & $-0.063$ & $(0.008)$ & --- &  \\
 & $0.996$ & $-0.824$ & $(0.008)$ & $-0.460$ & $-0.096$ & $(0.008)$ & $+0.062$ & $(0.008)$ \\
\midrule
$0.9$ & $0.99999$ & $+0.077$ & $(0.008)$ & $+0.078$ & $+0.079$ & $(0.008)$ & --- &  \\
 & $0.999$ & $-0.231$ & $(0.008)$ & $-0.081$ & $+0.069$ & $(0.008)$ & --- &  \\
 & $0.998$ & $-0.498$ & $(0.008)$ & $-0.229$ & $+0.040$ & $(0.008)$ & --- &  \\
 & $0.997$ & $-0.727$ & $(0.008)$ & $-0.364$ & $-0.002$ & $(0.008)$ & $+0.077$ & $(0.008)$ \\
 & $0.9965$ & $-0.829$ & $(0.008)$ & $-0.428$ & $-0.027$ & $(0.008)$ & --- &  \\
 & $0.996$ & $-0.924$ & $(0.008)$ & $-0.489$ & $-0.055$ & $(0.008)$ & $+0.073$ & $(0.008)$ \\
\bottomrule
\end{tabular}
\caption{The featured storage pair, simulated and corrected. Entries are the pair's simulated
contrast $\ell^{\top}(\bar\theta_i-\bar\theta_j)$ (Section~\ref{sec:revcaiso}) under the
current update, using $8\times10^{6}$ steps and $128$ replications, with Monte Carlo standard
errors in parentheses. Replication $r$ of both rules uses one seed, as in the storage-pairs study
(Section~\ref{app:expt}), so each contrast is paired within a replication and the replications are
independent. The LSTD contrast is $+0.082$, so a negative entry has the wrong sign,
and is a wrong ordering by the rule of Section~\ref{sec:reversal} when negative by more than two standard
errors. ``Deployed'' is the run at $\beta_2$, and ``at $\beta_2'$'' the second run on the same
trajectory at the rate that halves the effective gain, \eqref{eq:beta2prime}. ``Two runs'' is
their combination $2\bar\theta(\beta_2')-\bar\theta(\beta_2)$ of Corollary~\ref{cor:rr}, and
``three runs'' adds a third run at a quarter of the effective gain with weights
$(\tfrac13,-2,\tfrac83)$. It was run at every gain and is shown at $\beta_2=0.997$ and $0.996$. The lagged
update differs by at most $0.04$ in every column, with the same signs. The first
row of each block is a control with negligible displacement, so the simulation must reproduce
the LSTD contrast within Monte Carlo error.}
\label{tab:caisorun}
\end{table}

\begin{figure}[!htp]
\centering
\flushcaption
\includegraphics[width=\textwidth]{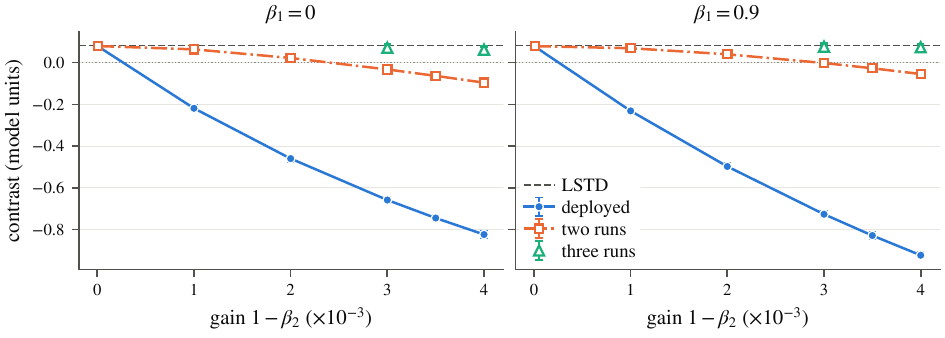}
\caption{The storage model's wrong ordering and its correction against the gain. Each panel plots the data of Table~\ref{tab:caisorun} against
$1-\beta_2$, without momentum (left) and with Adam's $\beta_1=0.9$ (right), under the current
update. Filled circles are the deployed contrast, open squares the two-run combination, and
open triangles the three-run combination at $\beta_2=0.997$ and $0.996$. Bars are $\pm2$ Monte
Carlo standard errors, the dashed line is the LSTD contrast $+0.082$, and the leftmost points
are from the control run at gain $10^{-5}$.}
\label{fig:caisorr}
\end{figure}

A single run at the smaller gain does not restore the pair's ordering. The ``at $\beta_2'$''
column of Table~\ref{tab:caisorun} is such a run: it keeps the wrong sign at every
$\beta_2\le0.999$, with or without momentum ($-0.081$ and $-0.078$ at the defaults). Even runs
four times longer, with more steps than the remedy's two runs combined, move these contrasts by at most
$0.22$ standard errors of the difference (Section~\ref{app:expt}). In Table~\ref{tab:caisorun}, one run recovers the LSTD sign
only in the control, and that run does not by itself show that the gain is small enough.

Two runs give the pair's contrast the LSTD sign at the default and at twice its gain, under both updates and with or
without momentum. Their residual roughly follows the quadratic scaling of Corollary~\ref{cor:rr}, and two
runs suffice until the deployed displacement reaches about eight to ten times the LSTD gap. Beyond that,
the three-run combination recovers the LSTD sign in every tested case, within $0.023$ of the LSTD contrast under either update (Figure~\ref{fig:caisorr}), and each combination's standard error stays within $0.001$ of a single run's. Across the simulated pairs, the recursion confirms seven of the scan's ten current-update wrong
orderings at the default, and two runs restore all ten. At $0.998$ and $0.996$, two runs leave up
to six ordered wrongly, and three runs leave none, with one pair undecided (Section~\ref{app:expt}).

Exact features make the wrong orderings more numerous, not rarer, but they involve closer pairs. With one indicator
feature per state, LSTD values equal exact values, so all $3{,}486$ pairs count. Under the current
update without momentum, the recursion orders $285$ pairs wrongly at $\beta_2=0.999$ ($8.2$
percent) and $478$ at $0.998$, with mean exact gaps of $0.19$ and $0.30$ model units per wrongly
ordered pair. The deployed values still select the LSTD-best rule, and its value falls by $0.24$ percent at the
default. Two runs leave $45$ and $152$ pairs ordered wrongly, and three runs leave $3$ and $16$.
First order identifies which rules move (correlation $0.88$ across rules) but overstates the
displacement by a median factor of five to seven, and halving the default gain reduces it by a median factor of
only $1.4$, which is why two runs fall short here (Section~\ref{app:features}).

\subsubsection{Taxi repositioning on a faster-mixing chain}
\label{sec:revtaxi}

The taxi model is a one-cab repositioning model fitted to $1{,}594{,}524$ weekday daytime trips in
New York City's January 2024 yellow-taxi records. Its eight states are regions, and its three
actions are to wait, to reposition to John F. Kennedy (JFK) Airport, or to reposition to the
Manhattan core. Twenty-seven ergodic threshold rules give $350$ counted pairs, and the contrast is
the value at the core in dollars, with discount factor $\gamma=0.9$ per decision. The region chain mixes faster than the price
chain ($\kappa=0.23$--$0.96$ across rules).

\begin{table}[!htp]
\centering
\small
\flushcaption
\setlength{\tabcolsep}{7pt}
\begin{tabular}{lccccccccccc}
\toprule
& \multicolumn{4}{c}{wrongly ordered pairs, deployed} & \multicolumn{3}{c}{best rule's displacement, \$} & \multicolumn{4}{c}{two runs, current update} \\
\cmidrule(lr){2-5}\cmidrule(lr){6-8}\cmidrule(lr){9-12}
& \multicolumn{2}{c}{current} & \multicolumn{2}{c}{lagged} & \multicolumn{2}{c}{current} & \multicolumn{1}{c}{lagged} & \multicolumn{2}{c}{misordered} & \multicolumn{2}{c}{displacement, \$} \\
\cmidrule(lr){2-3}\cmidrule(lr){4-5}\cmidrule(lr){6-7}\cmidrule(lr){8-8}\cmidrule(lr){9-10}\cmidrule(lr){11-12}
\multicolumn{1}{c}{$\beta_2$} & \multicolumn{1}{c}{$\beta_1{=}0$} & \multicolumn{1}{c}{$0.9$} & \multicolumn{1}{c}{$\beta_1{=}0$} & \multicolumn{1}{c}{$0.9$} & \multicolumn{1}{c}{$\beta_1{=}0$} & \multicolumn{1}{c}{$0.9$} & \multicolumn{1}{c}{$0.9$} & \multicolumn{1}{c}{$\beta_1{=}0$} & \multicolumn{1}{c}{$0.9$} & \multicolumn{1}{c}{$\beta_1{=}0$} & \multicolumn{1}{c}{$0.9$} \\
\midrule
$0.999$ & $0$ & $0$ & $0$ & $0$ & $-0.17$ & $-0.18$ & $-0.16$ & $0$ & $0$ & $+0.01$ & $+0.01$ \\
$0.99$ & $0$\rlap{$^{\dagger}$} & $0$ & $0$ & $0$ & $-1.68$ & $-1.61$ & $-1.44$ & $0$ & $0$ & $-0.06$ & $-0.05$ \\
$0.98$ & $1$ & $1$ & $0$ & $1$ & $-3.08$ & $-2.79$ & $-2.51$ & $0$ & $0$ & $-0.28$ & $-0.19$ \\
$0.95$ & $4$ & $4$ & $0$ & $3$\rlap{$^{\dagger}$} & $-5.99$ & $-4.91$ & $-4.40$ & $0$ & $0$ & $-1.38$ & $-0.76$ \\
$0.9$ & $5$ & $5$ & $0$\rlap{$^{\dagger}$} & $4$\rlap{$^{\dagger}$} & $-8.67$ & $-6.29$ & $-5.62$ & $1$ & $0$ & $-3.32$ & $-1.75$ \\
\bottomrule
\end{tabular}
\caption{Wrong repositioning-rule orderings on taxi-calibrated primitives, simulated and corrected.
Counts of the $350$ counted pairs that the constant-gain recursion \eqref{eq:momentum} orders wrongly,
using $8\times10^{6}$ steps and $128$ replications per rule, under both updates, without
momentum ($\beta_1=0$) and with Adam's ($\beta_1=0.9$). A dagger marks a cell with one
further pair undecided within two Monte Carlo standard errors. ``Best rule's displacement'' is
the simulated deployed value of the best rule (\#23, exact \$51.47, LSTD \$51.09) minus its
LSTD value, in dollars, with standard error at most \$0.005. ``Two runs'' gives
the count and displacement under the current update for the combination of
Corollary~\ref{cor:rr}, which adds a second run on the same trajectory at $\beta_2'$. Under
lagging, the two-run counts are zero in every cell. Pairs are counted as in Table~\ref{tab:caisorev}.}
\label{tab:taxirev}
\end{table}

Under the current update, wrong orderings appear as the gain grows: none at $\beta_2=0.999$ or $0.99$, then one, four, and
five at $0.98$, $0.95$, and $0.9$, with or without momentum (Table~\ref{tab:taxirev}). Without
momentum, these are all among the pairs first order predicts. Lagging prevents them only without
momentum: with $\beta_1=0.9$, lagging keeps $0.87$ to $0.96$ of each misordered pair's
displacement, close to the fraction $\beta_1$ of Corollary~\ref{cor:iid}.

The remedy removes every wrong ordering. With momentum, two runs leave none under either update
at any tested gain. Without momentum, they leave one of the five at $\beta_2=0.9$, where the
effective gain is largest, and the third run restores it, with a standard error $1.44$ times a single run's. Under the current update, the ratio of two-run to single-run standard error is at most $1.51$, largest at $\beta_2=0.9$ without momentum, where two runs cut the largest exact-value
cost of a misordered pair from $\$0.61$ to $\$0.03$. After two or three runs, every configuration selects the best rule.

In $10$ to $27$ of the $3{,}456$ replications per configuration and gain ($27$ rules, $128$ each), all under the lagged update without
momentum, we flag a possible \emph{confinement
failure}: in the last quarter of the run, the iterate's largest distance from its running average since mid-run exceeds ten
times the median of that largest distance over the rule's replications. Removing these replications changes no wrong-ordering count.

Adam as software libraries implement it, with $\epsilon=10^{-8}$ added after the square root and
bias correction, displaces values far more on a larger model. On the taxi model refitted at zone resolution with one
indicator feature for each of $264$ states, the best rule loses $20.8$ percent of its
value at Adam's defaults, against $1.4$ percent with our damping, and $6$ of $190$ pairs are
ordered wrongly, but the selected rule does not change. Two runs remove every wrong ordering but
only $55$ percent of the best rule's displacement, and three runs $83$ percent, because halving
the effective gain shrinks the displacement only by a factor of $1.38$, outside the regime our theory
describes. With $129$ dense features instead, the displacement is $0.39$ percent, halves
with the gain, and two runs remove it within Monte Carlo error (Section~\ref{app:highdim}).

One close pair of the eight-state model, featured here, shows how quickly a wrong ordering becomes significant. Its two rules are $\$0.55$ apart in exact value,
yet under the current update, the deployed recursion at $\beta_2=0.95$ orders them wrongly by $45$ to $92$ Monte Carlo
standard errors. Without momentum, the misordering would exceed two asymptotic LSTD standard errors after about
six weekdays of the fleet's daytime trips per rule. With or without momentum, two runs recover the
LSTD contrast within $13$ percent.

\subsubsection{Interval coverage on the two featured pairs}
\label{sec:coverage}

At $8\times10^{6}$ steps, the two-block batch-means interval of step (3) covers reliably, but whether it settles a
comparison depends on the pair. On each featured pair, with $500$ replications per case, a coverage near $95$ percent has a Monte Carlo standard error
of about one percentage point, and the band from $92.1$ to $97.9$ percent spans about three of them either
side of the nominal $95$. The two-block interval covered each of the
combination's own across-replication mean, an estimate of the corrected target, and the LSTD
contrast in $93$ to $96$ percent of replications under either update, inside the band. The
nominal $95$ percent self-normalized interval, about a quarter as wide, covered that mean in
$85$ to $93$ percent, below the band in seven of the eight cases, consistent with the
finite-sample size distortion reported for self-normalized statistics
\citep{kiefervogelsang2005,shao2015}. The single-run interval showed the same undercoverage, so
the shortfall is not unique to the combination (Section~\ref{app:tier4}).

The two-block interval's width decides which pairs it settles. On the taxi pair, with LSTD
contrast $-0.675$, its median half-width is $0.59$ to $0.61$ at $8\times10^{6}$ steps, about the
contrast's magnitude, and $0.28$ to $0.31$ at $3.2\times10^{7}$, where a typical run excludes zero.
On the storage pair, with LSTD contrast $+0.082$, the median half-width is $1.2$ to $1.3$ at
$8\times10^{6}$ steps and still $0.31$ to $0.34$ at $1.28\times10^{8}$, so a typical run does not
confirm the correct ordering at any run length we tried.

Longer runs let the second-order residual reduce the two-block interval's coverage of the LSTD
contrast. At sixteen times the original run length, its LSTD coverage fell to 70 percent on the
taxi model under the current update without momentum, where the residual was $0.60$
of the median half-width. The third run of step (4) raised that coverage to 90 and 86 percent at four and
sixteen times the original run length, still below the band, with the remaining residual at
$0.16$ and $0.28$ of the median half-width (Section~\ref{app:tier4}).

\section{Discussion}
\label{sec:discussion}

Adaptive optimizers at their usual settings can rank decision rules wrongly when the rules' values
are fitted by TD learning on correlated data, and with the rates fixed, more data only makes the wrong ranking look more
certain. The averaging rates therefore belong with the discount factor and the feature map among
the choices an analyst reports, because they can change what is estimated. A
squared-gradient averaging rate that makes the displacement negligible on one problem need not do so on another: the sample size at which the displacement reaches one
asymptotic standard error depends on the problem (Remark~\ref{rem:n1se}).

The remedy can correct the ranking without changing the deployed run. One extra run on the same
data at a second averaging rate, combined with the deployed run, removes the leading displacement
for any momentum and either update, and the two runs alone give a confidence interval for the corrected target. In the
storage and taxi models as fitted, two runs restore the ranking at moderate gains, with a standard error at most $1.51$ times a single run's under the current update, and a third run restores it at larger ones.

Three limitations matter in practice. First, the interval that covers reliably is wide. The two-block interval
is about four times as wide as the self-normalized one, which undercovers at our run lengths,
and it does not typically confirm the storage pair's correct ordering at any run length we tried. Second, two runs leave a second-order residual. At large gains, LSTD coverage declines on very long runs, and a third run slows but does not stop the decline. Third, the theory and the evidence have limited scope. The
analysis assumes a fixed rule generating the data, and its first-order description can require a gain small relative to the visit frequency of the rarest state. With Adam's library damping and indicator features, halving the gain shrinks the displacement only by a factor of about $1.4$, so two runs remove about half of it. The evidence comes from models with computable ground truth rather than
raw operational data.

\clearpage
\ECSwitch\ECnumbering\vspace*{-\baselineskip}

\noindent{\Large\bfseries Electronic Companion}

\medskip

\noindent
This electronic companion supplements the paper ``Adaptive Optimizers Can Get a Policy Ranking Wrong:
Momentum, Dependent Data, and a Two-Run Remedy.'' Section~\ref{sec:setup} sets up the tuple
chain, collects standing bounds and standard facts, and establishes the regularity of the chain
on tuples, accumulators, and momentum (Proposition~\ref{prop:controlled}). Section~\ref{app:proofs} proves
Theorems~\ref{thm:momentum} and~\ref{thm:expansion}, Corollary~\ref{cor:iid}, and Proposition~\ref{prop:knob}. Section~\ref{app:imported} proves
Theorems~\ref{thm:deployedcons} and~\ref{thm:rr}. Section~\ref{app:numerics} describes the simulation designs and results.

\bigskip

\section{The tuple chain, standing bounds, and standard facts}
\label{sec:setup}

\subsection{The temporal-difference field, the tuple chain, and the standing bounds}
\label{app:tdfield}

With $A(\zeta_t)$ and $b(\zeta_t)$ as in Section~\ref{sec:tdsetup}, the stationary mean of the TD estimating function is affine,
\begin{equation*}
  \bar{g}(\theta) \;:=\; \E_{\mu}\big[ g(\theta, \zeta) \big] \;=\; A\theta - b,
  \qquad b := \E_{\mu}\big[\varphi(x) r\big],
\end{equation*}
with $A$ as in \eqref{eq:Adef}, so $\thetast=A^{-1}b$ is the LSTD target, the TD fixed point of
\citet{tsitsiklis1997}.

The Poisson equation of the residual acts on the \emph{tuple} chain, because the estimating function is
evaluated at tuples, not states, while Assumption~\ref{a:ergodic} is stated for the state chain. For
$\zeta=(x,a,r,x')$, the next tuple is $\zeta'=(x',a',r',x'')$ with $a'\sim\pi(\cdot\mid x')$ and
$(r',x'')$ drawn jointly from the decision process given $(x',a')$, so the tuple kernel
\[
  \Qcal(\zeta,\cdot)\;=\;\iota_{x'},
  \qquad
  \iota_{x}:=\text{law of one tuple started at state } x,
\]
depends on $\zeta$ only through its terminal coordinate.

The tuple chain inherits geometric ergodicity from the state chain, with the same rate. Its
stationary law is $\mu_\zeta:=\int\iota_x\,\mu(dx)$ and
$\Qcal^{k}(\zeta,\cdot)=\int(\Pcal^{\pi})^{k-1}(x',dx)\,\iota_x$. The kernel $x\mapsto\iota_x$ does not
increase total variation, and the bounds $1\le V\le V_\infty$ relate the weighted and total-variation norms, with $\widetilde V(\zeta):=V(x')$:
\begin{equation}
\label{eq:tupleergodic}
\begin{aligned}
  \big\|\Qcal^{k}(\zeta,\cdot)-\mu_\zeta\big\|_{\widetilde V}
  &\;\le\;V_\infty\big\|\Qcal^{k}(\zeta,\cdot)-\mu_\zeta\big\|_{\mathrm{TV}}
   \;\le\;V_\infty\big\|(\Pcal^{\pi})^{k-1}(x',\cdot)-\mu\big\|_{\mathrm{TV}}\\
  &\;\le\;C_{\mathrm{mix}}V_\infty^{2}\,\kappa^{k-1},
  \qquad k\ge1.
\end{aligned}
\end{equation}
The mixing bound \eqref{eq:tupleergodic} is used in the form
\begin{equation}
\label{eq:Qcontract}
  \|\Qcal^{k}f-\mu f\|_\infty\;\le\;C_{\mathrm Q}\,\kappa^{k}\,\|f\|_\infty
  \qquad\text{for bounded $f$ on the tuple space and every }k\ge0,
\end{equation}
with $C_{\mathrm Q}:=\max\{2,\,2C_{\mathrm{mix}}V_\infty^{2}/\kappa\}$. A covariance between
bounded functions of tuples $k$ steps apart is therefore at most $C_{\mathrm Q}\kappa^{k}$ times the
product of their sup norms in absolute value: for $f$, $f'$ bounded and $f$ centered,
$|\E[f(\zeta_0)f'(\zeta_k)]|=|\E[f(\zeta_0)(\Qcal^{k}f'-\mu f')(\zeta_0)]|\le C_{\mathrm Q}\kappa^{k}\|f\|_\infty\|f'\|_\infty$.

The one-sided TD bound of Section~\ref{sec:assumptions} is used below in the form
\begin{equation}
\label{eq:Apd}
  \theta^{\top} A \theta \;\ge\; \lambda_A \|\theta\|^{2},
  \qquad \lambda_A := (1-\gamma)\lambda_{\varphi} > 0.
\end{equation}
Under Assumption~\ref{a:bounded}, the field is \emph{uniformly bounded} on $\Bcal_R:=\Bcal(0,R)$,
\begin{equation}
\label{eq:Gbound}
  \|g(\theta,\zeta)\| \;\le\; G \;=\; C_{\varphi}^{2}(1+\gamma) R + C_{\varphi} C_r ,
  \qquad \theta\in\Bcal_R .
\end{equation}
The field is Lipschitz in the parameter with constant $L_g=C_\varphi^{2}(1+\gamma)$, since
$g(\theta,\zeta)-g(\theta',\zeta)=A(\zeta)(\theta-\theta')$ and $\|A(\zeta)\|_{\op}\le L_g$. For $F_\theta:=g(\theta,\cdot)^{\odot2}$, the Jacobian
$\nabla_\theta F_\theta(\zeta)=2\Diag(g(\theta,\zeta))A(\zeta)$ has
$\|\nabla_\theta F_\theta\|_{\op}\le2GL_g=:C_S$ on $\Bcal_R$. The preconditioner map
$\phi(v):=\Diag(v+\epsilon\ind)^{-1/2}$ has derivatives bounded on $[0,\infty)^d$ by
\begin{equation}
\label{eq:phibounds}
 \|\phi\|_{\op}\le\epsilon^{-1/2},\quad
 \|D\phi\|_{\op}\le\tfrac12\epsilon^{-3/2},\quad
 \|D^2\phi\|_{\op}\le\tfrac34\epsilon^{-5/2},\quad
 \|D^3\phi\|_{\op}\le\tfrac{15}{8}\epsilon^{-7/2}.
\end{equation}
These bounds follow from differentiating each diagonal entry.

\subsection{Coercivity, preconditioned stability, and the Poisson equation}

\begin{lemma}[Standard facts]
Let $\mu$ be invariant for $\Pcal^\pi$ and let the trajectory be started in stationarity, as
Section~\ref{sec:tdsetup} assumes throughout.
\begin{enumerate}[label=(\roman*),ref=\thelemma(\roman*),leftmargin=2.2em,itemsep=2pt]
\item\label{lem:Apd}\emph{One-sided coercivity of the TD drift} \citep[Lem.~3]{bhandari2018}, after
  \citet[Lem.~9]{tsitsiklis1997}. Under Assumptions~\ref{a:bounded} and~\ref{a:nondegen},
  \eqref{eq:Apd} holds for every $\theta$. Consequently, $A$ is invertible with
  $\|A^{-1}\|_{\op}\le\lambda_A^{-1}$ and every eigenvalue of $A$ has positive real part.
\item\label{lem:posstable}\emph{Positive stability under symmetric positive-definite left
  preconditioning}. Under the assumptions of~(i), for every symmetric positive-definite $P$, $PA$ is positive stable. This follows from
  Lyapunov's theorem \citep[Thm.~2.2.1]{hornjohnson1991} with the certificate $\Qly:=P^{-1}$:
  $\Qly(PA)+(PA)^{\top}\Qly=A+A^{\top}\succeq2\lambda_AI\succ0$ by \eqref{eq:Apd}, and no symmetry
  of $A$ is used. The diagonal case, which covers
  $J_0=\phi(\bar v)A$, is $D$-stability in the sense of \citet{arrow1958}.
\item\emph{Poisson solution for the residual}. Under Assumption~\ref{a:data}, the Poisson equation
  \begin{equation*}
    \hat{\varepsilon} - \Qcal \hat{\varepsilon} \;=\; \varepsilon(\cdot)
  \end{equation*}
  has the bounded series solution $\hat\varepsilon=\sum_{k\ge0}\Qcal^{k}\varepsilon$, with
  \begin{equation}
  \label{eq:poissonbounds}
    \|\hat\varepsilon\|_{\infty} \le C_{\mathrm{Poi}}\,G,
    \qquad C_{\mathrm{Poi}}:=C_{\mathrm Q}/(1-\kappa),
  \end{equation}
  by \eqref{eq:Qcontract}. \citet{glynnmeyn1996} and
  \citet[Thm.~17.4.2]{meyntweedie2009} give the general theory.
\end{enumerate}
\end{lemma}

\begin{proof}
Parts (ii) and (iii) are proved in their statements. For (i), we follow \citet[Lem.~3]{bhandari2018}: Cauchy--Schwarz and invariance of $\mu$ give
$\theta^\top A\theta\ge(1-\gamma)\theta^\top\Sigma_\varphi\theta$, which, with $\Sigma_\varphi\succeq\lambda_\varphi I$, gives \eqref{eq:Apd}. Hence, $\Sym(A)\succeq\lambda_AI$, and
$\|Aq\|\|q\|\ge q^\top Aq$ gives $\|A^{-1}\|_{\op}\le\lambda_A^{-1}$. For a unit eigenvector $q_1+\mathrm{i}q_2$ with eigenvalue $\varsigma$,
$\Re\varsigma=q_1^\top\Sym(A)q_1+q_2^\top\Sym(A)q_2\ge\lambda_A$ (Bendixson's inequality, \citealt{bhatia1997}).
\end{proof}

\subsection{Conventions, the accumulator's fluctuation, and the iterate increment}

Throughout the proofs, $C<\infty$ may change from line to line. Unless stated otherwise, it depends only
on $\gamma$, $d$, $\epsilon$, $R$, and the constants of Assumption~\ref{a:data}. The step size is $\eta_t=\eta_0t^{-\alpha}$.

\emph{The accumulator's fluctuation.} Let $a:=1-\rho$, $\bar v(\theta):=\E_\mu F_\theta$, and
$\widetilde F_\theta:=F_\theta-\bar v(\theta)$, and on the stationary two-sided tuple chain, put
$\delta:=\rho\sum_{k\ge0}a^{k}\widetilde F_\theta(\zeta_{-1-k})$, the deviation of the stationary
accumulator from its mean, where $\widetilde F_\theta$ is centered and bounded coordinatewise by $G^{2}$. Then,
\[
\begin{aligned}
  \E\|\delta\|^{2}
  &=\rho^{2}\sum_{k,k'\ge0}a^{k+k'}\,\E\big[\widetilde F_\theta(\zeta_{-1-k})^{\top}\widetilde F_\theta(\zeta_{-1-k'})\big]\\
  &\le\rho^{2}\,dG^{4}C_{\mathrm Q}\sum_{k,k'\ge0}a^{k+k'}\kappa^{|k-k'|}
  \le\rho^{2}\,dG^{4}C_{\mathrm Q}\,\frac{2}{1-\kappa}\sum_{k\ge0}a^{2k}
\end{aligned}
\]
by \eqref{eq:Qcontract} applied coordinatewise, and since $\sum_ka^{2k}=1/(1-a^{2})\le1/\rho$,
\begin{equation}
\label{eq:vvar}
 \E\|\delta\|^2\;\le\;C_v\,\rho,
 \qquad C_v:=\frac{2dG^{4}C_{\mathrm Q}}{1-\kappa},
\end{equation}
uniformly in $\theta\in\Bcal_R$. The prefactor $\rho^2$ is balanced by the accumulator's memory, of length of order $1/\rho$, so $\E\|\delta\|^{2}$ is $O(\rho)$, not $O(\rho^{2})$ in general. When the $F_\theta(\zeta_t)$ are independent, it equals $\rho\,\E\|\widetilde F_\theta(\zeta_0)\|^{2}/(2-\rho)$.

\begin{lemma}[Iterate increment bound]
\label{lem:increment}
Under Assumption~\ref{a:bounded}, and with initial accumulators in $[0,\infty)^d$,
gains $\rho_t\in[0,1]$, and initial momentum $\|u_0\|\le G$ (Adam starts it at zero), the recursions \eqref{eq:update} and
\eqref{eq:momentum} with $P_t=\phi(v_{t-m})$ satisfy the increment bound $\|\eta_t\phi(v_{t-m})u_t\|\le\epsilon^{-1/2}G\,\eta_t$
whenever $\theta_k\in\Bcal_R$ for every $k\le t$, for every delay $m\ge0$ and every $\lambda\in[0,1)$.
\end{lemma}

\begin{proof}
The increment is $-\eta_t\phi(v_{t-m})u_t$ by \eqref{eq:momentum}, with $u_t=g(\theta_t,\zeta_t)$ when $\lambda=0$.
By induction from the initial accumulators, $v_t=(1-\rho_t)v_{t-1}+\rho_tg(\theta_t,\zeta_t)^{\odot2}\succeq0$, so $\phi(v_{t-m})$ is diagonal with
entries $(v_{t-m,j}+\epsilon)^{-1/2}\le\epsilon^{-1/2}$. Unrolling gives $u_t=\lambda^{t}u_0+(1-\lambda)\sum_{1\le k\le t}\lambda^{t-k}g(\theta_k,\zeta_k)$,
with nonnegative weights summing to one, $\|u_0\|\le G$, and $\|g(\theta_k,\zeta_k)\|\le G$ by \eqref{eq:Gbound}, as
$\|\theta_k\|\le R$. Thus, $\|u_t\|\le G$, and the bound follows.
\end{proof}

\subsection{The enlarged chain and its regularity}
\label{app:enlarged}

Both the first-order expansion and the limit laws use the enlarged chain that the recursion
\eqref{eq:momentum} drives at a frozen parameter. Its state consists of the recent accumulators, the momentum, and the
transition tuple. Its regularity is established here, with constants explicit in $\rho$ and
$\lambda$, although it need not mix in total variation. Throughout this subsection,
$\rho\in(0,1)$, $\lambda\in[0,1)$, and $m\ge0$ are fixed, and $\bar m:=\max\{m,1\}$.

At a frozen parameter $\theta\in\Bcal_R$, put $a:=1-\rho$, $g_\theta:=g(\theta,\cdot)$, and
$F_\theta:=g_\theta^{\odot2}$. The recursion carries the last $\bar m$ accumulators and the momentum, so
its state is
\[
  X_t:=(v_{t-\bar m},\dots,v_{t-1},u_{t-1},\zeta_t)\in\mathcal X:=[0,G^{2}]^{d\bar m}\times\mathcal U\times\mathcal Z,
  \qquad
  \mathcal U:=\{u\in\R^{d}:\|u\|\le G\},
\]
with $\mathcal Z$ the space of transition tuples. The set is preserved, because
$v_t=av_{t-1}+\rho F_\theta(\zeta_t)$ and $u_t=\lambda u_{t-1}+(1-\lambda)g_\theta(\zeta_t)$ are convex
combinations of points of the respective sets by \eqref{eq:Gbound}. The kernel
$\Pi_\theta$ computes $v_t$ and $u_t$ by these two maps, shifts $v_t$ into the accumulator block, and moves $\zeta$ by $\Qcal$.
The update applies the field
\[
  H_\theta(\mathsf x):=\phi(v_{t-m})\,u_t=\phi(v_{t-m})\big\{\lambda u_{t-1}+(1-\lambda)g_\theta(\zeta_t)\big\},
\]
where, at $m=0$, the accumulator $v_t=av_{t-1}+\rho F_\theta(\zeta_t)$ is itself a function of $(\theta,\mathsf x)$.
The chain and its kernel are then those of $m=1$, and only the field differs: it also depends on $\theta$
through $\phi$.

The steps below use two facts about the recursion, both valid at every $m\ge0$: $\theta_t$ is
$\Fcal_{t-1}$-measurable, and $X_{t+1}\sim\Pi_{\theta_t}(X_t,\cdot)$ given $\Fcal_t$.
Predictability of the preconditioner itself is not used, so the argument covers the current
update ($m=0$), whose divisor already includes the current observation.
For an invariant law $\pi_\theta$ of $\Pi_\theta$, write $h_\rho(\theta):=\pi_\theta H_\theta$ and
$f_\theta:=H_\theta-h_\rho(\theta)$. Given the data path, the accumulator block and the momentum, which we call the
\emph{deterministic coordinates}, evolve deterministically. The block is measured by the norm
$\max_{1\le i\le\bar m}\|v_{t-i}\|$, in which its update is nonexpansive, and the momentum by
the Euclidean norm. We write $\operatorname{Lip}_v$ and $\operatorname{Lip}_u$ for Lipschitz constants in the block and
in the momentum for these norms, uniform over the other coordinates. For $\lambda=0$, neither $H_\theta$ nor $\Pi_\theta$ depends on
$u_{t-1}$.

Given the data path, the deterministic coordinates carry no noise, so the laws of $X_k$ from two starting states can be mutually singular for every $k$. The standard argument \citep[Thm.~2]{care2019}, through the contraction
of \citet{hairer2011}, needs a minorization condition for the enlarged chain, which our assumptions do not supply. The proof uses this determinism instead: after $k$ steps, these coordinates are affine in their
starting values, with remainders that depend on the data alone.

\begin{proposition}[Controlled-chain regularity from the accumulator and the momentum]
\label{prop:controlled}
\label{prop:controlledmom}
Let Assumptions~\ref{a:ergodic} and~\ref{a:bounded} hold and let $X_1\in\mathcal X$.
\begin{enumerate}[label=(\roman*),leftmargin=2.2em,itemsep=2pt]
\item For every $\theta\in\Bcal_R$, $\Pi_\theta$ has a unique invariant law $\pi_\theta$.
\item $h_\rho$ is $C^{3}$ on $\Bcal_R$, with derivatives of orders one to three bounded uniformly in
  $\theta\in\Bcal_R$, $\rho$, and $\lambda$.
\item The Poisson equation $\hat f_\theta-\Pi_\theta\hat f_\theta=f_\theta$ has a solution $\hat f_\theta$ satisfying
  $\|\hat f\|_\infty:=\sup_{\theta\in\Bcal_R}\|\hat f_\theta\|_\infty<\infty$. This solution is continuously
  differentiable in the accumulator block and the momentum, with
  \begin{equation}
  \label{eq:hatfmomlip}
    \operatorname{Lip}_v(\hat f_\theta)\le\Big(\bar m+\frac1\rho\Big)\frac{G}{2\epsilon^{3/2}},
    \qquad
    \operatorname{Lip}_u(\hat f_\theta)\le\frac{\lambda}{(1-\lambda)\,\epsilon^{1/2}},
  \end{equation}
  and
  \begin{equation}
  \label{eq:hatflip}
    \|\hat f_{\theta}-\hat f_{\theta'}\|_{\infty}\;\le\;L_{\hat f,\rho}\,\|\theta-\theta'\|
    \qquad\text{for all }\theta,\theta'\in\Bcal_R,
  \end{equation}
  where $L_{\hat f,\rho}<\infty$ need not stay bounded as $\rho\downarrow0$ or $\lambda\uparrow1$.
\item Put $M_{\theta}(\mathsf x,\mathsf y):=\hat f_{\theta}(\mathsf y)-\Pi_{\theta}\hat f_{\theta}(\mathsf x)$ and
  $Q_{\theta}(\mathsf x):=\int M_{\theta}(\mathsf x,\mathsf y)M_{\theta}(\mathsf x,\mathsf y)^{\top}\,\Pi_{\theta}(\mathsf x,d\mathsf y)$. If
  $\theta_t\in\Bcal_R$ is $\Fcal_{t-1}$-measurable and $X_{t+1}\sim\Pi_{\theta_t}(X_t,\cdot)$
  conditionally on $\Fcal_t:=\sigma(\theta_1,X_1,\dots,X_t)$, then for every $\theta^\circ\in\Bcal_R$,
  $n^{-1}\sum_{t\le n}Q_{\theta_t}(X_t)\to\int Q_{\theta^\circ}\,d\pi_{\theta^\circ}$ almost surely on the
  event $\{\theta_t\to\theta^\circ\}$.
\end{enumerate}
\end{proposition}

\begin{proof}
With $L_g$, $C_S=2GL_g$, $C_{\mathrm Q}$, $\kappa$, and \eqref{eq:phibounds} as in Section~\ref{app:tdfield}, we have
\begin{equation}
\label{eq:controlledbasiclip}
  \|g_\theta-g_{\theta'}\|\le L_g\|\theta-\theta'\|,\qquad
  \|F_\theta-F_{\theta'}\|\le2GL_g\|\theta-\theta'\|.
\end{equation}
The second inequality follows from the coordinatewise identity $g_j^{2}-g_j'^{2}=(g_j-g_j')(g_j+g_j')$.

\emph{Step 1: affine structure and the invariant law.} Unrolling the two deterministic recursions from a generic starting state $X_0=(v_{-\bar m},\dots,v_{-1},u_{-1},\zeta_0)$ gives, for $k\ge1$,
\begin{equation}
\label{eq:momaffine}
\begin{aligned}
  v_{k-i}&=a^{k-i+1}v_{-1}+\rho\sum_{l=0}^{k-i}a^{k-i-l}F_\theta(\zeta_l)\qquad(1\le i\le\min\{\bar m,k\}),\\
  u_{k-1}&=\lambda^{k}u_{-1}+(1-\lambda)\sum_{l=0}^{k-1}\lambda^{k-1-l}g_\theta(\zeta_l).
\end{aligned}
\end{equation}
So, given the data path, once $k\ge\bar m$, the accumulator block of $X_k$ depends on the starting state only through
$v_{-1}$, with slopes at most $a^{k-\bar m+1}$, the momentum depends on it only through $u_{-1}$, with
slope $\lambda^{k}$, and both remainders are functions of the data path.

The invariant law is the law of the backward series, and it is unique. On the stationary two-sided
tuple chain, which exists by Kolmogorov's extension theorem because $\mu_\zeta$ is invariant for
$\Qcal$, the series
\[
  v^{-}_i(\theta):=\rho\sum_{l\ge0}a^{l}F_\theta(\zeta_{-i-l})\quad(0\le i\le\bar m),
  \qquad
  u^{-}(\theta):=(1-\lambda)\sum_{l\ge0}\lambda^{l}g_\theta(\zeta_{-1-l})
\]
converge absolutely and uniformly in $\theta\in\Bcal_R$ and in the path, each being a convex
combination of points of $[0,G^2]^d$ or $\mathcal U$. The point
$X^{-}(\theta):=(v^{-}_{\bar m},\dots,v^{-}_1,u^{-},\zeta_0)$ satisfies the recursion under the shift, so its
law $\pi_\theta$ is invariant. If $\pi'$ is another invariant law, a stationary version started from
it has $\zeta$-marginal $\mu_\zeta$, and \eqref{eq:momaffine} applied $k$ steps back writes its
accumulator block and momentum at time $0$ as the functionals above plus terms bounded by
$2a^{k-\bar m+1}\sqrt d\,G^{2}$ and $2\lambda^{k}G$. Letting $k\to\infty$ identifies $\pi'$ with $\pi_\theta$.
This is (i).

\emph{Step 2: smoothness of the field.} By Step 1,
$h_\rho(\theta)=\E\big[\phi(v^{-}_m(\theta))\{\lambda u^{-}(\theta)+(1-\lambda)g_\theta(\zeta_0)\}\big]$, with
$v^{-}_0(\theta)=av^{-}_1(\theta)+\rho F_\theta(\zeta_0)$ at $m=0$.
Each $F_\theta$ is a quadratic polynomial in $\theta$ with $\nabla_\theta F_\theta=2\Diag(g_\theta)A(\zeta)$
and $\nabla^{2}_\theta F_\theta[q,q']=2(A(\zeta)q)\odot(A(\zeta)q')$, bounded on $\Bcal_R$ by $C_S$ and
$2L_g^{2}$, respectively, and $g_\theta$ is affine with $\nabla_\theta g_\theta=A(\zeta)$. The weights of both series
sum to one, so $v^{-}_m$ and $u^{-}$ are $C^{\infty}$ in $\theta$ with derivatives bounded uniformly in
the path, in $\rho$, and in $\lambda$. Composing with $\phi$, whose first three derivatives are bounded
by \eqref{eq:phibounds}, gives an integrand whose first three $\theta$-derivatives are dominated
uniformly in the path, $\rho$, and $\lambda$, and dominated convergence justifies differentiating three times under
the expectation. This is (ii). In particular, $h_\rho$ is Lipschitz on $\Bcal_R$ with a constant $L_h$
free of $\rho$ and $\lambda$.

\emph{Step 3: moduli of the centered field.} On $\mathcal X$, $\|H_\theta\|\le\epsilon^{-1/2}G$. The
field depends on the accumulator block only through its oldest entry, and $\partial H_\theta/\partial
u_{t-1}=\lambda\phi(v_{t-m})$. By \eqref{eq:phibounds},
\begin{equation}
\label{eq:fmoduli}
\begin{aligned}
  &\|f_\theta\|_\infty\le2\epsilon^{-1/2}G,\qquad
  \operatorname{Lip}_v(f_\theta)\le\tfrac12\epsilon^{-3/2}G,\qquad
  \operatorname{Lip}_u(f_\theta)\le\lambda\epsilon^{-1/2},\\
  &\operatorname{Lip}_v(\nabla_vf_\theta)\le\tfrac34\epsilon^{-5/2}G,\qquad
  \operatorname{Lip}_v(\nabla_uf_\theta)+\operatorname{Lip}_u(\nabla_vf_\theta)\le\lambda\epsilon^{-3/2},\qquad
  \operatorname{Lip}_u(\nabla_uf_\theta)=0 .
\end{aligned}
\end{equation}
For $m\ge1$, $H_\theta(\mathsf x)-H_{\theta'}(\mathsf x)=(1-\lambda)\phi(v_{t-m})\{g_\theta-g_{\theta'}\}(\zeta_t)$, so by
\eqref{eq:controlledbasiclip} and Step 2,
$\|f_\theta-f_{\theta'}\|_\infty\le\{(1-\lambda)\epsilon^{-1/2}L_g+L_h\}\|\theta-\theta'\|$,
$\operatorname{Lip}_v(f_\theta-f_{\theta'})\le\frac12(1-\lambda)\epsilon^{-3/2}L_g\|\theta-\theta'\|$, and
$\operatorname{Lip}_u(f_\theta-f_{\theta'})=0$. At $m=0$, the field depends on the block through
$av_{t-1}+\rho F_\theta(\zeta_t)$, so the $v$-moduli in \eqref{eq:fmoduli} gain the factors $a$ and $a^{2}$,
at most one, and the argument of $\phi$ moves with $\theta$ by at most $2\rho GL_g\|\theta-\theta'\|$. By
\eqref{eq:phibounds}, the three parameter moduli are at most
$(1-\lambda)\epsilon^{-1/2}L_g+\rho\epsilon^{-3/2}G^{2}L_g+L_h$,
$\frac12(1-\lambda)\epsilon^{-3/2}L_g+\frac32\rho\epsilon^{-5/2}G^{2}L_g$, and
$\lambda\rho\epsilon^{-3/2}GL_g$, each multiplied by $\|\theta-\theta'\|$. All are bounded (they enter Step~6
only through $B'$).

\emph{Step 4: forgetting.} The chain forgets its starting state geometrically, at the rate
$\bar\lambda:=\max\{a,\lambda,\kappa\}^{1/2}$, for test functions with finite moduli. Let $w$ be bounded with
finite $\operatorname{Lip}_v(w)$ and $\operatorname{Lip}_u(w)$, let $\mathsf x,\tilde{\mathsf x}\in\mathcal X$, let $k\ge2\bar m$ and
$j:=\lfloor k/2\rfloor$, so that $\bar m\le j\le k-\bar m$. By \eqref{eq:momaffine} applied from time $j$, the
accumulator block and the momentum of $X_k$ differ from functions of $(\zeta_j,\dots,\zeta_k)$ alone
by at most $a^{k/2-\bar m}\sqrt d\,G^{2}$ and $\lambda^{k/2}G$. Hence,
$|\Pi_\theta^{k}w(\mathsf x)-\E_\zeta[w(\mathcal R_{j,k})]|\le\operatorname{Lip}_v(w)a^{k/2-\bar m}\sqrt d\,G^{2}+\operatorname{Lip}_u(w)\lambda^{k/2}G$,
where $\mathcal R_{j,k}$ is that function of the data and $\E_\zeta$ the expectation given
$\zeta_0=\zeta$. By the Markov property,
$\E_\zeta[w(\mathcal R_{j,k})]=(\Qcal^{j}\Psi)(\zeta)$ with $\Psi(z):=\E[w(\mathcal R_{j,k})\mid\zeta_j=z]$, and
\eqref{eq:Qcontract} gives
$|\Qcal^{j}\Psi(\zeta)-\Qcal^{j}\Psi(\tilde\zeta)|\le2C_{\mathrm Q}\kappa^{-1/2}\kappa^{k/2}\|w\|_\infty$. The
same bounds at $\tilde{\mathsf x}$, and the trivial bound $2\|w\|_\infty$ for $k<2\bar m$, give for every $k\ge1$
\begin{equation}
\label{eq:controlledforget}
  \big|\Pi_\theta^kw(\mathsf x)-\Pi_\theta^kw(\tilde{\mathsf x})\big|
  \le C_F\,\bar\lambda^{k}\big\{\|w\|_\infty+\operatorname{Lip}_v(w)+\operatorname{Lip}_u(w)\big\},
\end{equation}
with $C_F:=\max\{2\sqrt d\,G^{2}a^{-\bar m},\,2G,\,2C_{\mathrm Q}\kappa^{-1/2},\,2a^{-\bar m}\}$.
The constant is free of $\theta$ and $\lambda$ and, for $\rho\le\frac12$, at most its value at $\rho=\frac12$,
so only the rate degrades as $\rho\downarrow0$ or $\lambda\uparrow1$. The sum
$\sum_k\bar\lambda^{k}\le2/\min\{\rho,1-\lambda,1-\kappa\}$ is the source of every non-uniformity below. Integrating \eqref{eq:controlledforget} against $\tilde{\mathsf x}\sim\pi_\theta$,
\begin{equation}
\label{eq:controlledergodic}
  \big|\Pi_\theta^kw(\mathsf x)-\pi_\theta w\big|\le C_F\,\bar\lambda^{k}\big\{\|w\|_\infty+\operatorname{Lip}_v(w)+\operatorname{Lip}_u(w)\big\}.
\end{equation}

For test functions with finite moduli, both the invariant law and the kernel are Lipschitz in the parameter. First, the backward series of Step 1, evaluated at $\theta$ and
$\theta'$ on one path, move by at most $2GL_g\|\theta-\theta'\|$ in each accumulator and
$L_g\|\theta-\theta'\|$ in the momentum, so
\begin{equation}
\label{eq:pilip}
  |\pi_\theta w-\pi_{\theta'}w|\le\big\{2GL_g\operatorname{Lip}_v(w)+L_g\operatorname{Lip}_u(w)\big\}\|\theta-\theta'\|.
\end{equation}
Second, $\Pi_\theta$ and $\Pi_{\theta'}$ draw $\zeta'$ identically and differ only by the shifts
$\delta_v:=\rho\{F_\theta-F_{\theta'}\}(\zeta)$ in the newest accumulator and
$\delta_u:=(1-\lambda)\{g_\theta-g_{\theta'}\}(\zeta)$ in the momentum, which do not depend on the
deterministic coordinates. The deterministic update is affine in those coordinates and nonexpansive
in the norms fixed above. Thus, the first bound below holds for $w$ with finite moduli, and the second for $w$ continuously
differentiable in the deterministic coordinates with Lipschitz gradients:
\begin{equation}
\label{eq:kernelgap}
\begin{aligned}
  \big|(\Pi_\theta-\Pi_{\theta'})w(\mathsf x)\big|
  &\le\big\{2\rho GL_g\operatorname{Lip}_v(w)+(1-\lambda)L_g\operatorname{Lip}_u(w)\big\}\|\theta-\theta'\|,\\
  \operatorname{Lip}_v\big((\Pi_\theta-\Pi_{\theta'})w\big)+\operatorname{Lip}_u\big((\Pi_\theta-\Pi_{\theta'})w\big)
  &\le\Big\{2\rho GL_g\big[\operatorname{Lip}_v(\nabla_vw)+\operatorname{Lip}_u(\nabla_vw)\big]\\
  &\qquad+(1-\lambda)L_g\big[\operatorname{Lip}_v(\nabla_uw)+\operatorname{Lip}_u(\nabla_uw)\big]\Big\}\|\theta-\theta'\|.
\end{aligned}
\end{equation}

\emph{Step 5: the Poisson solution.} Since $\pi_\theta f_\theta=0$, \eqref{eq:controlledergodic} makes
$\hat f_\theta:=\sum_{k\ge0}\Pi_\theta^{k}f_\theta$ converge in sup norm, with
\begin{equation}
\label{eq:hatfbound}
  \|\hat f\|_\infty\le\frac{C_F}{1-\bar\lambda}\Big\{2\epsilon^{-1/2}G+\tfrac12\epsilon^{-3/2}G+\lambda\epsilon^{-1/2}\Big\},
\end{equation}
and it solves the Poisson equation term by term, with $\pi_\theta\hat f_\theta=0$.

The solution's moduli in the deterministic coordinates follow from coupling two copies using the same
data path, started at different deterministic coordinates. By \eqref{eq:momaffine}, a
difference in the accumulator block leaves a difference in the block of $X_k$ at most as large for
$k<\bar m$, the older entries being shifted and the newer ones scaled by powers of $a$, and at most
$a^{k-\bar m+1}$ times as large for $k\ge\bar m$. A difference in the momentum is multiplied by $\lambda^{k}$,
and neither coordinate feeds the other. Thus, for every $w$ with finite moduli,
$\operatorname{Lip}_v(\sum_k\Pi_\theta^kw)\le(\bar m+\rho^{-1})\operatorname{Lip}_v(w)$ and
$\operatorname{Lip}_u(\sum_k\Pi_\theta^kw)\le(1-\lambda)^{-1}\operatorname{Lip}_u(w)$ whenever the series converges.
With $w=f_\theta$, \eqref{eq:fmoduli} gives \eqref{eq:hatfmomlip}. For gradient moduli, the same slopes enter twice, or once each for mixed
moduli, and the derivative series converge uniformly. Thus, $\hat f_\theta$ is continuously
differentiable in the deterministic coordinates, with
\begin{equation}
\label{eq:hatfsecond}
\begin{aligned}
  &\operatorname{Lip}_v(\nabla_v\hat f_\theta)\le\Big(\bar m+\frac1\rho\Big)\frac{3G}{4\epsilon^{5/2}},\qquad
  \operatorname{Lip}_u(\nabla_u\hat f_\theta)=0,\\
  &\operatorname{Lip}_v(\nabla_u\hat f_\theta)+\operatorname{Lip}_u(\nabla_v\hat f_\theta)\le\Big(\bar m+\frac1{1-a\lambda}\Big)\frac{\lambda}{\epsilon^{3/2}},
\end{aligned}
\end{equation}
using $1-a^{2}\ge\rho$.

\emph{Step 6: parameter-Lipschitz continuity.} Put $D:=\hat f_\theta-\hat f_{\theta'}$. Subtracting the
two Poisson equations gives $D-\Pi_\theta D=\Phi$, with
$\Phi:=(f_\theta-f_{\theta'})+(\Pi_\theta-\Pi_{\theta'})\hat f_{\theta'}$. Since $\pi_\theta\Phi=0$ and $\Phi$
has finite moduli, \eqref{eq:controlledergodic} gives $D=\sum_{k\ge0}\Pi_\theta^{k}\Phi+\pi_\theta D$ and
\[
  \|D\|_\infty\le\frac{C_F}{1-\bar\lambda}\big\{\|\Phi\|_\infty+\operatorname{Lip}_v(\Phi)+\operatorname{Lip}_u(\Phi)\big\}+\|\pi_\theta D\| .
\]
The normalizations $\pi_\theta\hat f_\theta=0=\pi_{\theta'}\hat f_{\theta'}$ give
$\pi_\theta D=(\pi_{\theta'}-\pi_\theta)\hat f_{\theta'}$, which \eqref{eq:pilip} and \eqref{eq:hatfmomlip} bound
by $\{(\bar m+\rho^{-1})G^{2}L_g\epsilon^{-3/2}+\lambda(1-\lambda)^{-1}L_g\epsilon^{-1/2}\}\|\theta-\theta'\|$.
For $\Phi$, take $w=\hat f_{\theta'}$ in \eqref{eq:kernelgap}. With \eqref{eq:hatfmomlip} and
\eqref{eq:hatfsecond}, the gain factors cancel: $\rho(\bar m+\rho^{-1})\le1+\bar m\rho$ and
$(1-\lambda)(\bar m+(1-a\lambda)^{-1})\le1+\bar m$. Hence,
\[
  \big\|(\Pi_\theta-\Pi_{\theta'})\hat f_{\theta'}\big\|_\infty
  \le\Big\{(1+\bar m\rho)\frac{G^{2}L_g}{\epsilon^{3/2}}+\frac{\lambda L_g}{\epsilon^{1/2}}\Big\}\|\theta-\theta'\|,
\]
and its two moduli are at most $B\|\theta-\theta'\|$ for a constant $B$ depending on $(\epsilon,G,L_g,m)$
alone. With Step 3, this gives \eqref{eq:hatflip} with
\[
  L_{\hat f,\rho}\le\frac{C_F}{1-\bar\lambda}\,B'+\Big(\bar m+\frac1\rho\Big)\frac{G^{2}L_g}{\epsilon^{3/2}}+\frac{\lambda}{1-\lambda}\,\frac{L_g}{\epsilon^{1/2}},
\]
where $B'$ depends on $(\epsilon,G,L_g,L_h,m)$ alone.

\emph{Step 7: variance averaging.} $Q_\theta$ is Lipschitz in $\theta$ uniformly in $\mathsf x$, with constant
$L_Q:=8\|\hat f\|_\infty(L_{\hat f,\rho}+K)$, $K:=(1+\bar m\rho)G^{2}L_g\epsilon^{-3/2}+\lambda L_g\epsilon^{-1/2}$: in
$M_\theta(\mathsf x,\mathsf y)$, bounded by $2\|\hat f\|_\infty$, the solution moves by at most $L_{\hat f,\rho}\|\theta-\theta'\|$
in sup norm, and the shifts $\delta_v,\delta_u$ of $\mathsf y$ change $\hat f_{\theta'}(\mathsf y)$ and its kernel average by
at most $K\|\theta-\theta'\|$ by \eqref{eq:hatfmomlip}. The deterministic update being nonexpansive,
$Q_\theta$ is Lipschitz in the deterministic coordinates with constants
$8\|\hat f\|_\infty\operatorname{Lip}_v(\hat f_\theta)$ and $8\|\hat f\|_\infty\operatorname{Lip}_u(\hat f_\theta)$.

The Ces\`aro average of $Q_{\theta_t}(X_t)$ converges to its frozen mean at $\theta^\circ$ on
$\{\theta_t\to\theta^\circ\}$. Fix $\theta^\circ$, write $Q^\circ:=Q_{\theta^\circ}$, $\Gamma^\circ:=\pi_{\theta^\circ}Q^\circ$, and
$q^\circ:=Q^\circ-\Gamma^\circ$, and split
$Q_{\theta_t}(X_t)-\Gamma^\circ=\{Q_{\theta_t}(X_t)-Q^\circ(X_t)\}+q^\circ(X_t)$. On
$\{\theta_t\to\theta^\circ\}$, the first term is bounded in norm by $L_Q\|\theta_t-\theta^\circ\|\to0$, so its
average tends to zero. For the second, $\hat q^\circ:=\sum_{k\ge0}\Pi_{\theta^\circ}^{k}q^\circ$ converges by
\eqref{eq:controlledergodic}, since $\pi_{\theta^\circ}q^\circ=0$ and $q^\circ$ has finite moduli, and by the
propagation bound of Step~5, $\operatorname{Lip}_v(\hat q^{\circ})\le(\bar m+\rho^{-1})\operatorname{Lip}_v(q^{\circ})$ and
$\operatorname{Lip}_u(\hat q^{\circ})\le(1-\lambda)^{-1}\operatorname{Lip}_u(q^{\circ})$. The Poisson equation gives
\begin{equation*}
 q^{\circ}(X_t)
 =\underbrace{\hat q^{\circ}(X_t)-\hat q^{\circ}(X_{t+1})}_{\text{telescoping}}
 +\underbrace{\hat q^{\circ}(X_{t+1})-\Pi_{\theta_t}\hat q^{\circ}(X_t)}_{\text{martingale difference}}
 +\underbrace{\big\{\Pi_{\theta_t}-\Pi_{\theta^\circ}\big\}\hat q^{\circ}(X_t)}_{\text{drift}} .
\end{equation*}
Summed over $t<n$ and divided by $n$, the telescoping term is $O(1/n)$. The martingale differences
$N_{t+1}$ are bounded, so $\sum_{t<n}N_{t+1}/(t+1)$ converges almost surely as an $L^{2}$-bounded martingale, and Kronecker's lemma
gives $n^{-1}\sum_{t<n}N_{t+1}\to0$ almost surely. By \eqref{eq:kernelgap} and these moduli, the
drift is bounded in norm by
$\{2GL_g(1+\bar m\rho)\operatorname{Lip}_v(q^\circ)+L_g\operatorname{Lip}_u(q^\circ)\}\|\theta_t-\theta^\circ\|$, which tends to
zero on the event. This is (iv).
\end{proof}

\section{Proofs for the displaced target}
\label{app:proofs}

\subsection{Small-gain approximation of the mean field}

At frozen $\theta$ and a fixed gain $\rho_t\equiv\rho\in(0,1)$, the accumulator generally approaches a
stationary law rather than a fixed value, so the recursion's field is the update averaged under
the joint law of the accumulator and the tuple. Write $X_t:=(v_{t-1},\zeta_t)$ for the enlarged state of the lagged recursion without momentum,
and let $\pi_\theta$ be its stationary law and $\Pi_\theta$ the kernel that generates it. Both expectations in
$h_\rho(\theta)=\E_{\pi_\theta}[PA(\zeta)]\,\theta-\E_{\pi_\theta}[Pb(\zeta)]$, with $P=\phi(v_{t-1})$, depend on $\theta$ through $\pi_\theta$.

\begin{proposition}[Small-gain approximation of the constant-gain lagged mean field]
\label{prop:smallgain}
Let Assumption~\ref{a:data} hold and let $\rho\in(0,1)$ be constant.
Write $\bar v(\theta):=\E_{\mu}[g(\theta,\zeta)^{\odot2}]$ and
$\bar P(\theta):=\phi(\bar v(\theta))$ for the preconditioner at the mean accumulator, so that
$\bar v=\bar v(\thetast)$ and $\bar P(\thetast)=\phi(\bar v)$, and let
$h_{\rho}(\theta)=\E_{\pi_\theta}[P\,g(\theta,\zeta)]$, where $P=\phi(v^-)$ is the lagged preconditioner
of \eqref{eq:sarmsprop} evaluated at the stationary accumulator $v^-$ of part (i).
\begin{enumerate}[leftmargin=1.6em,itemsep=2pt]
\item[(i)] $\pi_\theta$ exists uniquely and is induced by the backward representation
  \begin{equation}
  \label{eq:backward}
    v^{-}(\theta) \;:=\; \rho\sum_{k\ge0}(1-\rho)^{k}\,g(\theta,\zeta_{-1-k})^{\odot2}
  \end{equation}
  on a stationary two-sided chain. The series converges uniformly on $\Bcal_R$.
\item[(ii)] There is a $C<\infty$ depending only on $\gamma$, $d$, $\epsilon$, $R$, and the constants of
  Assumption~\ref{a:data}---in particular, \emph{not} on $\rho$---such that
  for every $\rho\in(0,1)$,
  \begin{equation}
  \label{eq:smallgain}
    \sup_{\theta\in\Bcal_R}\Big\{
      \big\|h_{\rho}(\theta) - \bar P(\theta)(A\theta-b)\big\|
      + \big\|\nabla h_{\rho}(\theta) - \nabla\big[\bar P(\theta)(A\theta-b)\big]\big\|_{\op}
    \Big\} \;\le\; C\rho .
  \end{equation}
\item[(iii)] The $\rho\to0$ field is $\bar P(\theta)(A\theta-b)$. It vanishes only at $\thetast$, and
  \begin{equation}
  \label{eq:J0}
    \nabla\big[\bar P(\theta)(A\theta-b)\big]\big|_{\theta=\thetast} \;=\; \phi(\bar v)\,A\;=\;J_0 ,
  \end{equation}
  which is positive stable by Lemma~\ref{lem:posstable}.
\item[(iv)] For some $\rho_0>0$ and neighborhood $\mathcal O$ of $\thetast$, every $\rho<\rho_0$ gives a unique zero $\thetast_{\rho}$ of $h_\rho$ in $\mathcal O$, with
  $\|\thetast_{\rho}-\thetast\| = O(\rho)$ and Jacobian
  $J_{\rho} := \nabla h_{\rho}(\thetast_{\rho}) = J_0 + O(\rho)$, which is positive stable.
\end{enumerate}
\end{proposition}

\begin{proof}[Proof of Proposition~\ref{prop:smallgain}]
Throughout, $a:=1-\rho$, $g_\theta:=g(\theta,\cdot)$, $F_\theta:=g_\theta^{\odot2}$, $\bar v(\theta)=\E_\mu F_\theta$,
$\widetilde F_\theta:=F_\theta-\bar v(\theta)$,~and $\phi(v)=\Diag(v+\epsilon\ind)^{-1/2}$, with the
constants $C_{\mathrm Q}$, $\kappa$, $L_g$, $C_S$, and the derivative bounds \eqref{eq:phibounds} of
Section~\ref{app:tdfield}.

\emph{Part (i).} This is Proposition~\ref{prop:controlled}(i) at $(m,\lambda)=(1,0)$. Step~1 of its proof
establishes it by the argument of \citet[Thm.~1]{brandt1986}: \eqref{eq:backward} is the series $v^{-}_1$
of that step, with values in $[0,G^{2}]^{d}$.

\emph{Part (ii), the second moment of the accumulator.} The accumulator's deviation from its mean
and its centered sensitivity both have $O(\rho)$ second moments, uniformly on $\Bcal_R$. For
$\delta:=v^{-}(\theta)-\bar v(\theta)=\rho\sum_{k\ge0}a^{k}\widetilde F_\theta(\zeta_{-1-k})$,
\eqref{eq:vvar} gives $\E\|\delta\|^2\le C_v\rho$ uniformly in $\theta\in\Bcal_R$. The
accumulator's sensitivity is $S(\theta):=\nabla_\theta v^{-}(\theta)=\rho\sum_ka^{k}\nabla_\theta F_\theta(\zeta_{-1-k})$,
and the centered sensitivity is $\widetilde S:=S(\theta)-\nabla\bar v(\theta)$. The same computation in the
Frobenius norm, with $\widetilde F_\theta$ replaced by $\nabla_\theta F_\theta-\nabla_\theta\bar v(\theta)$, bounded in operator norm
by $2C_S$, gives $\E\|\widetilde S\|_{\op}^{2}\le\E\|\widetilde S\|_{\mathrm F}^{2}\le C_S'\rho$ for a constant $C_S'$ free of $\rho$.
Cauchy--Schwarz then bounds the cross term by
$\E\|\delta\|\|\widetilde S\|_{\op}\le(C_vC_S'\rho^{2})^{1/2}$. Also,
$\E S(\theta)=\nabla\bar v(\theta)$ because the weights sum to one.

\emph{Part (ii), the field gap.} The field differs from the $\rho\to0$ field by $O(\rho)$, uniformly on
$\Bcal_R$. By stationarity, $h_\rho(\theta)=\E[\phi(v^{-}(\theta))g_\theta(\zeta_0)]$
with $v^{-}$ built from $\zeta_{-1},\zeta_{-2},\dots$, and $\E[\phi(\bar v(\theta))g_\theta(\zeta_0)]=\bar P(\theta)\bar g(\theta)$
exactly, where $\bar g(\theta):=A\theta-b$. Taylor's theorem at $\bar v(\theta)$ gives
$\phi(\bar v(\theta)+\delta)=\phi(\bar v(\theta))+D\phi(\bar v(\theta))[\delta]+R_2$ with
$\|R_2\|_{\op}\le\tfrac12\|D^{2}\phi\|_{\op}\|\delta\|^{2}\le\tfrac38\epsilon^{-5/2}\|\delta\|^{2}$
by \eqref{eq:phibounds}, so
\[
  h_\rho(\theta)-\bar P(\theta)\bar g(\theta)
  =\E\big[D\phi(\bar v(\theta))[\delta]\,g_\theta(\zeta_0)\big]+\E\big[R_2\,g_\theta(\zeta_0)\big].
\]
The second term has norm at most $\tfrac38\epsilon^{-5/2}G\,\E\|\delta\|^{2}\le\tfrac38\epsilon^{-5/2}GC_v\rho$
by \eqref{eq:vvar}.

The first term is also $O(\rho)$, but through mixing rather than \eqref{eq:vvar}: Cauchy--Schwarz
would give only $\E\|\delta\|=O(\sqrt\rho)$. The term is a covariance, and \eqref{eq:backward}
writes it as
\[
  \E\big[D\phi(\bar v(\theta))[\delta]\,g_\theta(\zeta_0)\big]
  =\rho\sum_{k\ge0}a^{k}\,\E\Big[D\phi(\bar v(\theta))\big[\widetilde F_\theta(\zeta_{-1-k})\big]\,g_\theta(\zeta_0)\Big],
\]
whose $k$th expectation has $j$th coordinate $-\tfrac12(\bar v_j(\theta)+\epsilon)^{-3/2}\E[\widetilde F_{\theta,j}(\zeta_{-1-k})g_{\theta,j}(\zeta_0)]$,
a covariance of two bounded functions of tuples $k+1$ steps apart, with the first function centered. By
\eqref{eq:Qcontract}, it is at most $\tfrac12\epsilon^{-3/2}G^{3}C_{\mathrm Q}\kappa^{k+1}$ in
absolute value. Since $a^{k}\le1$, the sum has norm at most
$\rho\cdot\tfrac12\epsilon^{-3/2}\sqrt d\,G^{3}C_{\mathrm Q}\kappa/(1-\kappa)$, with a coefficient free of
$\rho$. Both bounds are uniform on $\Bcal_R$.

\emph{Part (ii), the Jacobian gap.} The Jacobians also differ by $O(\rho)$, uniformly on $\Bcal_R$.
The integrand $\phi(v^{-}(\theta))g_\theta(\zeta_0)$ is
continuously differentiable in $\theta$ with derivative bounded uniformly in the path, since
$\nabla_\theta g_\theta=A(\zeta)$, $\|S(\theta)\|_{\op}\le C_S$, and $\|D\phi\|_{\op}\le\tfrac12\epsilon^{-3/2}$. Dominated convergence therefore permits differentiation under the expectation, and
\begin{gather*}
  \nabla h_\rho(\theta)=\E\big[\phi(v^{-})A(\zeta_0)\big]+\E\big[D\phi(v^{-})[S(\theta)\,\cdot\,]\,g_\theta(\zeta_0)\big],\\
  \nabla\big[\bar P(\theta)\bar g(\theta)\big]=\bar P(\theta)A+D\phi(\bar v(\theta))[\nabla\bar v(\theta)\,\cdot\,]\,\bar g(\theta),
\end{gather*}
the second summand of each being the matrix whose $j$th column is obtained by inserting the $j$th standard basis vector $e_j$ in
the slot marked by the dot. The first summands differ by
$\E[D\phi(\bar v(\theta))[\delta]A(\zeta_0)]+\E[R_2A(\zeta_0)]$, which the field-gap argument bounds
by $C\rho$ with $G$ replaced by $L_g$ in the second factor.

The second summands differ by $O(\rho)$ as well. Expand
$D\phi(v^{-})=D\phi(\bar v(\theta))+D^{2}\phi(\bar v(\theta))[\delta,\cdot]+R_3$ with
$\|R_3\|\le\tfrac12\|D^{3}\phi\|\|\delta\|^{2}$, write $S=\nabla\bar v(\theta)+\widetilde S$ and
$g_\theta(\zeta_0)=\bar g(\theta)+\tilde g_0$ with $\tilde g_0$ centered and bounded by $2G$, and
multiply out. The product of the three leading terms is the target
$D\phi(\bar v(\theta))[\nabla\bar v(\theta)\,\cdot\,]\,\bar g(\theta)$.

Every other term has expectation $O(\rho)$ uniformly on
$\Bcal_R$, for one of three reasons. First, it contains no remainder and exactly one of
the centered factors $\delta$, $\widetilde S$, and $\tilde g_0$, so its expectation vanishes. Second, it contains two such factors, so its expectation either is a covariance at geometrically decaying lags, bounded by
$C\rho$ as the field gap was, or has norm at most
$C(\E\|\delta\|^{2}\E\|\widetilde S\|_{\op}^{2})^{1/2}\le C\rho$ by Cauchy--Schwarz and \eqref{eq:vvar}. Third, it contains a remainder $R_3$ or three centered factors, and its expectation has norm at most $C\E\|\delta\|^{2}\le C\rho$ or
$C\,\E[\|\delta\|\|\widetilde S\|_{\op}]\le C\rho$. Collecting the constants gives
\eqref{eq:smallgain} with a $C$ depending only on $\epsilon$, $G$, $L_g$, $d$, and the mixing
constants, not on $\rho$.

\emph{Part (iii).} The $\rho\to0$ field vanishes only at $\thetast$, where its Jacobian is $J_0$. The
coordinatewise bound $0\le\bar v(\theta)\le G^{2}$ gives
$(G^2+\epsilon)^{-1/2}I\preceq\bar P(\theta)\preceq\epsilon^{-1/2}I$, so $\bar P(\theta)$ is invertible
and $\bar P(\theta)(A\theta-b)=0$ exactly when $A\theta=b$, that is, at $\thetast$, since $A$ is invertible
by Lemma~\ref{lem:Apd}. In the product rule displayed above, the second term carries the factor
$\bar g(\thetast)=0$, and what remains at $\thetast$ is $\bar P(\thetast)A=J_0$, which is \eqref{eq:J0}.
It is positive stable by Lemma~\ref{lem:posstable}, since $\bar P(\thetast)$ is symmetric positive
definite.

\emph{Part (iv).} With $h_0(\theta):=\bar P(\theta)(A\theta-b)$, the zeros of $h_\rho$ are the fixed
points of $T_\rho(\theta):=\theta-J_0^{-1}h_\rho(\theta)$. As $\nabla h_0$ is locally Lipschitz with
$\nabla h_0(\thetast)=J_0$, there is $r_1>0$ with $\Bcal(\thetast,r_1)\subset\Bcal_R$ and
$\|I-J_0^{-1}\nabla h_0\|_{\op}\le\tfrac14$ on it. By \eqref{eq:smallgain}, for $\rho$ small, $T_\rho$ is
a $\tfrac12$-contraction of $\Bcal(\thetast,r_1)$ into itself with
$\|T_\rho(\thetast)-\thetast\|\le\|J_0^{-1}\|_{\op}C\rho$, so Banach's theorem gives a unique zero
$\thetast_\rho$ there, with $\|\thetast_\rho-\thetast\|\le2\|J_0^{-1}\|_{\op}C\rho$. Then,
$J_\rho=\nabla h_0(\thetast_\rho)+O(\rho)=J_0+O(\rho)$, and a Lyapunov certificate for $J_0$
\citep[Thm.~2.2.1]{hornjohnson1991} remains one for $J_\rho$ when $\rho$ is small, so $J_\rho$ is positive stable.
\end{proof}

\subsection{Proof of Theorem~\ref{thm:momentum} and Corollary~\ref{cor:iid}}
\label{app:thmmomentum}

The proof expands the mean field at $\thetast$ in $\rho$, as \citet[Thm.~2.5(i),
extended version]{lauand2023} (see also \citealt{lauand2022}) and \citet{huo2026} expand the iterate's stationary mean in the step size.

Two bounds on the tuple chain drive the proof: centered third-order moments decay with the larger of the two gaps
between their times (F1), and the stationary accumulator satisfies $\E\|v-\bar v\|^p=O(\rho^{p/2})$ for $2\le p\le4$ (F2).
Write $f_a:=f(\zeta_a)$. For bounded centered $f$,
$\|\Qcal^{k}f\|_\infty\le C_{\mathrm Q}\kappa^{k}\|f\|_\infty$ by \eqref{eq:Qcontract}.

\emph{(F1) Third-order decay.} For bounded centered $f,g,h$ and times $a\le b\le c$,
\[
  \big|\E[f_ag_bh_c]\big|\le C_{\mathrm Q}\|f\|_\infty\|g\|_\infty\|h\|_\infty\,\kappa^{\max(b-a,\,c-b)} .
\]
Conditioning on $(\zeta_a,\zeta_b)$ gives $|\E[f_ag_bh_c]|=|\E[f_ag_b(\Qcal^{c-b}h)(\zeta_b)]|\le
C_{\mathrm Q}\kappa^{c-b}\|f\|_\infty\|g\|_\infty\|h\|_\infty$. Conditioning on $\zeta_a$ instead gives
$\E[g_bh_c\mid\zeta_a]-\E[g_bh_c]=(\Qcal^{b-a}q-\mu q)(\zeta_a)$ for the function $q:=g\,\Qcal^{c-b}h$, with
$\|q\|_\infty\le\|g\|_\infty\|h\|_\infty$, so, as $\E[f_a]=0$, $|\E[f_ag_bh_c]|\le C_{\mathrm Q}\kappa^{b-a}\|f\|_\infty\|g\|_\infty\|h\|_\infty$.
Take the smaller bound.

\emph{(F2) Moments of the stationary accumulator.} With $a:=1-\rho$ and
$\Delta:=v-\bar v=\rho\sum_{i\ge0}a^{i}Y_{-i}$, $Y_{-i}:=F(\zeta_{-i})-\bar v$ centered with coordinates bounded by $G^{2}$,
$\E\|\Delta\|^{p}\le C_pG^{2p}\rho^{p/2}$ for $2\le p\le4$, with $C_p$ depending on $p$, $d$, and the mixing
constants. The case $p=2$ follows from \eqref{eq:vvar} by stationarity. For $p=4$, since $C_{\mathrm Q}\ge2$, conditioning as in (F1) bounds the absolute value of a fourth moment
of bounded centered coordinates at ordered times with gaps $n_1,n_2,n_3$ by
$C_{\mathrm Q}^{2}G^{8}\min\{\kappa^{n_1},\kappa^{n_3},\kappa^{n_1+n_3}+\kappa^{n_2}\}\le C_{\mathrm Q}^{2}G^{8}\{\kappa^{n_1+n_3}+\kappa^{(n_1+n_2+n_3)/3}\}$.
The third bound comes from the two pairs' means and their covariance, and the inequality from $\min\{a_1,a_2,a_3+a_4\}\le a_3+(a_1a_2a_4)^{1/3}$. A quadruple whose first index is $i$ carries weight
$\rho^{4}a^{4i+3n_1+2n_2+n_3}$. Since $\sum_ia^{4i}\le\rho^{-1}$ and $\sum_{n_2}a^{2n_2}\le\rho^{-1}$, summing over at most $4!$
orderings gives $\E[\Delta_j^{4}]\le24C_{\mathrm Q}^{2}G^{8}\{\rho^{2}(1-\kappa)^{-2}+\rho^{3}(1-\kappa^{1/3})^{-3}\}$. Hence,
$\E\|\Delta\|^{4}\le C_4G^{8}\rho^{2}$ with $C_4:=48d^{2}C_{\mathrm Q}^{2}{(1-\kappa^{1/3})^{-3}}$ for $\rho\le1$, and
$\E\|\Delta\|^{3}\le C_3G^{6}\rho^{3/2}$, $C_3:=C_4^{3/4}$, used for the Taylor remainder below.

\begin{lemma}[First-order covariance at every lag]
\label{lem:offset}
Under Assumption~\ref{a:data}, for every lag $\tau\in\mathbb Z$, the number of steps by which
the accumulator precedes the residual,
\[
  H_\tau:=\E\big[\phi(v_{t-\tau})\,\varepsilon_t\big]
  =\rho\sum_{i\ge0}(1-\rho)^{i}c(\tau+i)+r_\tau,\qquad \|r_\tau\|\le C_L\rho^{3/2},
\]
with $v$ the stationary accumulator at frozen $\thetast$ and $C_L<\infty$ depending only on $\epsilon$,
$G$, $d$, $C_{\mathrm Q}$, $\kappa$, and the constant $C_3$ of (F2), and free of $\tau$. A negative $\tau$ means that the accumulator comes after the residual.
\end{lemma}

\begin{proof}
The first-order Taylor term of $\phi$ gives the main term exactly, and the higher terms contribute
$O(\rho^{3/2})$. Stationarity gives the backward representation
$v_{t-\tau}-\bar v=\rho\sum_{i\ge0}(1-\rho)^{i}(F(\zeta_{t-\tau-i})-\bar v)$ for every
$\tau\in\mathbb Z$. The representation is the same whether or not the
accumulator includes $\varepsilon_t^{\odot2}$, so one argument covers the current and lagged updates, and only the lags
that appear differ. Set $\Delta:=v_{t-\tau}-\bar v$. Taylor's theorem at $\bar v$ with
integral remainder and \eqref{eq:phibounds} give
\[
  \phi(v_{t-\tau})=\phi(\bar v)+D\phi(\bar v)[\Delta]+\tfrac12D^{2}\phi(\bar v)[\Delta,\Delta]+R_3,
  \qquad \|R_3\|_{\op}\le\tfrac{5}{16}\epsilon^{-7/2}\|\Delta\|^{3}.
\]
Multiply by $\varepsilon_t$ and take expectations. The zeroth-order term vanishes because
$\varepsilon$ is centered. The first-order term is
$\rho\sum_i(1-\rho)^{i}\E[D\phi(\bar v)[F(\zeta_{t-\tau-i})-\bar v]\varepsilon_t]=\rho\sum_i(1-\rho)^{i}c(\tau+i)$,
absolutely convergent since $\|c(\tau)\|\le\tfrac12\epsilon^{-3/2}\sqrt d\,G^{3}C_{\mathrm Q}\kappa^{|\tau|}$
by \eqref{eq:Qcontract}.

The second-order term is $O(\rho^{2})$ uniformly in $\tau$. It equals
$\tfrac12\rho^{2}\sum_{i,i'}(1-\rho)^{i+i'}\E[D^{2}\phi(\bar v)[F_{t-\tau-i}-\bar v,F_{t-\tau-i'}-\bar v]\varepsilon_t]$,
each summand a vector of third-order moments of centered bounded functions at the times $t-\tau-i$, $t-\tau-i'$,
and $t$, with norm at most $\tfrac34\epsilon^{-5/2}\sqrt d\,G^{5}C_{\mathrm Q}\kappa^{n/2}$ by (F1) when their
spread is $n$. When $i=i'$, the two accumulator factors share a time, which (F1) allows. At most $6(n+1)$ pairs $(i,i')$ realize a given spread, so the double sum is at most a constant
times $\sum_n(n+1)\kappa^{n/2}<\infty$, uniformly in $\tau$. The $R_3$ term has norm at most
$\tfrac{5}{16}\epsilon^{-7/2}G\,\E\|\Delta\|^{3}\le\tfrac{5}{16}\epsilon^{-7/2}G\,C_3G^{6}\rho^{3/2}$
by (F2). Collecting the two remainders gives $r_\tau$ with a constant $C_L$ of the stated dependence.
\end{proof}

\begin{proof}[Proof of Theorem~\ref{thm:momentum}]
\emph{Field at $\thetast$.} The field at $\thetast$ is $\rho$ times a kernel-weighted sum of the
cross-cumulants $c(\tau)$, up to $O(\rho^{3/2})$ uniformly in $\lambda$. Under the invariant law at $\thetast$ (Proposition~\ref{prop:controlledmom}(i),
Step~1 of its proof), the momentum is $u_t=(1-\lambda)\sum_{k\ge0}\lambda^{k}\varepsilon_{t-k}$, an absolutely
convergent series with norm at most $G$, so expectation and sum interchange:
$h^{\lambda}_{\rho,m}(\thetast)=\E[\phi(v_{t-m})u_t]=(1-\lambda)\sum_{k\ge0}\lambda^{k}\E[\phi(v_{t-m})\varepsilon_{t-k}]$.
By stationarity, $\E[\phi(v_{t-m})\varepsilon_{t-k}]=\E[\phi(v_{t-m+k})\varepsilon_t]=H_{m-k}$, so
$h^{\lambda}_{\rho,m}(\thetast)=(1-\lambda)\sum_{k\ge0}\lambda^{k}H_{m-k}$. Inserting
Lemma~\ref{lem:offset} and using $(1-\lambda)\sum_k\lambda^{k}=1$ gives
\[
  h^{\lambda}_{\rho,m}(\thetast)=\rho\sum_{k\ge0}(1-\lambda)\lambda^{k}\sum_{i\ge0}(1-\rho)^{i}c(m-k+i)+r,
  \qquad \|r\|\le C_L\rho^{3/2},
\]
uniformly in $\lambda$. The double sum converges absolutely, so it can be reordered. The
coefficient of $\rho\,c(\tau)$ is $(1-\lambda)\sum_{k\ge0}\lambda^{k}(1-\rho)^{\tau-m+k}\ind\{\tau-m+k\ge0\}$,
which is \eqref{eq:kernel}. Summing the geometric series gives its closed form.

\emph{Bound on the coefficient.} The coefficient $B^{\lambda}_m$ decays geometrically in $m$, with a
constant that stays bounded as $\lambda\to1$. The bound follows from
$\|c(\tau)\|\le C\kappa^{|\tau|}$: the $m+1$ terms with $0\le\tau\le m$
each have norm at most $C\max(\kappa,\lambda)^{m}$. The norms of the terms with
$\tau>m$ sum to at most $C\kappa^{m+1}/(1-\kappa)$. The terms with $\tau<0$ have norms at most
$C\lambda^{m}(\lambda\kappa)^{|\tau|}$, a geometric series with ratio
$\lambda\kappa\le\kappa$, summing to at most $C\lambda^{m}\kappa/(1-\kappa)$. Hence,
$\|B^{\lambda}_m\|\le C(1+m)\max(\kappa,\lambda)^{m}/(1-\kappa)$, where $1/(1-\kappa)$ is a mixing
constant.

\emph{Limit of the kernel.} At fixed $\lambda$, $B_{\rho,\lambda,m}$ is within $O(\rho)$ of
$B^{\lambda}_m$. The inequality $(1-\rho)^{n}\ge1-n\rho$ gives
$|w_{\rho,\lambda,m}(\tau)-\lambda^{(m-\tau)_+}|\le\lambda^{(m-\tau)_+}\{(\tau-m)_+\rho+\lambda\rho/(1-\lambda)\}$, so
$\|B_{\rho,\lambda,m}-B^{\lambda}_m\|\le C\rho\{1+\lambda/(1-\lambda)\}\sum_\tau(1+|\tau|)\kappa^{|\tau|}$.
The factor multiplying $\rho$ in this bound grows as $\lambda\to1$.

\emph{Small-gain bound, uniform in $\lambda$.} The momentum field is within $O(\rho)$ of the limiting
field, with a constant free of $\lambda$. For $\theta\in\Bcal_R$, write
$g_\theta(\zeta):=g(\theta,\zeta)$, $F_\theta:=g_\theta^{\odot2}$, $\bar g(\theta)=A\theta-b$,
$\bar v(\theta)=\E_\mu F_\theta$, $\bar P(\theta)=\phi(\bar v(\theta))$, and $\Delta:=v_{t-m}-\bar v(\theta)$, with $v$ the stationary accumulator at frozen $\theta$. Then,
\begin{align*}
  h^{\lambda}_{\rho,m}(\theta)-\bar P(\theta)\bar g(\theta)
  &=(1-\lambda)\sum_k\lambda^{k}\Big\{\E\big[(\phi(v_{t-m})-\phi(\bar v(\theta)))(g_\theta(\zeta_{t-k})-\bar g(\theta))\big]\\
  &\qquad+\E\big[\phi(v_{t-m})-\phi(\bar v(\theta))\big]\bar g(\theta)\Big\}.
\end{align*}
The second expectation is $\E[D\phi(\bar v(\theta))[\Delta]]+O(\E\|\Delta\|^{2})=0+O(\rho)$ by \eqref{eq:vvar}.
The first has linear part $\rho\sum_i(1-\rho)^{i}\E[D\phi(\bar v(\theta))[\widetilde F_\theta(\zeta_{t-m-i})](g_\theta(\zeta_{t-k})-\bar g(\theta))]$,
and $\sum_i(1-\rho)^{i}\kappa^{|m+i-k|}\le2/(1-\kappa)$ uniformly in $k$, so it is $O(\rho)$.
Its Taylor remainder is $O(\E\|\Delta\|^{2})=O(\rho)$. Every term is thus $O(\rho)$ uniformly in $k$,
and so is the average with weights $(1-\lambda)\lambda^{k}$, uniformly in $\lambda$.

The Jacobian gap is also $O(\rho)$ uniformly in $\lambda$, once the accumulator's sensitivity is
centered. With $g_k:=g_\theta(\zeta_{t-k})$,
$S:=\nabla_\theta v_{t-m}=\rho\sum_i(1-\rho)^{i}2\Diag(g_\theta(\zeta_{t-m-i}))A(\zeta_{t-m-i})$, and $\E[S]=\nabla\bar v(\theta)$, the unweighted $k$th
term of $\nabla h^{\lambda}_{\rho,m}-\nabla(\bar P\bar g)$ is
$\E[(\phi(v_{t-m})-\phi(\bar v(\theta)))A(\zeta_{t-k})]+\E[(D\phi(v_{t-m})-D\phi(\bar v(\theta)))[S\,\cdot\,]g_k]+\E[D\phi(\bar v(\theta))[(S-\nabla\bar v(\theta))\,\cdot\,]g_k]$.
The first and third are $O(\rho)$ by mixing, $\widetilde S:=S-\E[S]$ being a centered $\rho$-weighted sum of
variance $O(\rho)$. In the second, writing $S=\nabla\bar v(\theta)+\widetilde S$ splits the linear part into a term
linear in $\Cov(\Delta,g_k)=O(\rho)$ and a term of norm at most
$C(\E\|\Delta\|^{2}\,\E\|\widetilde S\|_{\op}^{2})^{1/2}=O(\rho)$. The remainder is $O(\E\|\Delta\|^{2})$. Without
this centering, the bound would be only $O(\rho^{1/2})$. This extends Proposition~\ref{prop:smallgain}(ii) to the momentum field. The $\rho\to0$ field is $\bar P(\theta)\bar g(\theta)$, whose Jacobian at $\thetast$ is $J_0=\phi(\bar v)A$, free of $\lambda$.

\emph{Field to target.} Applying $-J_0^{-1}$ to the field at $\thetast$ gives the
leading-order displacement in \eqref{eq:momentumfirst}. The proof of Proposition~\ref{prop:smallgain}(iii)--(iv) rests only on the $\rho\to0$ field and
the small-gain bound, so it applies to $h^{\lambda}_{\rho,m}$ with a constant free of $\lambda$: for $\rho<\rho_0$,
there is a zero $\thetast_{\rho,m,\lambda}$, unique in a neighborhood of $\thetast$, with
$\|\thetast_{\rho,m,\lambda}-\thetast\|=O(\rho)$. The integral-form mean value theorem, applied
coordinatewise to $0=h^{\lambda}_{\rho,m}(\thetast_{\rho,m,\lambda})$, gives
$\thetast_{\rho,m,\lambda}-\thetast=-\bar J^{-1}h^{\lambda}_{\rho,m}(\thetast)$ with
$\bar J:=\int_0^1\nabla h^{\lambda}_{\rho,m}(\thetast+u(\thetast_{\rho,m,\lambda}-\thetast))\,du$. By the
small-gain bound on the Jacobian and the Lipschitz continuity of $\nabla[\bar P(\theta)\bar g(\theta)]$ near
$\thetast$, $\bar J=J_0+O(\rho)$. Since $J_0$ is invertible by Lemma~\ref{lem:posstable}, after a further
reduction of $\rho_0$, $\bar J^{-1}=J_0^{-1}+O(\rho)$. The field at $\thetast$ is
$\rho B_{\rho,\lambda,m}+O(\rho^{3/2})$ with $\|B_{\rho,\lambda,m}\|\le\sum_\tau\|c(\tau)\|$, since the kernel
\eqref{eq:kernel} is at most one, and \eqref{eq:momentumfirst} follows with $\|r\|\le C\rho^{3/2}$. Since
$C_L$ and the small-gain constant do not depend on $\lambda$, neither does $C$.
\end{proof}

\begin{proof}[Proof of Corollary~\ref{cor:iid}]
With $\Qcal(\zeta,\cdot)=\mu_\zeta$, \eqref{eq:Qcontract} holds with $C_{\mathrm Q}=2$ for every $\kappa\in(0,1)$, and the proofs above use no other property of the chain.
Independence and the centering of both factors give $c(\tau)=0$ for $\tau\ne0$, so $B_{\rho,\lambda,m}=w_{\rho,\lambda,m}(0)c(0)$, where
$w_{\rho,\lambda,m}(0)=0$ if and only if $\lambda=0$ and $m\ge1$, and $c(0)_j\propto\E_\mu[\varepsilon_j^{3}]$ by \eqref{eq:cumulant}. The lagged case without momentum is consistent with \citet[Thm.~5]{zhou2019}.
\end{proof}

\subsection{The second-order coefficient, and the proof of Theorem~\ref{thm:expansion}}
\label{app:expansion}

Write $h_0(\theta):=\bar P(\theta)(A\theta-b)$ for the $\rho\to0$ field, whose Jacobian at $\thetast$
is $J_0$, and $h_1:=B^{\lambda}_m$, so that $b_1=-J_0^{-1}h_1$. Expectations are taken under the
stationary law at frozen $\thetast$. Set $\widetilde F:=F-\bar v$, $\widetilde F_u:=\widetilde F(\zeta_u)$, and
$\Delta_{t-\tau}:=v_{t-\tau}-\bar v=\rho\sum_{i\ge0}a^{i}\widetilde F(\zeta_{t-\tau-i})$ with
$a:=1-\rho$, as in (F2). Each diagonal matrix $D^{k}\phi(\bar v)[\cdot]$ is identified with
the vector of its diagonal entries. Five objects enter $b_2$.

\emph{The accumulator's variance.} Summing $\rho^{2}\sum_{i,i'}a^{i+i'}\Cov(F(\zeta_{-i}),F(\zeta_{-i'}))$ along the
diagonals ${i-i'=\tau}$ gives the exact stationary covariance
$\E[\Delta\Delta^{\top}]=\rho(2-\rho)^{-1}\sum_{\tau\in\mathbb Z}a^{|\tau|}\Cov(F(\zeta_0),F(\zeta_\tau))$.
The lag-$\tau$ covariance is $O(\kappa^{|\tau|})$, so
\begin{equation}
\label{eq:Sigmav}
  \E[\Delta\Delta^{\top}]=\rho\,\Sigma_v+O(\rho^{2}),
  \qquad
  \Sigma_v:=\tfrac12\sum_{\tau\in\mathbb Z}\Cov\big(F(\zeta_0),F(\zeta_\tau)\big),
\end{equation}
half the long-run covariance of the squared residual. Only its diagonal entries
$\sigma^{2}_{v,j}:=(\Sigma_v)_{jj}$ enter below, because $\phi$ acts entrywise.

\emph{The second derivative of $h_0$.} The matrix $\nabla^{2}h_0(\thetast)[b_1]$ is
$q\mapsto D\phi(\bar v)[\nabla\bar v\cdot b_1]Aq+D\phi(\bar v)[\nabla\bar v\cdot q]Ab_1$ with
$\nabla\bar v:=\nabla\bar v(\thetast)$ and $\nabla\bar v\cdot q=\E_\mu[2\varepsilon\odot(A(\zeta)q)]$. The remaining terms vanish
because $A\thetast-b=0$.

\emph{The raw cross-cumulant.} Write
$\tilde c(\tau):=\E[\widetilde F(\zeta_0)\odot\varepsilon(\zeta_\tau)]$ for the raw cross-cumulant
function, so that $c(\tau)=D\phi(\bar v)[\tilde c(\tau)]$ coordinatewise. The resummed form of $\tilde c$ is
$\widetilde B^{\lambda}_m:=\sum_\tau\lambda^{(m-\tau)_+}\tilde c(\tau)$.

\emph{The first-order correction to the Jacobian.}
\begin{equation}
\label{eq:K1}
\begin{aligned}
  K_1&=\Phi_2A+\sum_{k\ge0}(1-\lambda)\lambda^{k}\sum_{i\ge0}\big(e(m-k+i)+e_S(m-k+i)\big)+\Psi_S,\\
  \Phi_2&:=\tfrac12D^{2}\phi(\bar v)[\Sigma_v],
  \qquad
  \Psi_S:=D^{2}\phi(\bar v)\big[\widetilde B^{\lambda}_m,\nabla\bar v\,\cdot\,\big],
\end{aligned}
\end{equation}
where $\Phi_2$ is the diagonal matrix with entries $\tfrac38(\bar v_j+\epsilon)^{-5/2}\sigma^{2}_{v,j}$.
The mean preconditioner satisfies ${\E[\phi(v)]-\phi(\bar v)=\rho\Phi_2+O(\rho^{2})}$.
The double sum involves $e(\tau):=\E[D\phi(\bar v)[\widetilde F(\zeta_0)](A(\zeta_\tau)-A)]$ and the scaled covariance
$e_S(\tau)_{\cdot j}:=\E[D\phi(\bar v)[2\varepsilon(\zeta_0)\odot A(\zeta_0)e_j]\,\varepsilon(\zeta_\tau)]$
of the squared residual's sensitivity to $\theta$ with the residual. Finally, $\Psi_S$,
with entries $(\Psi_S)_{ij}=\tfrac34(\bar v_i+\epsilon)^{-5/2}(\widetilde B^{\lambda}_m)_i\,\partial_{\theta_j}\bar v_i$,
pairs the mean sensitivity $\nabla\bar v$ with the resummed cross-cumulant through the curvature
of $\phi$. It contributes at first order even though it involves no fluctuation of the
sensitivity, because the mean sensitivity is of order one.

\emph{The second-order coefficient of the field.}
\begin{equation}
\label{eq:h2}
\begin{aligned}
  h_2&:=h_2^{\mathrm{ker}}+h_2^{(2)}+h_2^{(3)},\qquad
  h_2^{\mathrm{ker}}:=-\sum_\tau\lambda^{(m-\tau)_+}(\tau-m)_+\,c(\tau),\\
  h_2^{(2)}&:=\tfrac12\sum_{k\ge0}(1-\lambda)\lambda^{k}\sum_{i,i'\ge0}
    \E\big[D^{2}\phi(\bar v)[\widetilde F(\zeta_{-(m-k)-i}),\widetilde F(\zeta_{-(m-k)-i'})]\,\varepsilon(\zeta_0)\big],\\
  h_2^{(3)}&:=\tfrac12D^{3}\phi(\bar v)\big[\Sigma_v,\widetilde B^{\lambda}_m\big],
  \qquad\text{that is,}\qquad
  (h_2^{(3)})_j=-\tfrac{15}{16}(\bar v_j+\epsilon)^{-7/2}\,\sigma^{2}_{v,j}\,(\widetilde B^{\lambda}_m)_j .
\end{aligned}
\end{equation}
The first sums $c(\tau)$ against the derivative in $s$ at $s=0$ of the lag weights $\rho w_{\rho,\lambda,m}/s$ of \eqref{eq:kernelfactor}. The second sums scaled third cumulants of two squared-residual deviations and the residual over lags. The third pairs the accumulator's variance with the resummed cross-cumulant. A second-order Taylor
expansion would miss it, since it comes from the \emph{third} derivative of $\phi$ through $\E[\Delta_j^{3}\varepsilon_j]=3\E[\Delta_j^{2}]\E[\Delta_j\varepsilon_j]+O(\rho^{3})$.
Then,
\begin{equation}
\label{eq:b2}
  b_2=J_0^{-1}\big(\tfrac12\nabla^{2}h_0(\thetast)[b_1]+K_1\big)J_0^{-1}h_1-J_0^{-1}h_2 .
\end{equation}

Every term is built from expectations of bounded functions of the tuple chain at finitely many times,
summed over lags against polynomial or geometric weights. For a finite-state chain, resolvents of the state kernel give these sums in closed form, as for $B_{\mathrm{mem}}$. Neither $b_1$ nor $b_2$ depends on $\rho$, as Corollary~\ref{cor:rr}
requires.

A numerical check, independent of the simulation code, confirms \eqref{eq:b2} on two-state, one-feature TD instances (so $b_2$ is a scalar) with independent or persistent states, at $(m,\lambda)=(0,0)$, $(1,0)$, and $(1,0.9)$. The roots
of the frozen field are computed in 80-digit arithmetic at effective gains $s$ from $10^{-2}$ to $3.9\times10^{-5}$.
The second-order coefficient extrapolated from these roots agrees with \eqref{eq:b2} to within $10^{-10}$. Omitting $\Psi_S$ would change $b_2$ by $0.27$ to $14.6$ in absolute value and by $0.7$ to $72$ times $|b_2|$, where $\Psi_S\ne0$.

\subsubsection*{Cumulant decay and a counting rule}

For scalar random variables $Y_1,\dots,Y_k$ with finite moments, the joint cumulant is
$\operatorname{cum}(Y_1,\dots,Y_k):=\sum_{\pi}(-1)^{|\pi|-1}(|\pi|-1)!\prod_{B\in\pi}\E[\prod_{i\in B}Y_i]$,
the sum over partitions $\pi$ of $\{1,\dots,k\}$, and the moment--cumulant formula inverts it:
$\E[Y_1\cdots Y_k]=\sum_\pi\prod_{B\in\pi}{\operatorname{cum}(Y_i:i\in B)}$. Cumulants are multilinear, the
first-order cumulant is the mean, and every partition containing a singleton block of a
centered variable contributes zero to the formula. Write $\kappa$ and $C_{\mathrm Q}$ as in
\eqref{eq:Qcontract}. Cumulant decay bounds are classical for mixing sequences (see \citealt[Sec.~1.4]{doring2022} and the references there). The proof below keeps the constant explicit for bounded functions of a uniformly ergodic chain.

\begin{lemma}[Cumulant decay for the tuple chain]
\label{lem:cumdecay}
Let $f_1,\dots,f_k$ be bounded functions on the tuple space and $t_1\le\dots\le t_k$. Then,
\[
  \big|\operatorname{cum}\big(f_1(\zeta_{t_1}),\dots,f_k(\zeta_{t_k})\big)\big|
  \le C^{\mathrm{cum}}_k\,\prod_{i\le k}\|f_i\|_\infty\;\kappa^{\,g},
  \qquad g:=\max_{i<k}(t_{i+1}-t_i),
\]
with $C^{\mathrm{cum}}_k$ depending on $k$ and $C_{\mathrm Q}$ only ($g:=0$ when $k=1$).
\end{lemma}

\begin{proof}
We argue by induction on $k$: the cumulant is a covariance across the largest gap minus products
over straddling partitions, and both decay across that gap. For $k=1$, $|\E f_1|\le\|f_1\|_\infty$ gives $C^{\mathrm{cum}}_1=1$. For $k\ge2$, let $j$ attain the largest gap,
$L:=\{1,\dots,j\}$, $R:=\{j+1,\dots,k\}$, and write $Y_i:=f_i(\zeta_{t_i})$, $Y_S:=\prod_{i\in S}Y_i$, and $\operatorname{cum}_B:=\operatorname{cum}(Y_i:i\in B)$.

The covariance of $Y_L$ and $Y_R$ decays as $\kappa^{g}$. With $\Psi(z):=\E[Y_R\mid\zeta_{t_{j+1}}=z]$, bounded by
$\prod_R\|f_i\|_\infty$, the Markov property and stationarity give
$\E[Y_LY_R]=\E[Y_L(\Qcal^{g}\Psi)(\zeta_{t_j})]$ and $\E[Y_R]=\mu\Psi$, so by \eqref{eq:Qcontract},
\[
  \big|\E[Y_LY_R]-\E[Y_L]\E[Y_R]\big|=\big|\E[Y_L(\Qcal^{g}\Psi-\mu\Psi)(\zeta_{t_j})]\big|
  \le C_{\mathrm Q}\kappa^{g}\prod_{i\le k}\|f_i\|_\infty .
\]

The cumulant differs from that covariance by products over straddling partitions, each decaying as
$\kappa^{g}$ by induction. The moment--cumulant formula on $L$ and on $R$ separately gives
$\E[Y_L]\E[Y_R]=\sum_{\pi}\prod_{B\in\pi}\operatorname{cum}_B$ over the partitions $\pi$ of
$\{1,\dots,k\}$ none of whose blocks meets both $L$ and $R$. Subtracting this from the formula for
$\E[Y_1\cdots Y_k]$ and isolating the one-block partition gives
\[
  \operatorname{cum}(Y_1,\dots,Y_k)=\big\{\E[Y_LY_R]-\E[Y_L]\E[Y_R]\big\}
   -\sum_{\pi\ \text{straddling},\ |\pi|\ge2}\ \prod_{B\in\pi}\operatorname{cum}_B,
\]
where a straddling partition has at least one block $B$ meeting both $L$ and $R$. Since $|\pi|\ge2$, such a block
has $|B|<k$ and, between its largest member in $L$ and its smallest in $R$, a time gap of at least
$g$, so by the induction hypothesis, $|\operatorname{cum}_B|\le C^{\mathrm{cum}}_{|B|}\prod_B\|f_i\|_\infty\kappa^{g}$,
while every other block is bounded by the hypothesis with its decay factor replaced by one. The sum has fewer
terms than the Bell number $\mathrm{Bell}_k$, so the claim holds with
$C^{\mathrm{cum}}_k:=C_{\mathrm Q}+\mathrm{Bell}_k\,(\max_{b<k}C^{\mathrm{cum}}_b)^{k}$.
\end{proof}

\begin{lemma}[Counting rule]
\label{lem:counting}
Let $Y_r:=\rho\sum_{i\ge0}a^{i}\widetilde\psi_r(\zeta_{\bar t_r-i})$, $r=1,\dots,k$, be weighted
backward sums of centered functions bounded by $\Lambda$, with arbitrary anchors
$\bar t_r\in\mathbb Z$, and let $Z_1,\dots,Z_{k'}$ be centered functions of the tuple at fixed times, bounded by
$\Lambda$. In the moment--cumulant expansion of $\E[Y_1\cdots Y_kZ_1\cdots Z_{k'}]$, a partition
$\pi$ with no singleton blocks contributes at most $C\rho^{\,k-k_0(\pi)}$ in modulus, where $k_0(\pi)$ counts the
blocks of $\pi$ with no $Z$, and $C$ depends on $k$, $k'$, $\Lambda$, $C_{\mathrm Q}$,
and $\kappa$, not on $\rho$, the anchors, or fixed times. Thus,
\[
  \E[Y_1\cdots Y_kZ_1\cdots Z_{k'}]=O\big(\rho^{\,k-\lfloor k/2\rfloor}\big)\qquad\text{for every }k'\ge0.
\]
If, moreover, every block of $\pi$ contains a $Z$, so that $k_0(\pi)=0$, then replacing every
weight $a^{i}$ by $1$ inside $\pi$'s contribution changes it by at most
$C'\rho^{\,k+1}\big(1+\max_{r,p}(\bar t_r-t_p)_+\big)$, where $t_1,\dots,t_{k'}$ are the fixed times and
$C'$ depends on the same quantities as $C$.
\end{lemma}

\begin{proof}
Each block without a $Z$ leaves one index sum bounded by $1/\rho$. Together with the prefactor $\rho^{k}$, this gives the bound $O(\rho^{k-k_0(\pi)})$. Expand each $Y_r$ and use multilinearity: the contribution of $\pi$ is
\[
  \rho^{k}\sum_{i_1,\dots,i_k\ge0}\prod_ra^{i_r}\prod_{B\in\pi}\operatorname{cum}(W_B),
\]
where $W_B$ lists the variables of block $B$ at their times, $\widetilde\psi_r(\zeta_{\bar t_r-i_r})$ or a
$Z$. Fix a block $B$ with $b:=|B|\ge2$ members at times $t'_r$. By Lemma~\ref{lem:cumdecay},
$|\operatorname{cum}(W_B)|\le C^{\mathrm{cum}}_b\Lambda^{b}\kappa^{g_B}$ with $g_B$ the largest gap
among these times, and since their spread is at most $(b-1)g_B$,
$\kappa^{g_B}\le\kappa^{\mathrm{spread}/(b-1)}\le\prod_{r}\kappa^{|t'_r-t'_{r_0}|/(b-1)^{2}}$
for any member $r_0$ of $B$, where the product runs over the other
members.

Only a block without a $Z$ leaves an index sum of order $1/\rho$. If
$B$ contains a $Z$, take one as $r_0$: its time is fixed, the decay factor of any other $Z$ is at most
one, and as $a^{i_r}\le1$, each weighted member's index sum, including its decay factor, is at most
$2/(1-\kappa^{1/(b-1)^{2}})$. If $B$ contains no $Z$, take any member as $r_0$: summing its index against
$a^{i_{r_0}}$ gives exactly $1/\rho$, and the other $b-1$
indices are summed as before. With the prefactor $\rho^{k}$ and exactly $k_0(\pi)$ factors $1/\rho$,
the contribution is $O(\rho^{k-k_0(\pi)})$, with a constant free of the anchors
and fixed times. Since each block counted by $k_0$ contains at least two $Y$-variables, we have $k_0\le\lfloor k/2\rfloor$. Summing over the finitely many partitions, each with at most $\lfloor(k+k')/2\rfloor$ blocks, gives the displayed moment bound.

When every block contains a $Z$, replacing the weights by one changes the contribution by at most $\rho^{k+1}$ times an index sum that grows at most linearly in the lags. Indeed, $0\le1-a^{i_1+\dots+i_k}\le\rho(i_1+\dots+i_k)$, so the change is at most
$\rho^{k+1}\sum_{i_1,\dots,i_k\ge0}(i_1+\dots+i_k)\prod_{B\in\pi}|\operatorname{cum}(W_B)|$, with
each weight now bounded by one. Each index
$i_r$ is summed as above against $\kappa_b^{|\bar t_r-i_r-t_p|}$, $\kappa_b:=\kappa^{1/(b-1)^{2}}$, with $t_p$ the time of
$r_0$ in its block, and
$\sum_{i\ge0}i\,\kappa_b^{|\bar t_r-i-t_p|}\le C''\big(1+(\bar t_r-t_p)_+\big)$: the terms with
$i\le(\bar t_r-t_p)_+$ contribute at most $(\bar t_r-t_p)_+\cdot2/(1-\kappa_b)$, and writing
$i=(\bar t_r-t_p)_++j$ for the rest gives at most $(\bar t_r-t_p)_+/(1-\kappa_b)+\sum_{j\ge1}j\kappa_b^{j}$. In the
term with factor $i_r$, the other indices are summed as before. This gives the stated bound.
\end{proof}

\begin{proof}[Proof of Theorem~\ref{thm:expansion}]
\emph{Step 1: the field at $\thetast$ to order $s^{2}$.} Only the Taylor terms of orders one to three
contribute to $H_\tau$ through order $\rho^{2}$. Fix $\tau\in\mathbb Z$ and expand
$\phi$ in $H_\tau=\E[\phi(v_{t-\tau})\varepsilon_t]$ by Taylor's theorem at $\bar v$ to fifth order in $\Delta:=v_{t-\tau}-\bar v$,
$\phi(\bar v+\Delta)=\sum_{n\le5}\tfrac1{n!}D^{n}\phi(\bar v)[\Delta^{\otimes n}]+R_6$ with
$\|R_6\|\le\tfrac1{720}\|D^{6}\phi\|_\infty\|\Delta\|^{6}$, $D^{6}\phi$ being bounded by
$\tfrac{10395}{64}\epsilon^{-13/2}$ entrywise. Since $\phi$ acts entrywise, with $\phi_j(x):=(x+\epsilon)^{-1/2}$, the $j$th coordinate
of the $n$th term in $H_\tau$ is $\tfrac1{n!}\phi_j^{(n)}(\bar v_j)$ times $\E[\Delta_j^{n}\varepsilon_j]$, a joint
moment of $n$ copies of the weighted sum $\Delta_j$ and the centered fixed-time factor
$\varepsilon_j(\zeta_t)$, to which Lemma~\ref{lem:counting} applies with $k=n$, $k'=1$. The
$n=0$ term vanishes. The $n=1$ term is computed exactly:
$\E[D\phi(\bar v)[\Delta]\varepsilon_t]=\rho\sum_ia^{i}c(\tau+i)$.

The $n=2$ term is $\rho^{2}$ times a double sum of third cumulants of two squared-residual deviations and the
residual, up to $O\big((1+\tau_-)\rho^{3}\big)$, $\tau_-:=\max(-\tau,0)$. The three
variables are centered, so only the one-block partition contributes. It contains the fixed-time
$\varepsilon_j(\zeta_t)$, so the replacement clause of the lemma applies, with anchors
$\bar t_r=t-\tau$ and fixed time $t$, and replacing $a^{i+i'}$ by $1$ changes the term by $O\big((1+\tau_-)\rho^{3}\big)$:
the term is $\rho^{2}\cdot\tfrac12\sum_{i,i'\ge0}\E[D^{2}\phi(\bar v)[\widetilde F_{-\tau-i},\widetilde F_{-\tau-i'}]\varepsilon_0]+O\big((1+\tau_-)\rho^{3}\big)$.

The $n=3$ term pairs the accumulator's variance with the raw cross-cumulant, at order
$\rho^{2}$. The partitions of four centered variables without singletons are the one block,
contributing $O(\rho^{3})$, and the three pairings, contributing equally because the three copies of $\Delta_j$
are identical: $\E[\Delta_j^{3}\varepsilon_j]=3\E[\Delta_j^{2}]\E[\Delta_j\varepsilon_j]+O(\rho^{3})$,
and by \eqref{eq:Sigmav} and the exact $n=1$ computation,
$\E[\Delta_j^{2}]\E[\Delta_j\varepsilon_j]=\rho^{2}\sigma^{2}_{v,j}\sum_i\tilde c_j(\tau+i)+O\big((1+\tau_-)\rho^{3}\big)$,
the lag factor again coming from replacing $a^{i}$ by one in $\sum_ia^{i}\tilde c_j(\tau+i)$.
With $\tfrac1{3!}\cdot3\cdot\phi_j'''(\bar v_j)=-\tfrac{15}{16}(\bar v_j+\epsilon)^{-7/2}$, this
term becomes $\rho^{2}h_2^{(3)}$ after the momentum mixture.

The terms with $n=4$ and $n=5$, and the remainder term, are $O(\rho^{3})$. For $n=4$ and $n=5$, every partition of five or six
variables without singletons has $k_0\le1$ and $k_0\le2$, respectively, so the terms are
$O(\rho^{3})$, and $\E\|\Delta\|^{6}=O(\rho^{3})$ by Lemma~\ref{lem:counting} with $k=6$, $k'=0$, coordinatewise. Hence, with a remainder bounded by
$C(1+\tau_-)\rho^{3}$ for a constant $C$ free of $\tau$,
\begin{equation}
\label{eq:Hell2}
\begin{aligned}
  H_\tau={}&\rho\sum_{i\ge0}a^{i}c(\tau+i)
  +\rho^{2}\Big\{\tfrac12\sum_{i,i'\ge0}\E\big[D^{2}\phi(\bar v)[\widetilde F_{-\tau-i},\widetilde F_{-\tau-i'}]\varepsilon_0\big]
   +\tfrac12D^{3}\phi(\bar v)\big[\Sigma_v,\textstyle\sum_{i\ge0}\tilde c(\tau+i)\big]\Big\}\\
  &+O\big((1+\tau_-)\rho^{3}\big).
\end{aligned}
\end{equation}

The momentum mixture of \eqref{eq:Hell2} keeps the remainder at $O(\rho^{3})$, with a constant
that carries the factor $1/(1-\lambda)$. At $\thetast$, the momentum field is the mixture $h^{\lambda}_{\rho,m}(\thetast)=(1-\lambda)\sum_k\lambda^{k}H_{m-k}$
(proof of Theorem~\ref{thm:momentum}). The remainder in \eqref{eq:Hell2} grows at most linearly in the lag, so the mixture of the remainders is at most
$C\rho^{3}(1-\lambda)\sum_k\lambda^{k}(1+(k-m)_+)\le C\rho^{3}\{1+\lambda/(1-\lambda)\}$. The first-order terms resum to
$\rho\sum_\tau w_{\rho,\lambda,m}(\tau)c(\tau)=s\sum_\tau\lambda^{(m-\tau)_+}a^{(\tau-m)_+}c(\tau)$ by \eqref{eq:kernelfactor}, and the $D^{3}\phi$ terms, linear in
$\sum_i\tilde c(\tau+i)$, resum to $\rho^{2}\cdot\tfrac12D^{3}\phi(\bar v)[\Sigma_v,\widetilde B^{\lambda}_m]+O(\rho^{3})$, since $(1-\lambda)\sum_k\lambda^{k}\sum_i\tilde c(m-k+i)=\widetilde B^{\lambda}_m$.

The field at $\thetast$ expands in the effective gain as $sh_1+s^{2}h_2+O(s^{3})$. From \eqref{eq:seff}, $\rho=s(1-\lambda)/(1-\lambda-s\lambda)=s+O(s^{2})$
with a constant depending on $\lambda$, and $a^{(\tau-m)_+}=1-\rho(\tau-m)_++O(\rho^{2}(\tau-m)^{2})$.
Both corrections are summable against $c(\tau)=O(\kappa^{|\tau|})$, so
$s\sum_\tau\lambda^{(m-\tau)_+}a^{(\tau-m)_+}c(\tau)=sh_1+s^{2}h_2^{\mathrm{ker}}+O(s^{3})$, and
$\rho^{2}=s^{2}+O(s^{3})$ in the remaining terms. Altogether,
\begin{equation}
\label{eq:fieldexp}
  h^{\lambda}_{\rho,m}(\thetast)=s\,h_1+s^{2}h_2+O(s^{3}),
\end{equation}
with $h_2$ as in \eqref{eq:h2}. The $\lambda$-dependence of $\rho(s)$ enters only at third order
because the first-order term is written with its exact kernel.

\emph{Step 2: the Jacobian at $\thetast$ to first order.} The Jacobian at $\thetast$ is
$J_0+sK_1+O(s^{2})$, assembled from three pieces. By Proposition~\ref{prop:controlled}(ii), for every $m\ge0$, the field is $C^{3}$ on $\Bcal_R$ with derivatives bounded uniformly
in $\rho$, and differentiating under the expectation gives
\[
\begin{aligned}
  \nabla h^{\lambda}_{\rho,m}(\theta)&=\E\big[\phi(v_{t-m})\nabla_\theta u_t\big]
   +\E\big[D\phi(v_{t-m})[S_{t-m}\,\cdot\,]\,u_t\big],\\
  S_{t-m}&:=\nabla_\theta v_{t-m}
   =\rho\sum_ia^{i}\,2\Diag(g_\theta(\zeta_{t-m-i}))A(\zeta_{t-m-i}),
\end{aligned}
\]
and at $\thetast$, $u_t=(1-\lambda)\sum_k\lambda^{k}\varepsilon_{t-k}$ and
$\nabla_\theta u_t=(1-\lambda)\sum_k\lambda^{k}A(\zeta_{t-k})$.

(a) The mean preconditioner contributes $\E[\phi(v_{t-m})]A=\phi(\bar v)A+\rho\Phi_2A+O(\rho^{2})$.
Expanding $\phi$ to third order in $\Delta:=v_{t-m}-\bar v$ gives $\E[D\phi(\bar v)[\Delta]]=0$,
$\tfrac12\E[D^{2}\phi(\bar v)[\Delta,\Delta]]=\tfrac12\rho D^{2}\phi(\bar v)[\Sigma_v]+O(\rho^{2})$
by \eqref{eq:Sigmav}, a cubic term of $O(\rho^{2})$ by Lemma~\ref{lem:counting} with $k=3$, $k'=0$, and a remainder of $O(\E\|\Delta\|^{4})=O(\rho^{2})$ by (F2).

(b) The covariances $(1-\lambda)\sum_k\lambda^{k}\E[\phi(v_{t-m})(A(\zeta_{t-k})-A)]$ give
$\rho\sum_k(1-\lambda)\lambda^{k}\sum_ie(m-k+i)+O(\rho^{2})$. For each $k$, the first-order
Taylor term is $\rho\sum_ia^{i}e(m-k+i)$, whose weights may be replaced by one at cost
$\rho\sum_i(1-a^{i})\|e(m-k+i)\|\le C\rho^{2}(1+(k-m)_+)$, the sum $\sum_ii\,\kappa^{|m-k+i|}$
growing linearly in the lag as in the lemma's replacement clause. The second- and
third-order terms are $O(\rho^{2})$, uniformly in $k$, by the lemma's partitionwise
bound with one fixed-time factor $A(\zeta_{t-k})-A$, as is the remainder by (F2). Since
$(1-\lambda)\sum_k\lambda^{k}(1+(k-m)_+)\le1+\lambda/(1-\lambda)$, the mixture's error is at most
$C\rho^{2}/(1-\lambda)$.

(c) The $j$th column $\E[D\phi(v_{t-m})[S_{t-m}e_j]\,u_t]$ of the term from the sensitivity contributes
$\rho\sum_k(1-\lambda)\lambda^{k}\sum_ie_S(m-k+i)_{\cdot j}+\rho\,(\Psi_Se_j)+O(\rho^{2})$. With $\psi_j(\zeta):=2\varepsilon(\zeta)\odot(A(\zeta)e_j)$,
whose mean is $\nabla\bar v\,e_j$, split the sensitivity as $S_{t-m}e_j=\nabla\bar v\,e_j+\widetilde S_j$
with $\widetilde S_j:=\rho\sum_ia^{i}\{\psi_j(\zeta_{t-m-i})-\nabla\bar v\,e_j\}$ centered, the weights
$\rho a^{i}$ summing to one. Expand
$D\phi(v_{t-m})$ about $\bar v$ to third order in $\Delta:=v_{t-m}-\bar v$, with a remainder of at most
$C\|\Delta\|^{4}$, and take expected products with each part.
\begin{itemize}[leftmargin=1.6em,itemsep=1pt]
\item $D\phi(\bar v)[\nabla\bar v\,e_j]$ against $u_t$ has mean zero, $u_t$ being centered at $\thetast$.
\item $D\phi(\bar v)[\widetilde S_j]$ against $u_t$ is
$\rho\sum_k(1-\lambda)\lambda^{k}\sum_ie_S(m-k+i)_{\cdot j}+O(\rho^{2})$, the weights being replaced
by one at the cost computed in (b).
\item The linear term against the mean part is
$D^{2}\phi(\bar v)\big[\E(\Delta\odot u_t),\nabla\bar v\,e_j\big]$. Here,
$\E(\Delta\odot u_t)=(1-\lambda)\sum_k\lambda^{k}\rho\sum_ia^{i}\tilde c(m-k+i)=\rho\widetilde B^{\lambda}_m+O(\rho^{2})$,
with the same replacement cost, so this product is $\rho\,(\Psi_Se_j)+O(\rho^{2})$.
\item Every other product is $O(\rho^{2})$. The linear term against $\widetilde S_j$, and the
quadratic and cubic terms against either part, are mixtures over $k$ of joint moments of at least two weighted sums
and the fixed-time $\varepsilon_{t-k}$, so by Lemma~\ref{lem:counting}, every partition without singletons
contributes $O(\rho^{2})$, uniformly in $k$. The Taylor remainder is at most $C\,\E\|\Delta\|^{4}=O(\rho^{2})$ by (F2).
\end{itemize}
Collecting (a)--(c),
\begin{equation*}
  \nabla h^{\lambda}_{\rho,m}(\thetast)=J_0+\rho K_1+O(\rho^{2})=J_0+sK_1+O(s^{2}),
\end{equation*}
with $K_1$ as in \eqref{eq:K1}, the second equality following from $\rho=s+O(s^{2})$.

\emph{Step 3: from field to target.} By Step~2, the field is
$C^{3}$ on $\Bcal_R$ with third derivative bounded uniformly in $\rho$, and the argument of
Proposition~\ref{prop:smallgain}(ii) applied to the integrand's second
$\theta$-derivative---a polynomial with bounded coefficients in $D^{n}\phi(v^{-})$, the affine $g_\theta$, and the first and second
sensitivities, whose centered parts have $O(\rho)$ second moments---gives $\nabla^{2}h^{\lambda}_{\rho,m}(\thetast)=\nabla^{2}h_0(\thetast)+O(\rho)$.
With $\delta:=\thetast_{\rho,m,\lambda}-\thetast$ and the average Jacobian
$\bar J:=\int_0^1\nabla h^{\lambda}_{\rho,m}(\thetast+\upsilon\delta)\,d\upsilon$, the integral-form mean value theorem gives
$\bar J\delta=-h^{\lambda}_{\rho,m}(\thetast)$. Two passes expand $\bar J$, invertible for small $s$, using
only $\|\delta\|=O(\rho)=O(s)$ from Theorem~\ref{thm:momentum} at fixed $\lambda$.

\emph{First pass.} The Jacobian of the field is Lipschitz on $\Bcal_R$ with a constant free of
$\rho$, so $\bar J=\nabla h^{\lambda}_{\rho,m}(\thetast)+O(\|\delta\|)=J_0+O(s)$ by Step~2. With
\eqref{eq:fieldexp}, this gives $\delta=-(J_0^{-1}+O(s))(sh_1+O(s^{2}))=sb_1+O(s^{2})$.

\emph{Second pass.} With the third derivative bounded,
\[
\begin{aligned}
  \bar J&=\nabla h^{\lambda}_{\rho,m}(\thetast)+\tfrac12\nabla^{2}h^{\lambda}_{\rho,m}(\thetast)[\delta]+O(\|\delta\|^{2})\\
  &=J_0+s\big(K_1+\tfrac12\nabla^{2}h_0(\thetast)[b_1]\big)+O(s^{2}),
\end{aligned}
\]
the second line using the first-pass estimate for $\delta$ and $\rho=s+O(s^{2})$ for $\rho K_1$. So
$\bar J^{-1}=J_0^{-1}-sJ_0^{-1}(K_1+\tfrac12\nabla^{2}h_0(\thetast)[b_1])J_0^{-1}+O(s^{2})$, and
inserting \eqref{eq:fieldexp} into $\delta=-\bar J^{-1}h^{\lambda}_{\rho,m}(\thetast)$ gives
\[
  \delta=-sJ_0^{-1}h_1+s^{2}\Big\{J_0^{-1}\big(K_1+\tfrac12\nabla^{2}h_0(\thetast)[b_1]\big)J_0^{-1}h_1-J_0^{-1}h_2\Big\}+O(s^{3}),
\]
which is \eqref{eq:expansion} with \eqref{eq:b2}. The $O(s^{3})$ constant collects those of
Lemma~\ref{lem:counting} to order six, the small-gain bounds, and the factors
$\lambda/(1-\lambda)$ from the momentum mixture in Step~1, from Step~2(b), and from $\rho(s)$,
which is why it depends on $\lambda$ while the constant of Theorem~\ref{thm:momentum} does not. It depends on $m$ via the lags, not on $\rho$.
\end{proof}

\subsection{Proof of Proposition~\ref{prop:knob}}
\label{app:knob}

\begin{proof}[Proof of Proposition~\ref{prop:knob}]
Write $h_{\min}>0$ for the smallest eigenvalue of $-H$ and $\varDelta:=\hat z_\rho-z^\star$. Because $V$
is $C^{2}$ with $\nabla V(z^\star)=0$, there is $r_0>0$ with $\Bcal(z^\star,r_0)\subset T$
on which $V(z^\star)-V(z)\ge\tfrac14h_{\min}\|z-z^\star\|^{2}$ and
$\|\nabla V(z)-H(z-z^\star)\|\le\tfrac12h_{\min}\|z-z^\star\|$. Take
$\mathcal O:=\Bcal(z^\star,r_0)$.

\emph{Existence.} For small $\rho$, $V_\rho$ has a critical point in $\mathcal O$. On the closed ball,
$\sup|V_\rho-V|\le\rho\sup|\delta|+\sup|r|=O(\rho)$, so for small $\rho$, every point of the sphere
$\|z-z^\star\|=r_0$ has $V_\rho(z)\le V(z^\star)-\tfrac14h_{\min}r_0^{2}+O(\rho)<V_\rho(z^\star)$,
and the maximum of $V_\rho$ over the closed ball is attained at an interior point, a local
maximizer and, as $V_\rho$ is $C^{1}$, a critical point. No uniqueness is claimed, as the $C^{1}$ bound does not control
$\nabla^{2}r$.

\emph{Location and loss.} Each such critical point lies $O(\rho)$ from $z^\star$, displaced
to first order along $H^{-1}\nabla\delta(z^\star)$, and loses LSTD value only at second order. At a critical point in $\mathcal O$, $0=\nabla V(\hat z_\rho)-\rho\nabla\delta(\hat z_\rho)+\nabla r(\hat z_\rho,\rho)$. As $\|H\varDelta\|\ge h_{\min}\|\varDelta\|$, the
choice of $r_0$ gives $\|\nabla V(\hat z_\rho)\|\ge\|H\varDelta\|-\tfrac12h_{\min}\|\varDelta\|\ge\tfrac12h_{\min}\|\varDelta\|$, while the last two terms are
$O(\rho)$, so $\|\varDelta\|=O(\rho)$. As $\nabla^{2}V$ is uniformly continuous and $\nabla^{2}\delta$ bounded on the closed ball, $\nabla V(\hat z_\rho)=H\varDelta+o(\rho)$,
$\nabla\delta(\hat z_\rho)=\nabla\delta(z^\star)+O(\rho)$, and $\nabla r=O(\rho^{3/2})$, so
$H\varDelta=\rho\nabla\delta(z^\star)+o(\rho)$, which is the first expansion. Expanding $V$ about $z^\star$ gives
$V(z^\star)-V(\hat z_\rho)=-\tfrac12\varDelta^\top H\varDelta+o(\|\varDelta\|^{2})$, which is the second once the
first is substituted.

\emph{Reported value.} The reported value at the critical point is off by $-\rho\,\delta(z^\star)$
at first order:
$V_\rho(\hat z_\rho)-V(z^\star)=\{V(\hat z_\rho)-V(z^\star)\}-\rho\,\delta(\hat z_\rho)+r(\hat z_\rho,\rho)=-\rho\,\delta(z^\star)+O(\rho^{3/2})$.
\end{proof}

\section{Proofs for the iterates: consistency and limit laws}
\label{app:imported}

This section proves consistency at the displaced target on the confinement event, and the limit laws conditionally on it, for the recursion
\eqref{eq:momentum} at any delay $m\ge0$ and momentum $\lambda\in[0,1)$. At $(m,\lambda)=(1,0)$, the recursion is \eqref{eq:update} with $P^{\mathrm{lag}}_t$. The
framework is stochastic approximation with controlled Markov noise
\citep{metivier1984,benveniste1990,andrieu2005}. The proofs use the regularity of Proposition~\ref{prop:controlled}
(Section~\ref{app:enlarged}). Throughout, $\rho\in(0,1)$, $\lambda$, and $m$ are fixed,
$h_\rho:=h^{\lambda}_{\rho,m}$ and $\thetast_\rho:=\thetast_{\rho,m,\lambda}$.

\subsection{Coercivity at the displaced target}

\begin{lemma}[Coercivity in a Lyapunov metric]
\label{lem:basin}
Let Assumption~\ref{a:data} hold. There is $\rho_0>0$, not depending on $\lambda$,
such that for every $\rho<\rho_0$, the field $h_\rho$ has a zero $\thetast_\rho$ that is locally unique near
$\thetast$, with a positive-stable Jacobian $J_\rho:=\nabla h_\rho(\thetast_\rho)$. There are also a symmetric
positive-definite $\Qly_\rho$ and a radius $\varrho_\rho>0$ with
$\Bcal(\thetast_\rho,\varrho_\rho)\subset\Bcal_R$ such that
\begin{equation}
\label{eq:deployedcoercive}
  (\theta-\thetast_{\rho})^{\top}\Qly_\rho\,h_{\rho}(\theta)\;\ge\;\tfrac14\|\theta-\thetast_{\rho}\|^{2}
  \qquad\text{for every }\theta\in\Bcal(\thetast_\rho,\varrho_\rho).
\end{equation}
\end{lemma}

\begin{proof}
The small-gain bound in the proof of Theorem~\ref{thm:momentum} gives
$\|h_\rho-h_0\|+\|\nabla h_\rho-\nabla h_0\|_{\op}\le C\rho$ on $\Bcal_R$, with $h_0(\theta):=\bar P(\theta)(A\theta-b)$
and $C$ free of $\lambda$. The proof of Proposition~\ref{prop:smallgain}(iv) uses only this bound, so it gives
$\rho_0$ free of $\lambda$, the zero $\thetast_\rho$, and $J_\rho=J_0+O(\rho)$, which is
positive stable once $\rho$ is small, as the Lyapunov certificate of $J_0$ also certifies $J_\rho$. Lyapunov's
theorem \citep[Cor.~2.2.4]{hornjohnson1991} supplies $\Qly_\rho\succ0$ with
$J_\rho^{\top}\Qly_\rho+\Qly_\rho J_\rho=I$, so $q^{\top}\Qly_\rho J_\rho q=\frac12\|q\|^{2}$ for every $q\in\R^d$. By Proposition~\ref{prop:controlled}(ii), the field is $C^{1}$, so with $q:=\theta-\thetast_\rho$, $h_\rho(\theta)=J_\rho q+r(\theta)$ with $r(\theta)=\int_0^1\{\nabla h_\rho(\thetast_\rho+\upsilon q)-J_\rho\}\,d\upsilon\,q$. Continuity of $\nabla h_\rho$
gives $\varrho_\rho>0$ with $\|\nabla h_\rho-J_\rho\|_{\op}\le1/(4\|\Qly_\rho\|_{\op})$ on
$\Bcal(\thetast_\rho,\varrho_\rho)$, so
$q^{\top}\Qly_\rho h_\rho(\theta)\ge\frac12\|q\|^{2}-\|q\|\,\|\Qly_\rho\|_{\op}\|r(\theta)\|\ge\frac14\|q\|^{2}$ there.
Reducing $\varrho_\rho$ puts the closed ball inside $\Bcal_R$, since $\|\thetast\|<R$ and
$\thetast_\rho=\thetast+O(\rho)$.
\end{proof}

The argument is Lyapunov's indirect method \citep[Thm.~4.7]{khalil2002}, in the local coercivity form standard in stochastic approximation \citep{kushner2003,borkar2021}. The non-Euclidean metric is needed: Euclidean coercivity \citep{chenzw2022} would require
$\Sym(J_\rho)\succ0$, which fails at small $\rho$ for twelve of the $27$ taxi rules ($\Sym(J_0)$ has eigenvalues down to $-0.049$). The lemma concerns the field and gives no lower bound on
$\varrho_\rho$ in terms of the primitives. Assumption~\ref{a:deployedstable} imposes confinement of the iterates.

\subsection{Consistency and limit laws at the displaced target, and the proof of Theorem~\ref{thm:deployedcons}}
\label{app:deployedcons}

\begin{theorem}[Consistency and limit theory at the displaced target]
\label{thm:deployedclt}
Let $\rho<\rho_0$, let Assumptions~\ref{a:data} and~\ref{a:deployedstable} hold for
\eqref{eq:momentum} at a delay $m\ge0$, and let $X_1\in\mathcal X$. Let $\tilde\theta_t$ be the recursion
projected onto $\Theta$ (Step~0 of the proof), which equals $\theta_t$ for every $t$ on $\mathcal E_\Theta$.
\begin{enumerate}[label=(\roman*),leftmargin=2.2em,itemsep=2pt]
\item $\tilde\theta_t\to\thetast_\rho$ almost surely. Hence, $\theta_t\to\thetast_\rho$ almost surely on
  $\mathcal E_\Theta$, which is Theorem~\ref{thm:deployedcons}, and $J:=\nabla h_\rho(\thetast_\rho)$ is positive stable.
\item For every integer $p\ge1$, there is $C_{\rho,p}<\infty$ with
  \begin{equation}
  \label{eq:deployedmoment}
    \E\|\tilde\theta_t-\thetast_{\rho}\|^{2p}\;\le\;C_{\rho,p}\,t^{-p\alpha},
    \qquad\text{hence,}\qquad
    \E\big[\|\theta_t-\thetast_{\rho}\|^{2p};\,\mathcal E_\Theta\big]\;\le\;C_{\rho,p}\,t^{-p\alpha}.
  \end{equation}
\item Let $\Gamma_P$ be the long-run variance of $H_{\thetast_\rho}(X_t)=\phi(v_{t-m})u_t$ under the frozen
  stationary law at $\thetast_\rho$, and put
  $\Psi_{\rho,n}(\xi):=n^{-1/2}\sum_{t\le\lfloor n\xi\rfloor}(\theta_t-\thetast_\rho)$. Then, conditionally on
  $\mathcal E_\Theta$,
  \begin{equation*}
    \Psi_{\rho,n}\;\weakto\;J^{-1}\Gamma_P^{1/2}W\quad\text{in }D[0,1]^d,
    \qquad\text{and so}\qquad
    \sqrt n\,(\bth_n-\thetast_{\rho})\;\dto\;\mathcal N\big(0,\,J^{-1}\Gamma_PJ^{-\top}\big).
  \end{equation*}
\end{enumerate}
For the projected recursion, both limits hold unconditionally and the convergence is mixing: it
also holds conditionally on every event of positive probability in $\sigma(\cup_k\Fcal_k)$. The gain is fixed throughout, so
neither limit is a joint limit in $(n,\rho)$, and the constants are not uniform as $\rho\downarrow0$ or
$\lambda\uparrow1$.
\end{theorem}

The proof follows the Poisson-equation approach of \citet[Part~II, Ch.~1]{benveniste1990}, with
Proposition~\ref{prop:controlled} giving the Poisson regularity from the contracting recursions for $v$ and $u$, in place of the minorization in
\citet[Prop.~4.1(iii)]{fort2015}. The intervals of Section~\ref{sec:rrlaw} rest on this functional limit.

\begin{proof}[Proof of Theorem~\ref{thm:deployedclt}]
Throughout, $\Qly_\rho$ is the metric of Lemma~\ref{lem:basin}, $\|q\|^2_{\Qly_\rho}:=q^{\top}\Qly_\rho q$,
and $\kappa_{\Qly}$ is the square root of the condition number of $\Qly_\rho$. The constants $C$ below
depend on $\rho$ and $\lambda$ through $\|\Qly_\rho\|_{\op}$, $\lambda_{\min}(\Qly_\rho)$, $\|\hat f\|_\infty$, and
$L_{\hat f,\rho}$, and on $\epsilon$, $G$, $\eta_0$, and $\alpha$.

\emph{Step 0: localization.} We replace the recursion by a projected copy that coincides with it on
$\mathcal E_\Theta$, because conditioning on $\mathcal E_\Theta$, an event determined by the whole path, need
not preserve the martingale structure. Let
$\operatorname{proj}^{\Qly}_\Theta(\theta):=\arg\min_{\theta'\in\Theta}(\theta'-\theta)^\top\Qly_\rho(\theta'-\theta)$ be the projection onto the closed
convex set $\Theta$ in the metric $\Qly_\rho$, and let $\tilde\theta_t$ be the recursion driven by the same
data and initial state $X_1$,
$\tilde\theta_{t+1}=\operatorname{proj}^{\Qly}_\Theta\big(\tilde\theta_t-\eta_tH_{\tilde\theta_t}(\tilde X_t)\big)$ with
$\tilde\theta_1:=\operatorname{proj}^{\Qly}_\Theta(\theta_1)$, which is $\theta_1$ on $\mathcal E_\Theta$. The accumulator and momentum components of $\tilde X_t$ are updated using
$g(\tilde\theta_t,\zeta_t)$. Since $\tilde\theta_t\in\Theta\subseteq\Bcal_R$, these updates keep $\tilde X_t$ in $\mathcal X$. On
$\mathcal E_\Theta$, the unprojected update never leaves $\Theta$, so, by induction, the projection is never
active and $(\tilde\theta_t,\tilde X_t)=(\theta_t,X_t)$ for every $t$.

The projected recursion lies in
$\Theta\subseteq\Bcal_R$ on the whole probability space, so Proposition~\ref{prop:controlled} and
Lemma~\ref{lem:basin} apply to it without restriction. The projection is nonexpansive in
$\|\cdot\|_{\Qly_\rho}$ and fixes $\Theta$, which contains $\thetast_\rho$ and $\tilde\theta_t$. Hence,
$\|\tilde\theta_{t+1}-\thetast_\rho\|_{\Qly_\rho}\le\|\tilde\theta_t-\eta_tH_{\tilde\theta_t}(\tilde X_t)-\thetast_\rho\|_{\Qly_\rho}$
and $\|\tilde\theta_{t+1}-\tilde\theta_t\|_{\Qly_\rho}\le\eta_t\|H_{\tilde\theta_t}(\tilde X_t)\|_{\Qly_\rho}$.
Until Step~8, we write $(\theta_t,X_t)$ for $(\tilde\theta_t,\tilde X_t)$ and set $\Delta^\rho_t:=\theta_t-\thetast_\rho$.

\emph{Step 1: the noise decomposition.} Let $\Fcal_t$ be the filtration of Section~\ref{sec:deployed}, with $\theta_1$ unprojected, so that
the initial state $X_1=(v_{1-\bar m},\dots,v_0,u_0,\zeta_1)$ is $\Fcal_1$-measurable. Then, $\theta_t$ is
$\Fcal_{t-1}$-measurable, and conditionally on $\Fcal_t$, $X_{t+1}\sim\Pi_{\theta_t}(X_t,\cdot)$, because
the accumulators and the momentum at time $t$ are computed with $\theta_t$ and
$\zeta_{t+1}\sim\Qcal(\zeta_t,\cdot)$. Before projection, the update reads
\begin{equation}
\label{eq:deployedrec}
  \theta_t-\eta_tH_{\theta_t}(X_t)=\theta_t-\eta_t\,h_\rho(\theta_t)-\eta_t\,U_{t+1},
  \qquad U_{t+1}:=f_{\theta_t}(X_t),
\end{equation}
and the Poisson equation of Proposition~\ref{prop:controlled} gives
\begin{equation}
\label{eq:deployedpoissonsplit}
 U_{t+1}=M_{t+1}+\hat f_{\theta_t}(X_t)-\hat f_{\theta_t}(X_{t+1}),
 \qquad
 M_{t+1}:=\hat f_{\theta_t}(X_{t+1})-\Pi_{\theta_t}\hat f_{\theta_t}(X_t),
\end{equation}
with $\E[M_{t+1}\mid\Fcal_t]=0$ and $\|M_{t+1}\|\le2\|\hat f\|_\infty$, finite by \eqref{eq:hatfbound}. The projected iterate moves by at most
$\kappa_{\Qly}\eta_t\|\phi(v_{t-m})u_t\|\le\kappa_{\Qly}\epsilon^{-1/2}G\eta_t$ in each step, by
Lemma~\ref{lem:increment} and the second property of the projection.

\emph{Step 2: the mean-square rate and consistency.} A perturbed Lyapunov function, corrected by the
Poisson solution, isolates the martingale part of the noise. Subtracting $\thetast_\rho$, using the first
property of the projection, and expanding the quadratic form, with $\|h_\rho\|_\infty\le\epsilon^{-1/2}G$
and $\|f_\theta\|_\infty\le2\epsilon^{-1/2}G$ on $\Bcal_R$, we obtain
\[
  \|\Delta^\rho_{t+1}\|^{2}_{\Qly_\rho}
  \le\|\Delta^\rho_t\|^{2}_{\Qly_\rho}-2\eta_t(\Delta^\rho_t)^{\top}\Qly_\rho h_\rho(\theta_t)
   -2\eta_t(\Delta^\rho_t)^{\top}\Qly_\rho U_{t+1}+\eta_t^{2}\|h_\rho(\theta_t)+U_{t+1}\|^{2}_{\Qly_\rho},
\]
the last term being at most $C\eta_t^{2}$ since
$h_\rho(\theta_t)+U_{t+1}=H_{\theta_t}(X_t)$. The drift
term is at most $-\frac12\eta_t\|\Delta^\rho_t\|^{2}$ by \eqref{eq:deployedcoercive}, since
$\theta_t\in\Theta$. Insert \eqref{eq:deployedpoissonsplit} into the noise term and define the perturbed
Lyapunov function
\[
  \mathcal W_t:=\|\Delta^\rho_t\|^{2}_{\Qly_\rho}-2\eta_t(\Delta^\rho_t)^{\top}\Qly_\rho\hat f_{\theta_t}(X_t)+C_0\eta_t^{2},
  \qquad C_0:=2\|\Qly_\rho\|_{\op}^{2}\|\hat f\|_\infty^{2}/\lambda_{\min}(\Qly_\rho),
\]
whose linear correction comes from the $X_t$-dependent term of the coboundary
$\hat f_{\theta_t}(X_t)-\hat f_{\theta_t}(X_{t+1})$ in \eqref{eq:deployedpoissonsplit}. Young's inequality,
$2\eta_t|(\Delta^\rho_t)^{\top}\Qly_\rho\hat f_{\theta_t}(X_t)|\le\frac12\lambda_{\min}(\Qly_\rho)\|\Delta^\rho_t\|^{2}+C_0\eta_t^{2}$, gives
\begin{equation}
\label{eq:deployedWequiv}
 \tfrac12\lambda_{\min}(\Qly_\rho)\|\Delta_t^\rho\|^2
 \le \mathcal W_t\le C\{\|\Delta_t^\rho\|^2+\eta_{t}^2\}.
\end{equation}

In terms of $\mathcal W_t$, the term produced by the coboundary cancels the correction, and what remains is
\[
\begin{aligned}
  \mathcal W_{t+1}&\le \mathcal W_t-\tfrac12\eta_t\|\Delta^\rho_t\|^{2}-2\eta_t(\Delta^\rho_t)^{\top}\Qly_\rho M_{t+1}+C\eta_t^{2}\\
  &\quad+2\eta_t(\Delta^\rho_t)^{\top}\Qly_\rho\hat f_{\theta_t}(X_{t+1})-2\eta_{t+1}(\Delta^\rho_{t+1})^{\top}\Qly_\rho\hat f_{\theta_{t+1}}(X_{t+1}),
\end{aligned}
\]
using $C_0(\eta_{t+1}^{2}-\eta_t^{2})\le0$. Three estimates bound the last difference by
$C\eta_t^{2}(1+\|\Delta^\rho_t\|)$: $\eta_t-\eta_{t+1}\le\alpha\eta_0t^{-\alpha-1}\le C\eta_t^{2}$ because
$\alpha<1$, $\|\Delta^\rho_t-\Delta^\rho_{t+1}\|\le\kappa_{\Qly}\epsilon^{-1/2}G\eta_t$ by Step~1 and the second
property of the projection, and
$\|\hat f_{\theta_t}(X_{t+1})-\hat f_{\theta_{t+1}}(X_{t+1})\|\le L_{\hat f,\rho}\kappa_{\Qly}\epsilon^{-1/2}G\eta_t$ by
\eqref{eq:hatflip}. Young's inequality, $C\eta_t^{2}\|\Delta^\rho_t\|\le\frac18\eta_t\|\Delta^\rho_t\|^{2}+C\eta_t^{3}$, absorbs them, leaving
\begin{equation}
\label{eq:Wonestep}
  \mathcal W_{t+1}\le \mathcal W_t-\tfrac38\eta_t\|\Delta^\rho_t\|^{2}+C\eta_t^{2}+\eta_tN_{t+1},
  \qquad N_{t+1}:=-2(\Delta^\rho_t)^{\top}\Qly_\rho M_{t+1},
\end{equation}
with $\E[N_{t+1}\mid\Fcal_t]=0$ and $|N_{t+1}|\le C\mathcal W_t^{1/2}$.

This one-step inequality gives the mean-square rate and almost sure convergence. By the upper bound in \eqref{eq:deployedWequiv}, $\|\Delta^\rho_t\|^{2}\ge \mathcal W_t/C-\eta_t^{2}$,
so $\E[\mathcal W_{t+1}\mid\Fcal_t]\le(1-c_\rho\eta_t)\mathcal W_t+C\eta_t^{2}$ for some $c_\rho>0$, and
\citet[Lem.~4]{chung1954} gives $\E \mathcal W_t\le C_\rho t^{-\alpha}$, which yields \eqref{eq:deployedmoment} at $p=1$.
Moreover, $\mathcal W_t\ge0$ and $\sum_t\eta_t^{2}<\infty$, so, by the Robbins--Siegmund theorem \citep{robbins1971},
$\mathcal W_t$ converges almost surely and $\sum_t\eta_t\|\Delta^\rho_t\|^{2}<\infty$ almost surely. Since
$\sum_t\eta_t=\infty$, $\liminf_t\|\Delta^\rho_t\|=0$, so $\mathcal W_t\to0$ by \eqref{eq:deployedWequiv} and $\tilde\theta_t\to\thetast_\rho$ almost surely.

\emph{Consistency.} On $\mathcal E_\Theta$, the two recursions coincide (Step~0), so the unprojected iterates converge to $\thetast_\rho$
almost surely on $\mathcal E_\Theta$, which is \eqref{eq:deployedcons}. The Jacobian
$J=J_\rho$ is positive stable by Lemma~\ref{lem:basin}. This proves (i).

\emph{Step 3: higher moments.} The $2p$th-moment rate follows by induction on $p$, adapting the device in
the proof of \citet[Thm.~9]{srikant2019}: raise the perturbed Lyapunov function to the $p$th power.
Since $\|\Delta^\rho_t\|\le2R$ (Step~0), only the rate is at issue. Fix $t_0$ with $c_\rho\eta_t<1$ for
$t\ge t_0$, and absorb the finitely many earlier indices into the constants. From
\eqref{eq:Wonestep}, $0\le \mathcal W_{t+1}\le a_t+\eta_tN_{t+1}$ with $a_t:=(1-c_\rho\eta_t)\mathcal W_t+C\eta_t^2$ being $\Fcal_t$-measurable
and $|N_{t+1}|\le C\mathcal W_t^{1/2}$. Both sides are
nonnegative, so we may raise them to the $p$th power. Conditioning on $\Fcal_t$ then eliminates the term linear in $N_{t+1}$.

Every remaining term of the expanded power except the leading one has expectation $O(\eta_t^{p+1})$. Expanding both
binomials, with $k$ and $a'$ the powers of $\eta_tN_{t+1}$ and $C\eta_t^2$, we bound every remaining term
by a constant times $\eta_t^{k+2a'}\mathcal W_t^{p-k/2-a'}$, with $k\ne1$ and $0\le a'\le p-k$. The power of
$\mathcal W_t$ is at most $p-1$ unless $k=a'=0$, so the induction hypothesis applies through
Lyapunov's inequality, $\E \mathcal W_t^{p'}\le(\E \mathcal W_t^{p-1})^{p'/(p-1)}=O(\eta_t^{p'})$ for $0\le p'\le p-1$. Every
such term then has expectation $O(\eta_t^{p+1})$, and the leading term is bounded using
$(1-c_\rho\eta_t)^p\le1-pc_\rho\eta_t+C\eta_t^2$.
Chung's lemma applied to $\E \mathcal W_{t+1}^p\le(1-c_{\rho,p}\eta_t)\E \mathcal W_t^p+C_{\rho,p}\eta_t^{p+1}$ closes the
induction, and \eqref{eq:deployedWequiv} gives \eqref{eq:deployedmoment}. Step~0 transfers
the bound to $\mathcal E_\Theta$, which proves (ii).

\emph{Step 4: the summation identity.} By Proposition~\ref{prop:controlled}(ii),
$h_\rho(\thetast_\rho+\Delta)=J\Delta+r(\Delta)$ with $\|r(\Delta)\|\le C_{\mathrm T}\|\Delta\|^{2}$ on $\Theta$
for a constant $C_{\mathrm T}$. By \eqref{eq:deployedrec}, the projected update is
$\tilde\theta_{t+1}=\tilde\theta_t-\eta_th_\rho(\tilde\theta_t)-\eta_tU_{t+1}+\{\operatorname{proj}^{\Qly}_\Theta(y_t)-y_t\}$, with $y_t$ the pre-projection point. Substituting the expansion,
dividing by $\eta_t$, summing over $t\le k$, and applying Abel summation to the increments, we obtain,
for every $k\ge1$, the exact identity
\begin{equation}
\label{eq:deployedcltdecomp}
 J\sum_{t\le k}\Delta_t^\rho
 =\eta_1^{-1}\Delta_1^\rho-\eta_k^{-1}\Delta_{k+1}^\rho
 +\sum_{t=2}^k(\eta_t^{-1}-\eta_{t-1}^{-1})\Delta_t^\rho
 -\sum_{t\le k}U_{t+1}-\sum_{t\le k}r(\Delta_t^\rho)+\sum_{t\le k}\eta_t^{-1}\{\operatorname{proj}^{\Qly}_\Theta(y_t)-y_t\},
\end{equation}
and applying summation by parts to \eqref{eq:deployedpoissonsplit} yields
\begin{equation}
\label{eq:noisesum}
  \sum_{t\le k}U_{t+1}=\sum_{t\le k}M_{t+1}+\hat f_{\theta_1}(X_1)-\hat f_{\theta_k}(X_{k+1})
  +\sum_{t=2}^{k}\big\{\hat f_{\theta_t}(X_t)-\hat f_{\theta_{t-1}}(X_t)\big\}.
\end{equation}
Thus, $\Psi_{\rho,n}(\xi)=-J^{-1}n^{-1/2}\sum_{t\le\lfloor n\xi\rfloor}M_{t+1}+J^{-1}n^{-1/2}\mathcal R_{\lfloor n\xi\rfloor}$,
where $\mathcal R_k$ collects the initial term, the endpoint term $-\eta_k^{-1}\Delta^\rho_{k+1}$, the Abel
sum, the Taylor sum, the coboundary endpoints and drift sum of \eqref{eq:noisesum}, and the projection sum, with $\mathcal R_0:=0$.

\emph{Step 5: the remainders are uniformly negligible.} Every remainder other than the endpoint term
vanishes after division by $\sqrt n$. By \eqref{eq:deployedmoment} and Jensen's inequality,
$\E\|\Delta^\rho_t\|\le C_\rho^{1/2}t^{-\alpha/2}$. In \eqref{eq:deployedcltdecomp} and \eqref{eq:noisesum}, the initial term and the two coboundary endpoints are bounded uniformly in $k$, the latter by
$\|\hat f\|_\infty$, and the Abel, Taylor, drift, and projection sums are bounded in norm, uniformly in $k\le n$, by sums of
nonnegative quantities over $t\le n$. After division by $\sqrt n$, the bounded terms are $O(n^{-1/2})$. The normalized Abel sum, with
$0\le\eta_t^{-1}-\eta_{t-1}^{-1}\le2\alpha\eta_0^{-1}t^{\alpha-1}$ for $t\ge2$, has expected norm at most
$Cn^{\alpha/2-1/2}$. The normalized Taylor sum has expected norm at most
$C_{\mathrm T}C_\rho(1-\alpha)^{-1}n^{1/2-\alpha}$, which vanishes exactly because $\alpha>1/2$. The
normalized drift sum of \eqref{eq:noisesum} has norm at most
$n^{-1/2}L_{\hat f,\rho}\kappa_{\Qly}\epsilon^{-1/2}G\sum_{t<n}\eta_t=O(n^{1/2-\alpha})$ by \eqref{eq:hatflip} and
Step~1. For the projection sum,
$\|y_t-\tilde\theta_t\|\le\epsilon^{-1/2}G\eta_t\to0$ and $\tilde\theta_t\to\thetast_\rho$, which lies in the interior of
$\Theta$, so, almost surely, the projection is eventually inactive and the sum is eventually
constant.

The endpoint term needs a separate maximal bound, which a dyadic-block estimate gives. On each block
$I_j:=[2^{j},2^{j+1})\cap\{1,\dots,n\}$, $\eta_k^{-1}\le\eta_0^{-1}2^{(j+1)\alpha}$, and by \eqref{eq:deployedmoment}
with an integer $p>1/(1-\alpha)$,
$\E\max_{k\in I_j}\|\Delta^\rho_{k+1}\|\le(\sum_{k\in I_j}\E\|\Delta^\rho_{k+1}\|^{2p})^{1/(2p)}\le C2^{\,j(1/(2p)-\alpha/2)}$.
Summing over $0\le j\le\log_2n$ gives
$\E\max_{k\le n}\eta_k^{-1}\|\Delta^\rho_{k+1}\|\le Cn^{\alpha/2+1/(2p)}$, and
$n^{-1/2}n^{\alpha/2+1/(2p)}\to0$ exactly when $p>1/(1-\alpha)$. Hence,
$\sup_{\xi\in[0,1]}n^{-1/2}\|\mathcal R_{\lfloor n\xi\rfloor}\|\pto0$.

\emph{Step 6: the predictable quadratic variation.} The conditional covariance of $M_{t+1}$ given
$\Fcal_t$ is $Q_{\theta_t}(X_t)$. With $S_k:=\sum_{t<k}\{Q_{\theta_t}(X_t)-\Gamma_P\}$ and
$\Gamma_P=\int Q_{\thetast_\rho}\,d\pi_{\thetast_\rho}$, Proposition~\ref{prop:controlled}(iv) gives $S_k/k\to0$ almost surely, since $\tilde\theta_t\to\thetast_\rho$
almost surely by Step~2. The convergence is also uniform over the time index by a
deterministic argument. Given $\delta>0$, choose a random index $K$ with $\|S_k\|\le\delta k$ for $k\ge K$. Then,
$n^{-1}\max_{k\le n}\|S_k\|\le n^{-1}\max_{k<K}\|S_k\|+\delta$, whose limit superior is at most $\delta$. Since
the summands are bounded, endpoint shifts and rounding $n\xi$ down to an integer introduce only uniform $O(n^{-1})$ errors, so
$\sup_{\xi\in[0,1]}\|n^{-1}\sum_{t\le\lfloor n\xi\rfloor}Q_{\theta_t}(X_t)-\xi\Gamma_P\|\to0$ almost surely.

\emph{Step 7: the functional limit and the central limit theorem.} The increments $n^{-1/2}M_{t+1}$ are
$O(n^{-1/2})$, so the conditional Lindeberg condition holds, and by Step~6, their predictable
quadratic variation converges to $\xi\Gamma_P$ uniformly in $\xi$. The martingale central limit
theorems \citep[Thms.~18.1 and~18.2]{billingsley1999}, applied coordinatewise for tightness and, through
the Cram\'er--Wold device, to $n^{-1/2}\psi(t/n)^{\top}M_{t+1}$ for step functions $\psi:[0,1]\to\R^d$, give
$n^{-1/2}\sum_{t\le\lfloor n\cdot\rfloor}M_{t+1}\weakto\Gamma_P^{1/2}W$ in $D[0,1]^d$. With Steps~4--5 and
the continuous mapping theorem, $\Psi_{\rho,n}\weakto J^{-1}\Gamma_P^{1/2}W$, as $-W$ and $W$
have the same law. Evaluation at $\xi=1$, which is continuous at continuous paths,
gives the central limit theorem, since $\Psi_{\rho,n}(1)=\sqrt n(\bar{\tilde\theta}_n-\thetast_\rho)$.

\emph{Step 8: back to the unprojected recursion, conditionally on $\mathcal E_\Theta$.} The limits
transfer to the conditional law given $\mathcal E_\Theta$ because the convergence of Step~7 is mixing. The
bound $\Pr(\cdot\mid\mathcal E_\Theta)\le\Pr(\cdot)/\Pr(\mathcal E_\Theta)$ carries convergence in
probability over to the conditional law, but not convergence in distribution. Let $Z_n$ be
$\sqrt n(\bth_n-\thetast_\rho)$ or $\Psi_{\rho,n}$, and $\tilde Z_n$ the same statistic of the projected
path, so that $Z_n=\tilde Z_n$ on $\mathcal E_\Theta$, and write $\mathcal L$ for the limit law of
Step~7. The convergence $\tilde Z_n\dto\mathcal L$ is mixing: it also holds under $\Pr(\cdot\mid A)$
for every fixed $k$ and every $A\in\Fcal_k$ with $\Pr(A)>0$. Under $\Pr(\cdot\mid A)$, the increments $M_{t+1}$ of Step~1,
$t\ge k$, are still bounded martingale differences for $(\Fcal_t)$, the averages of their conditional
covariances still converge to $\Gamma_P$ in probability, the first $k$ increments contribute
$O(n^{-1/2})$, and every remainder of Step~5 that is $o_{\Pr}(1)$ is $o_{\Pr(\cdot\mid A)}(1)$. Thus, the
martingale limit theorem under $\Pr(\cdot\mid A)$ yields the same limit law $\mathcal L$. The extension below establishes mixing
convergence in the sense of \citet{aldous1978}.

To obtain the conditional limit given $\mathcal E_\Theta$, define
$\mathcal E_k:=\{\theta_t\in\Theta\text{ for every }t\le k\}\in\Fcal_k$. These events decrease to $\mathcal E_\Theta$.
For a continuity set $B$ of $\mathcal L$,
\[
  \big|\Pr(Z_n\in B,\ \mathcal E_\Theta)-\mathcal L(B)\Pr(\mathcal E_\Theta)\big|
  \le\big|\Pr(\tilde Z_n\in B,\ \mathcal E_k)-\mathcal L(B)\Pr(\mathcal E_k)\big|
  +2\Pr(\mathcal E_k\setminus\mathcal E_\Theta).
\]
Letting $n\to\infty$ removes the first term, by mixing at fixed $k$, and letting $k\to\infty$ then
removes the second. Dividing by $\Pr(\mathcal E_\Theta)>0$ gives (iii). For $A\in\sigma(\cup_k\Fcal_k)$, choose $A_k\in\Fcal_k$ with $\Pr(A_k\triangle A)\to0$. The bound with $\tilde Z_n$, $A$, and $A_k$ in place of $Z_n$, $\mathcal E_\Theta$, and $\mathcal E_k$, with last term $2\Pr(A_k\triangle A)$, extends the mixing to $A$.

The same argument applies to any
statistic that equals a functional of the partial-sum process plus a remainder negligible in probability,
provided that the functional is continuous at almost every limiting path. This covers the self-normalized ratio of Theorem~\ref{thm:rr}(iii) and the pivot of the two-block interval \eqref{eq:rrbm}: their functionals are ratios, and when $\ell^\top\Sigma_{\mathrm{RR}}\ell>0$,
the denominators are positive at almost every limiting path.

\emph{Identification of $\Gamma_P$.} Under the frozen stationary law at $\thetast_\rho$, $h_\rho(\thetast_\rho)=0$
makes $f_{\thetast_\rho}(X_t)=H_{\thetast_\rho}(X_t)=\phi(v_{t-m})u_t$. With the parameter frozen, summing the identity \eqref{eq:deployedpoissonsplit}
expresses the partial sums as a sum of stationary martingale differences plus two bounded coboundary
endpoints, so, by \citet{gordin1969}, the long-run variance is the second-moment matrix of the martingale
difference, $\pi_{\thetast_\rho}Q_{\thetast_\rho}=\Gamma_P$.
\end{proof}

\begin{remark}[Adam's bias corrections]
\label{rem:biascorr}
With Adam's bias corrections, parts~(i)--(iii) of Theorem~\ref{thm:deployedclt} hold on the corrected run's own
confinement event, and Theorem~\ref{thm:rr} holds on the two corrected runs' joint confinement event, each when it has positive probability. From zero
initial averages, dividing by $1-\beta_2^t$ and $1-\beta_1^t$ turns \eqref{eq:sarmsprop} and
\eqref{eq:momentum} into the same convex-combination recursions with rates
$\rho_t:=\rho/\{1-(1-\rho)^t\}$ and $\lambda_t:=\lambda(1-\lambda^{t-1})/(1-\lambda^t)$, where $0\le\rho_t-\rho\le(1-\rho)^t$ and $0\le\lambda-\lambda_t\le\lambda^t$. The state therefore stays in
$\mathcal X$ while the parameter is in $\Bcal_R$, and on $\Bcal_R\times\mathcal X$, the corrections change the update and field by at most $C\max\{1-\rho,\lambda\}^t\le C_{\rho,\lambda}\eta_t$. As
$\hat f_\theta$ and $\hat q^\circ$ are Lipschitz in the deterministic coordinates, this adds at most
$C\eta_t^2$ per step to \eqref{eq:Wonestep}, whose $C\eta_t^2$ term absorbs it.
Elsewhere, it adds terms with bounded sums.
\end{remark}

\subsection{Proof of Theorem~\ref{thm:rr}}
\label{app:rr}

\begin{proof}[Proof of Theorem~\ref{thm:rr}]
\emph{(i)} The two recursions, stacked as in \citet[Thm.~5]{zeng2026}, form one recursion of the
same kind, to which every step of the proofs of Proposition~\ref{prop:controlled}, Lemma~\ref{lem:basin},
and Theorem~\ref{thm:deployedclt} applies. The enlarged state is
$(v^{(1)}_{t-\bar m},\dots,v^{(1)}_{t-1},u^{(1)}_{t-1},v^{(2)}_{t-\bar m},\dots,v^{(2)}_{t-1},u^{(2)}_{t-1},\zeta_t)$,
driven by the same tuple chain, with parameter $(\theta^{(1)},\theta^{(2)})\in\R^{2d}$ and field
$(H^{(1)}_{\theta^{(1)}},H^{(2)}_{\theta^{(2)}})$. The forgetting rate is $\max\{1-\rho_1,1-\rho_2,\lambda,\kappa\}^{1/2}$ for the two gains
$\rho_1,\rho_2$. The Jacobian at the stacked target is $\Diag(J_{s},J_{\omega s})$, positive stable by
Lemma~\ref{lem:basin} applied to each block. The Lyapunov metric is block diagonal, the projection acts
on each parameter separately, and the confinement event is $\mathcal E$.

The normalized joint predictable quadratic variation converges to $\boldsymbol\Gamma$, whose off-diagonal block comes
from the shared tuple chain. The conditional covariance of the stacked martingale increment is a
$2\times2$ block matrix, and Step~7 of the proof of Proposition~\ref{prop:controlled} and Step~6 of the
proof of Theorem~\ref{thm:deployedclt} give the convergence. As in the identification of $\Gamma_P$, the
off-diagonal block $\Gamma_\times$ is the long-run cross-covariance of the two fields under the joint
frozen law. A joint limit does not follow from the two marginal limits: Steps~7 and~8 of that proof, applied to the
stacked chain, give the joint functional limit of both partial-sum processes, conditionally on
$\mathcal E$.

\emph{(ii)} $\hat\theta^{\mathrm{RR}}_n=\nu_1\bth_n(s)+\nu_2\bth_n(\omega s)$, so
$\sqrt n(\hat\theta^{\mathrm{RR}}_n-\thetast_{\mathrm{RR}}(s))$ is the $\nu$-weighted sum of the two blocks of the vector of part (i),
exactly and for every $n$, because $\nu_1\thetast(s)+\nu_2\thetast(\omega s)=\thetast_{\mathrm{RR}}(s)$ by definition.
Its limit is the image of the joint limit, with covariance $\Sigma_{\mathrm{RR}}$ of \eqref{eq:rrclt}. The
centering is exact at each $s$, so the remainder of the expansion $\thetast_{\mathrm{RR}}(s)=\thetast-\omega s^{2}b_2+O(s^{3})$
of Corollary~\ref{cor:rr} never enters the limit.

\emph{(iii)} Put $\vartheta:=\ell^\top\thetast_{\mathrm{RR}}(s)$ and
$\Phi_n(\xi):=n^{-1/2}\sum_{t\le\lfloor n\xi\rfloor}(Y_t-\vartheta)$. It is the image of the pair of partial-sum
processes of part (i) under the fixed linear map $(q_1,q_2)\mapsto\ell^\top(\nu_1q_1+\nu_2q_2)$, so
$\Phi_n\weakto\sigma_{\mathrm{RR}}W$ in $D[0,1]$, conditionally on $\mathcal E$, with $\sigma_{\mathrm{RR}}^{2}=\ell^\top\Sigma_{\mathrm{RR}}\ell>0$.
Since $\sum_{t\le k}(Y_t-\bar Y_n)=\sqrt n\,\widetilde\Phi_n(k/n)$ with $\widetilde\Phi_n(\xi):=\Phi_n(\xi)-\xi\Phi_n(1)$, the statistic $\widehat\sigma_{\mathrm{RR},n}^{2}=n^{-1}\sum_{k}\widetilde\Phi_n(k/n)^{2}$ differs from $\int_0^1\widetilde\Phi_n^{2}$ by a pathwise $O(n^{-1}|\Phi_n(1)|\sup|\widetilde\Phi_n|)$, and continuous
mapping gives the pivot: this step in \citet[Cor.~6 with $q=1$]{zeng2026} and \citet[Thm.~1]{lee2022} uses only the
functional limit, which part (i) supplies, and the limit has a continuous law and an almost surely positive denominator.
Set the ratio to $0$ when $\widehat\sigma_{\mathrm{RR},n}=0$, an event of vanishing probability.

For the two-block interval \eqref{eq:rrbm}, let $\bar Y_{\mathrm I}$ and $\bar Y_{\mathrm{II}}$ be the means
of $Y_t$ over the first and second halves, in time order, of the iterates entering the average, with $n$ even. By part (i), $\sqrt n(\bar Y_{\mathrm I}-\vartheta)$ and
$\sqrt n(\bar Y_{\mathrm{II}}-\vartheta)$ converge jointly to
multiples of the increments of $\sigma_{\mathrm{RR}}W$ over two disjoint intervals of equal length, which
are independent centered Gaussians with a common variance $\sigma_{\mathrm{blk}}^{2}$. Their sum and difference are
therefore independent $\mathcal N(0,2\sigma_{\mathrm{blk}}^{2})$ variables, so, by continuous mapping,
$\{\frac12(\bar Y_{\mathrm I}+\bar Y_{\mathrm{II}})-\vartheta\}/\{\frac12|\bar Y_{\mathrm I}-\bar Y_{\mathrm{II}}|\}\dto\mathcal T_1$,
Student's $t$ with one degree of freedom, as the limit of the $\sqrt n$-scaled denominator is positive almost surely.
Set the ratio to $0$ when $\bar Y_{\mathrm I}=\bar Y_{\mathrm{II}}$, an asymptotically null event.

Both pivots also hold on the second half of the run, where our numerical studies compute the
intervals, with $n'=n/2$ as the sample size. For simplicity, take $n$ divisible by four. The process
$\Phi'_n(\xi):=n'^{-1/2}\sum_{n/2<t\le n/2+\lfloor n'\xi\rfloor}(Y_t-\vartheta)=\sqrt2\{\Phi_n(\frac12+\frac{\xi}{2})-\Phi_n(\frac12)\}$ converges to
$\sqrt2\,\sigma_{\mathrm{RR}}\{W(\frac12+\frac{\xi}{2})-W(\frac12)\}$, again $\sigma_{\mathrm{RR}}$ times a standard Brownian motion, and the same continuous mapping arguments give both pivots for $\Phi'_n$. The qualifications carry over unchanged: the contrast is fixed before the data are examined, $\ell^\top\Sigma_{\mathrm{RR}}\ell>0$, and the limits are conditional on $\mathcal E$ by Step~8 of the proof of Theorem~\ref{thm:deployedclt}.
\end{proof}

\section{Supplementary simulations and designs}
\label{app:numerics}

This section gives the designs behind Section~\ref{sec:numerics} and the numbers for each
check the main text cites. The replication package records the remaining settings and the script that generates each table and figure. Every design has bounded rewards, as Assumption~\ref{a:bounded} requires, and every reported average and interval uses the second half of the run. 

\subsection{The extrapolation check of Table~\ref{tab:rrcheck}}
\label{app:rrcheck}

The instance is mean estimation, $g(\theta,Y_t)=\theta-Y_t$ with $Y_t=y_{x_t}$ and $y=(0,1,6)$, on
the three-state chain with transition rows $(0.8,0.15,0.05)$, $(0.2,0.6,0.2)$, and
$(0.1,0.2,0.7)$, so that $\kappa=0.68$ and the residual skewness is $-0.94$. The independent control draws
the same values i.i.d.\ from the stationary law. Every cumulant is exact. The runs
use $n=4\times10^{6}$, $64$ replications, and $\eta_t=0.5\,t^{-2/3}$. All gains share one trajectory. In every row, the runs at $\rho$ and $2\rho$ correlate above $0.998$, and both two-run standard errors are within a tenth of that of the run at $\rho$.

\subsection{Constant step size and interval coverage}
\label{app:tier4}

Both checks use the featured pair of each model. Each run starts at its rule's LSTD target, each rule's
deployed run and second run share one trajectory of $8\times10^{6}$ steps, and the two rules run on independent trajectories. With the decaying
step, the runs reproduce the contrasts of Section~\ref{sec:numerics} within Monte Carlo error.

\emph{Constant step.} The step is set to the decaying step's mid-run value and to ten times that, under the current update, with $128$ replications. In none of the twenty configurations other than the controls at $\beta_2=0.99999$ does the two-run combination order the pair wrongly, including
every case in which the deployed run does. In the controls, the deviation from the LSTD contrast is within two standard errors at the smaller
step. At the larger step without momentum, it is $+0.051$ ($3.6$ standard errors) for the storage
contrast of $+0.082$ and $-0.032$ ($5.1$ standard errors) for the taxi contrast of $-0.675$, and no control
orders its pair wrongly.

\emph{Coverage.} The eight cases are the storage pair at $\beta_2=0.999$ and the taxi pair at $\beta_2=0.95$, under both updates, with and without momentum. With $500$ replications per case, coverage near $95\%$ has a Monte Carlo standard error of about $1\%$, and the band $[92.1\%,97.9\%]$ spans about three such errors on either side of $95\%$.
At $8\times10^{6}$ steps, the self-normalized interval \eqref{eq:rrinterval} covers the
combination's own across-replication mean in $85.2$--$93.0\%$ of replications, below the band in seven of the
eight cases, and the deployed run's own interval does about the same. The two-block interval
\eqref{eq:rrbm} covers that mean in $93.4$--$96.0\%$ and the LSTD contrast in $92.6$--$95.8\%$, at $3.9$ to $4.5$ times the self-normalized width (median ratios). Only the two-block interval's coverage of that mean is inside the band in all eight cases at $8\times10^{6}$, $3.2\times10^{7}$, and $1.28\times10^{8}$ steps. Batch means with four and eight blocks each leave the band in at least one case at every length.

\emph{Long runs and the third run.} For the taxi case under the current update without momentum,
the two-run interval's LSTD coverage is $81.8\%$ at $3.2\times10^{7}$ steps and $69.8\%$ at $1.28\times10^{8}$. The three-run interval covers the LSTD contrast in $90.0\%$ and $86.4\%$, at essentially
the two-run width, with a remaining displacement of about $-0.044$, whose magnitude is $0.16$
and $0.28$ of the median half-width.

\subsection{Taxi repositioning at zone resolution}
\label{app:zoneperiod}

The taxi model is refitted on $264$ zone-and-period states: the $87$ zones with at least $300$
weekday daytime pickups and one pooled unit, in three daytime periods, with pickup shares as $\nu_0$. With state indicators as
features, LSTD is exact and the memory channel vanishes. At $\beta_2=0.999$ on the twenty best of $91$ ergodic threshold rules ($3\times10^{7}$ steps, $\eta_t=20(t+10^{4})^{-2/3}$, $20$ replications per case, started at $\thetast$ and $\bar v$), the lagged update without momentum, on the best rule, gives a displacement of $+0.002$ (standard error $0.006$), and the current update displaces the best rule's value by $-0.715$ (standard error $0.005$), $1.45$ percent. With momentum, the current update gives
$-0.707$ and the lagged one $-0.639$. The selected rule never changes, and at most two close
pairs are ordered wrongly beyond two standard errors. Between $88$ and $116$ states have
stationary probability below $10^{-3}$, and at first order, single coordinates of $\thetast_\rho$ and $\thetast$ differ by more than ten percent on fourteen rules under the current update or with momentum, while the contrast, which places at most $3.3$ percent of its weight on those states, shifts by at most $2.5$ percent. With $\epsilon$ lowered to $1.7\times10^{-5}$,
the first-order shift of the current update without momentum is $-78$ to $-84$ percent of the four best
LSTD values. Adam's library damping shifts the best rule by about
$21$ percent (Section~\ref{app:highdim}).

\subsection{Adam as libraries implement it, at higher dimension}
\label{app:highdim}

The zone-resolution taxi model of Section~\ref{app:zoneperiod} was fitted by Adam with bias
correction and $\epsilon=10^{-8}$ after the square root, under the current update, with two
feature sets: one indicator per state ($d=264$, LSTD exact; $20$ rules, $10^{7}$ steps,
$\eta_t=4(t+10^{4})^{-2/3}$) and a constant plus $128$ random Fourier features of zone
covariates and period ($d=129$, LSTD not exact; $12$ rules, $3\times10^{6}$ steps,
$\eta_t=(t+10^{4})^{-2/3}$). Each case has $12$ replications, all configurations of a replication
share one trajectory, and runs start at $\thetast$. For the best rule with indicators, single-step-size TD at $40(t+10^{4})^{-2/3}$ is not displaced
($-0.0005$, standard error $0.0067$), and our damping without momentum at $20(t+10^{4})^{-2/3}$ gives $-0.709$, near the $-0.715$ of
Section~\ref{app:zoneperiod}.

With indicator features, the best rule's displacement at $(\beta_1,\beta_2)=(0.9,0.999)$ is
$-10.29$ (standard error $0.008$), $-20.8$ percent ($-23.3$ to $-17.6$ percent across rules).
Without momentum, it is $-21.1$ percent, and at $\beta_2=0.99999$, it is $-0.47$ percent. With momentum, at half and a
quarter of the effective gain, it is $-15.1$ and $-10.0$ percent, so the two-run combination leaves
$-9.3$ percent and the three-run combination $-3.5$, at $1.36$ and $1.25$ times a single run's standard
error. The deployed runs order $6$ of $190$ pairs wrongly beyond two standard errors, among the
rules ranked ninth to seventeenth. The runs at half and a quarter of the gain order $4$ and $1$,
and neither combination orders any by this test. Without momentum, the best rule's values in each band of
stationary probability are displaced by $-23.9$ to $-18.5$ percent on average, and across states, the displacement in value units correlates $0.73$
with the median minus the mean of the one-step TD target. A reading consistent with this, not
proved here, is that with indicator features, many visit rates are below $\rho$, and the nearly undamped
divisor makes updates partly sign-like, pulling each value toward the median of its TD target.
A run of $3\times10^{7}$ steps with momentum enlarges the best rule's displacement by $0.9$ percent.

With the dense features, the best rule's displacement is $-0.193$ (standard error $0.026$),
$-0.39$ percent, and the two-run combination leaves $+0.002$ ($0.027$). Halving the effective gain divides
the displacements of the four best rules by $1.9$ to $2.2$, and no configuration orders a pair wrongly beyond two standard errors.

An exploratory network with one hidden layer of $32$ tanh units was fitted for two rules by semi-gradient TD,
which does not differentiate the TD target. With library Adam, its reported values average $0.7$ to $0.8$ percent
below empirical reference values from single-step-size TD, and their two-run combination is $0.013$ and $0.032$
below the reference values. With no LSTD target, runs start at a fit to the exact values.

\subsection{The storage model with exact features}
\label{app:features}

The storage model was refitted with eight feature sets, keeping the four-feature model's reward scale,
so that the rules, exact values, and units are unchanged.
With price bins merged two by two or polynomials in price
and charge, more pairs are dropped than with the four features, and only two sets nearly remove the error in the
reported values: one indicator per state ($d=24$, LSTD exact to $10^{-12}$, no pair dropped) and an aggregation
that resolves the two lowest price bins ($d=12$, largest LSTD error $0.45$, $112$ pairs dropped),
chosen among candidate aggregations. The four-feature model's step $\eta_t=40t^{-2/3}$
was unstable with indicator features, so the runs use $\eta_t=20(t+10^{4})^{-2/3}$, the schedule of
Section~\ref{app:zoneperiod}. For both sets, all $84$ rules ran under the current update without momentum for
$8\times10^{6}$ steps in $24$ replications, each sharing one price path across rules and gains. In the control
at $\beta_2=0.99999$, no rule is more than three standard errors from its LSTD value and no pair is
ordered wrongly beyond two standard errors.

With indicator features, the counts in the main text are $285$ and $478$ wrongly ordered pairs at
$0.999$ and $0.998$, $204$ and $324$ without the nine rules the run-length check flags (all randomizing with
probability $0.1$, whose random actions largely coincide under common random numbers). Two runs leave $45$ and $152$, and three runs $3$ and $16$
(twelve features give $319$ and $518$, two runs $40$ and $163$, and three runs $5$ and $19$). A
single run at $\beta_2'=0.9995$ orders $141$ pairs wrongly. First order predicts $2{,}421$ and $2{,}971$
and a change of the selected rule, which the recursion does not confirm. The ratio of simulated to
first-order displacement has median $0.22$ at $0.999$ and $0.14$ at $0.998$. Across rules, the two
displacements have correlations $0.88$ and $0.89$. Among the $51$ rules displaced by more than five
standard errors at half the default gain, halving the gain divides the displacement by a median factor
of $1.39$, where linear scaling would give $2$. States with
stationary probability down to $6\times10^{-6}$ can place these gains outside the first-order regime
(Section~\ref{sec:tdsetup}). One near-tied pair, LSTD contrast $-0.015$, is ordered wrongly at
$+0.278$ (standard error $0.011$) at $0.999$ ($+0.318$ with $\beta_1=0.9$), $+0.160$ after two runs, and $+0.071$ after
three. On seven rules, the lagged update without momentum displaces
no rule beyond $1.1$ standard errors, consistent with the vanishing memory channel under exact features.
Rescaling the rewards to the new features' residual scale, as the four-feature model does, gives
the same deployed targets as $\epsilon=0.121$ in model units. Under the current update at $0.999$, this $\epsilon$ enlarges six of the seven displacements by $22$ to $110$
percent, and the pair is still ordered wrongly.

\subsection{Further designs and checks}
\label{app:expt}

\emph{Models.} In the storage model, the state is the price bin and the state of charge, the
reward is price times flow at an efficiency of $0.9$ each way, and $\ell$ selects the lowest
price bin with an empty battery. In the taxi model, the state is the region where the cab
becomes idle. Waiting in region $s$ earns $\bar r_s$, the next trip's expected fare net of time costs, and
repositioning from $s$ to $k$ incurs the empty trip's cost $c_{sk}$. A threshold rule repositions to the target $k$ whose gain
$\bar r_k-c_{sk}-\bar r_s$ exceeds its threshold by the largest amount and otherwise waits.

\emph{Garnet.} Eight random instances have $20$ states, $3$ actions, $3$ successors per state--action pair, $4$ features, $\gamma=0.7$, $\epsilon=0.25$, and sixty statewise Dirichlet$(0.6)$ policies, giving $14{,}160$ pairs compared by LSTD value at the uniform policy's modal state. Runs
use the current update without momentum for $4\times10^{6}$ steps with
$\eta_t=t^{-2/3}$ and $64$ replications. One policy at $\beta_2=0.95$ and three at $0.8$ fail the
run-length check and are set aside. Among pairs first order does not reverse, the recursion orders none wrongly at $0.95$ and $11$ at $0.8$.

\emph{Fitted series.} Runs start at $\thetast$ with $\eta_t=\eta_0t^{-2/3}$: $\eta_0=100$ and
$6.4\times10^{7}$ steps for prices, $\eta_0=20$ and $1.6\times10^{7}$ steps for the taxi fits, with $64$ to
$1{,}477$ replications. The cell failing the run-length check, prices under the lagged update
without momentum at $0.999$, gives $-0.033$ at quarter length and $-0.017$ at full length. Of the other eleven cells at $0.999$, eight are within two Monte Carlo standard errors of the first-order prediction. The rest give $0.92$ times the prediction for prices under the current update without momentum and about twice it under the lagged update with momentum for prices and without momentum for fares.

\emph{Storage pairs.} The rules in Table~\ref{tab:caisorev}'s misordered pairs at $\beta_2=0.999$,
$0.998$, and $0.996$ ran at $\beta_1=0$ for $8\times10^{6}$ steps with $\eta_t=40t^{-2/3}$, with $256$ replications at the default and $128$ otherwise, on common price paths. Among pairs passing the run-length check, the recursion confirms $7$ of $10$, $12$ of $14$, and $19$ of $25$ wrong orderings at the three gains under the current update, and $11$ of $13$ and $21$ of $26$ at $0.998$ and $0.996$ under the lagged one. Across all pairs, two runs leave $0$, $1$, and $4$ ordered wrongly under the current update and $0$, $0$, and $6$ under the lagged one. At $0.998$ and $0.996$, three runs leave none, with one pair undecided.

\emph{Longer runs and confinement.} Runs four times as long move Table~\ref{tab:caisorun}'s single-run and two-run contrasts by at most $0.22$ standard errors of the difference. In Table~\ref{tab:taxirev}'s runs, $10$ to $27$ of the $3{,}456$ replications per case and gain ($27$ rules, $128$ each), all under the lagged update without momentum, are flagged as possible confinement failures (Section~\ref{sec:revtaxi}). Removing these replications changes no wrong-ordering count.


\begin{thebibliography}{99}
\small
\setlength{\itemsep}{0pt}
\setlength{\parskip}{0pt}

\bibitem[Agarwal et al.(2020)]{agarwal2020}
Agarwal, R., Schuurmans, D., and Norouzi, M. (2020). An optimistic perspective on offline
reinforcement learning. \emph{ICML}.

\bibitem[Aldous and Eagleson(1978)]{aldous1978}
Aldous, D. J. and Eagleson, G. K. (1978).
On mixing and stability of limit theorems.
\emph{Annals of Probability}, 6(2):325--331.

\bibitem[An and Huo(2026a)]{anhuo2025a}
An, S. and Huo, X. (2026a).
When does dynamic preconditioning preserve the Polyak--Ruppert CLT?
A stabilization threshold. arXiv:2604.23498.

\bibitem[An and Huo(2026b)]{anhuo2026b}
An, S. and Huo, X. (2026b).
A Polyak--Ruppert central limit theorem for SA-Adam with momentum and non-convergent
adaptive preconditioning. arXiv:2606.17364.

\bibitem[Andrieu et al.(2005)]{andrieu2005}
Andrieu, C., Moulines, \'E., and Priouret, P. (2005).
Stability of stochastic approximation under verifiable conditions.
\emph{SIAM Journal on Control and Optimization}, 44(1):283--312.

\bibitem[Andrychowicz et al.(2021)]{andrychowicz2021}
Andrychowicz, M., Raichuk, A., Sta\'nczyk, P., Orsini, M., Girgin, S., Marinier, R.,
Hussenot, L., Geist, M., Pietquin, O., Michalski, M., Gelly, S., and Bachem, O. (2021).
What matters for on-policy deep actor-critic methods? A large-scale study. \emph{ICLR}.

\bibitem[Arrow and McManus(1958)]{arrow1958}
Arrow, K. J. and McManus, M. (1958). A note on dynamic stability.
\emph{Econometrica}, 26(3):448--454.

\bibitem[Benveniste et al.(1990)]{benveniste1990}
Benveniste, A., M\'etivier, M., and Priouret, P. (1990).
\emph{Adaptive Algorithms and Stochastic Approximations}. Springer.

\bibitem[Bhandari et al.(2021)]{bhandari2018}
Bhandari, J., Russo, D., and Singal, R. (2021).
A finite time analysis of temporal difference learning with linear function
approximation. \emph{Operations Research}, 69(3):950--973; earlier version
\emph{COLT} 2018.

\bibitem[Bhatia(1997)]{bhatia1997}
Bhatia, R. (1997). \emph{Matrix Analysis}. Graduate Texts in Mathematics 169,
Springer.

\bibitem[Billingsley(1999)]{billingsley1999}
Billingsley, P. (1999). \emph{Convergence of Probability Measures}, 2nd ed. Wiley.

\bibitem[Borkar and Meyn(2000)]{borkarmeyn2000}
Borkar, V. S. and Meyn, S. P. (2000). The O.D.E.\ method for convergence of stochastic
approximation and reinforcement learning. \emph{SIAM Journal on Control and Optimization},
38(2):447--469.

\bibitem[Borkar et al.(2025)]{borkar2021}
Borkar, V. S., Chen, S., Devraj, A., Kontoyiannis, I., and Meyn, S. (2025).
The ODE method for asymptotic statistics in stochastic approximation and reinforcement
learning. \emph{Annals of Applied Probability}, 35(2):936--982.

\bibitem[Boyan(2002)]{boyan2002}
Boyan, J. A. (2002). Technical update: least-squares temporal difference learning.
\emph{Machine Learning}, 49(2--3):233--246.

\bibitem[Bradtke and Barto(1996)]{bradtke1996}
Bradtke, S. J. and Barto, A. G. (1996). Linear least-squares algorithms for temporal difference
learning. \emph{Machine Learning}, 22(1--3):33--57.

\bibitem[Brandt(1986)]{brandt1986}
Brandt, A. (1986).
The stochastic equation $Y_{n+1}=A_nY_n+B_n$ with stationary coefficients.
\emph{Advances in Applied Probability}, 18(1):211--220.

\bibitem[Braverman et al.(2019)]{braverman2019}
Braverman, A., Dai, J. G., Liu, X., and Ying, L. (2019).
Empty-car routing in ridesharing systems.
\emph{Operations Research}, 67(5):1437--1452.

\bibitem[Car\`e et al.(2025)]{care2019}
Car\`e, A., Cs\'aji, B. C., Gerencs\'er, B., Gerencs\'er, L., and R\'asonyi, M. (2025).
Stochastic approximation in a Markovian framework revisited: Lipschitz continuity of the
Poisson equation. arXiv:1906.09464v3.

\bibitem[Castro et al.(2018)]{castro2018}
Castro, P. S., Moitra, S., Gelada, C., Kumar, S., and Bellemare, M. G. (2018).
Dopamine: a research framework for deep reinforcement learning. arXiv:1812.06110.

\bibitem[Chang(2004)]{chang2004}
Chang, H. S. (2004). Technical note: On ordinal comparison of policies in Markov reward
processes. \emph{Journal of Optimization Theory and Applications}, 122(1):207--217.

\bibitem[Chen et al.(2020)]{chen2020}
Chen, X., Lee, J. D., Tong, X. T., and Zhang, Y. (2020).
Statistical inference for model parameters in stochastic gradient descent.
\emph{Annals of Statistics}, 48(1):251--273.

\bibitem[Chen et al.(2022)]{chenzw2022}
Chen, Z., Zhang, S., Doan, T. T., Clarke, J.-P., and Maguluri, S. T. (2022).
Finite-sample analysis of nonlinear stochastic approximation with applications in reinforcement
learning. \emph{Automatica}, 146:110623.

\bibitem[Chung(1954)]{chung1954}
Chung, K. L. (1954).
On a stochastic approximation method.
\emph{Annals of Mathematical Statistics}, 25(3):463--483.

\bibitem[Dai(1996)]{dai1996}
Dai, L. (1996). Convergence properties of ordinal comparison in the simulation of discrete event
dynamic systems. \emph{Journal of Optimization Theory and Applications}, 91(2):363--388.

\bibitem[Dereich and Jentzen(2024)]{dereich2024}
Dereich, S. and Jentzen, A. (2024). Convergence rates for the Adam optimizer. arXiv:2407.21078.

\bibitem[Dereich and Jentzen(2026)]{dereich2026}
Dereich, S. and Jentzen, A. (2026). Central limit theorem for the averaged Adam optimizer.
arXiv:2606.21433.

\bibitem[Dereich et al.(2025a)]{dereich2025sym}
Dereich, S., Do, T., Jentzen, A., and von Wurstemberger, P. (2025a).
Adam symmetry theorem: characterization of the convergence of the stochastic Adam
optimizer. arXiv:2511.06675.

\bibitem[Dereich et al.(2025b)]{dereich2025ode}
Dereich, S., Jentzen, A., and Kassing, S. (2025b).
ODE approximation for the Adam algorithm: general and overparametrized setting.
arXiv:2511.04622.

\bibitem[Dieuleveut et al.(2020)]{dieuleveut2020}
Dieuleveut, A., Durmus, A., and Bach, F. (2020). Bridging the gap between constant step size
stochastic gradient descent and Markov chains. \emph{Annals of Statistics}, 48(3):1348--1382.

\bibitem[D\"oring et al.(2022)]{doring2022}
D\"oring, H., Jansen, S., and Schubert, K. (2022). The method of cumulants for the normal
approximation. \emph{Probability Surveys}, 19:185--270.

\bibitem[Duchi et al.(2011)]{duchi2011}
Duchi, J., Hazan, E., and Singer, Y. (2011).
Adaptive subgradient methods for online learning and stochastic optimization.
\emph{Journal of Machine Learning Research}, 12:2121--2159.

\bibitem[Fort(2015)]{fort2015}
Fort, G. (2015). Central limit theorems for stochastic approximation with controlled Markov chain
dynamics. \emph{ESAIM: Probability and Statistics}, 19:60--80. arXiv:1309.3116.

\bibitem[George and Powell(2006)]{georgepowell2006}
George, A. P. and Powell, W. B. (2006). Adaptive stepsizes for recursive estimation with
applications in approximate dynamic programming. \emph{Machine Learning}, 65(1):167--198.

\bibitem[Glasserman and Yao(1992)]{glasserman1992}
Glasserman, P. and Yao, D. D. (1992). Some guidelines and guarantees for common random
numbers. \emph{Management Science}, 38(6):884--908.

\bibitem[Glynn and Iglehart(1990)]{glynniglehart1990}
Glynn, P. W. and Iglehart, D. L. (1990).
Simulation output analysis using standardized time series.
\emph{Mathematics of Operations Research}, 15(1):1--16.

\bibitem[Glynn and Meyn(1996)]{glynnmeyn1996}
Glynn, P. W. and Meyn, S. P. (1996).
A Liapounov bound for solutions of the Poisson equation.
\emph{Annals of Probability}, 24(2):916--931.

\bibitem[Gordin(1969)]{gordin1969}
Gordin, M. I. (1969).
On the central limit theorem for stationary processes.
\emph{Doklady Akademii Nauk SSSR}, 188:739--741;
English translation, \emph{Soviet Mathematics Doklady}, 10:1174--1176.

\bibitem[Hairer and Mattingly(2011)]{hairer2011}
Hairer, M. and Mattingly, J. C. (2011).
Yet another look at Harris' ergodic theorem for Markov chains.
In \emph{Seminar on Stochastic Analysis, Random Fields and Applications VI},
Progress in Probability 63, 109--117. Birkh\"auser.

\bibitem[Henderson et al.(2018a)]{henderson2018}
Henderson, P., Islam, R., Bachman, P., Pineau, J., Precup, D., and Meger, D. (2018a).
Deep reinforcement learning that matters. \emph{AAAI}.

\bibitem[Henderson et al.(2018b)]{hendersonoptim2018}
Henderson, P., Romoff, J., and Pineau, J. (2018b). Where did my optimum go? An empirical analysis
of gradient descent optimization in policy gradient methods. \emph{European Workshop on
Reinforcement Learning}. arXiv:1810.02525.

\bibitem[Hessel et al.(2018)]{hessel2018}
Hessel, M., Modayil, J., van Hasselt, H., Schaul, T., Ostrovski, G., Dabney, W., Horgan, D.,
Piot, B., Azar, M., and Silver, D. (2018). Rainbow: combining improvements in deep
reinforcement learning. \emph{AAAI}.

\bibitem[Horn and Johnson(1991)]{hornjohnson1991}
Horn, R. A. and Johnson, C. R. (1991). \emph{Topics in Matrix Analysis}.
Cambridge University Press.

\bibitem[Huo et al.(2026)]{huo2026}
Huo, D., Chen, Y., and Xie, Q. (2026).
Bias and extrapolation in Markovian linear stochastic approximation with constant step sizes.
\emph{Mathematics of Operations Research}, articles in advance, doi:10.1287/moor.2024.0471.
Earlier version: \emph{ACM SIGMETRICS Performance Evaluation Review}, 51(1):81--82 (2023);
arXiv:2210.00953.

\bibitem[Jiang and Powell(2015)]{jiangpowell2015}
Jiang, D. R. and Powell, W. B. (2015).
Optimal hour-ahead bidding in the real-time electricity market with battery storage
using approximate dynamic programming.
\emph{INFORMS Journal on Computing}, 27(3):525--543.

\bibitem[Khalil(2002)]{khalil2002}
Khalil, H. K. (2002). \emph{Nonlinear Systems}, 3rd ed. Prentice Hall.

\bibitem[Kiefer et al.(2000)]{kiefer2000}
Kiefer, N. M., Vogelsang, T. J., and Bunzel, H. (2000).
Simple robust testing of regression hypotheses.
\emph{Econometrica}, 68(3):695--714.

\bibitem[Kiefer and Vogelsang(2005)]{kiefervogelsang2005}
Kiefer, N. M. and Vogelsang, T. J. (2005).
A new asymptotic theory for heteroskedasticity-autocorrelation robust tests.
\emph{Econometric Theory}, 21(6):1130--1164.

\bibitem[Kingma and Ba(2015)]{kingma2015}
Kingma, D. P. and Ba, J. (2015). Adam: a method for stochastic optimization.
\emph{ICLR}.

\bibitem[Kolter(2011)]{kolter2011}
Kolter, J. Z. (2011). The fixed points of off-policy TD. \emph{NeurIPS}.

\bibitem[Kushner and Yin(2003)]{kushner2003}
Kushner, H. J. and Yin, G. G. (2003).
\emph{Stochastic Approximation and Recursive Algorithms and Applications}, 2nd
edition. Springer.

\bibitem[Lauand and Meyn(2022)]{lauand2022}
Lauand, C. K. and Meyn, S. (2022).
Bias in stochastic approximation cannot be eliminated with averaging.
\emph{58th Annual Allerton Conference on Communication, Control, and Computing}.

\bibitem[Lauand and Meyn(2023)]{lauand2023}
Lauand, C. K. and Meyn, S. (2023).
The curse of memory in stochastic approximation.
\emph{62nd IEEE Conference on Decision and Control (CDC)}, pages 7803--7809.
Extended version: arXiv:2309.02944v1.

\bibitem[Lee et al.(2022)]{lee2022}
Lee, S., Liao, Y., Seo, M. H., and Shin, Y. (2022).
Fast and robust online inference with stochastic gradient descent via random
scaling. \emph{AAAI}, 36(7):7381--7389.

\bibitem[L\"ohndorf et al.(2013)]{lohndorf2013}
L\"ohndorf, N., Wozabal, D., and Minner, S. (2013).
Optimizing trading decisions for hydro storage systems using approximate dual dynamic
programming. \emph{Operations Research}, 61(4):810--823.

\bibitem[Luckett et al.(2020)]{luckett2020}
Luckett, D. J., Laber, E. B., Kahkoska, A. R., Maahs, D. M., Mayer-Davis, E.,
and Kosorok, M. R. (2020).
Estimating dynamic treatment regimes in mobile health using V-learning.
\emph{Journal of the American Statistical Association}, 115(530):692--706.

\bibitem[Lyle et al.(2023)]{lyle2023}
Lyle, C., Zheng, Z., Nikishin, E., Avila Pires, B., Pascanu, R., and Dabney, W. (2023).
Understanding plasticity in neural networks. \emph{ICML}.

\bibitem[M\'etivier and Priouret(1984)]{metivier1984}
M\'etivier, M. and Priouret, P. (1984).
Applications of a Kushner and Clark lemma to general classes of stochastic
algorithms. \emph{IEEE Transactions on Information Theory}, 30(2):140--151.

\bibitem[Meyn and Tweedie(2009)]{meyntweedie2009}
Meyn, S. P. and Tweedie, R. L. (2009).
\emph{Markov Chains and Stochastic Stability}, 2nd edition.
Cambridge University Press.

\bibitem[Mnih et al.(2015)]{mnih2015}
Mnih, V., Kavukcuoglu, K., Silver, D., Rusu, A. A., Veness, J., Bellemare, M. G., Graves, A.,
Riedmiller, M., Fidjeland, A. K., Ostrovski, G., Petersen, S., Beattie, C., Sadik, A.,
Antonoglou, I., King, H., Kumaran, D., Wierstra, D., Legg, S., and Hassabis, D. (2015).
Human-level control through deep reinforcement learning. \emph{Nature}, 518(7540):529--533.

\bibitem[Obando-Ceron and Castro(2021)]{obandoceron2021}
Obando-Ceron, J. S. and Castro, P. S. (2021). Revisiting Rainbow: promoting more insightful and
inclusive deep reinforcement learning research. \emph{ICML}.

\bibitem[Patterson et al.(2024)]{patterson2024}
Patterson, A., Neumann, S., White, M., and White, A. (2024). Empirical design in
reinforcement learning. \emph{Journal of Machine Learning Research}, 25:1--63.

\bibitem[Polyak and Juditsky(1992)]{polyak1992}
Polyak, B. T. and Juditsky, A. B. (1992).
Acceleration of stochastic approximation by averaging.
\emph{SIAM Journal on Control and Optimization}, 30(4):838--855.

\bibitem[Ramprasad et al.(2023)]{ramprasad2023}
Ramprasad, P., Li, Y., Yang, Z., Wang, Z., Sun, W. W., and Cheng, G. (2023).
Online bootstrap inference for policy evaluation in reinforcement learning.
\emph{Journal of the American Statistical Association}, 118(544):2901--2914.

\bibitem[Reddi et al.(2018)]{reddi2018}
Reddi, S. J., Kale, S., and Kumar, S. (2018).
On the convergence of Adam and beyond.
\emph{ICLR}.

\bibitem[Robbins and Siegmund(1971)]{robbins1971}
Robbins, H. and Siegmund, D. (1971).
A convergence theorem for non negative almost supermartingales and some applications.
In J. S. Rustagi (ed.), \emph{Optimizing Methods in Statistics}, pages 233--257. Academic Press,
New York.

\bibitem[Romoff et al.(2021)]{romoff2021}
Romoff, J., Henderson, P., Kanaa, D., Bengio, E., Touati, A., Bacon, P.-L., and Pineau, J.
(2021). TDprop: does adaptive optimization with Jacobi preconditioning help temporal difference
learning? \emph{AAMAS}.

\bibitem[Ruppert(1988)]{ruppert1988}
Ruppert, D. (1988). Efficient estimations from a slowly convergent
Robbins--Monro process. Technical Report 781, Cornell University.

\bibitem[Saito and Nomura(2024)]{saito2024}
Saito, Y. and Nomura, M. (2024).
Hyperparameter optimization can even be harmful in off-policy learning and how to deal with it.
\emph{IJCAI}.

\bibitem[Schmidt et al.(2021)]{schmidt2021}
Schmidt, R. M., Schneider, F., and Hennig, P. (2021). Descending through a crowded valley:
benchmarking deep learning optimizers. \emph{ICML}.

\bibitem[Secomandi(2010)]{secomandi2010}
Secomandi, N. (2010).
Optimal commodity trading with a capacitated storage asset.
\emph{Management Science}, 56(3):449--467.

\bibitem[Shao(2010)]{shao2010}
Shao, X. (2010).
A self-normalized approach to confidence interval construction in time series.
\emph{Journal of the Royal Statistical Society: Series B}, 72(3):343--366.

\bibitem[Shao(2015)]{shao2015}
Shao, X. (2015).
Self-normalization for time series: a review of recent developments.
\emph{Journal of the American Statistical Association}, 110(512):1797--1817.

\bibitem[Shazeer and Stern(2018)]{shazeer2018}
Shazeer, N. and Stern, M. (2018). Adafactor: adaptive learning rates with sublinear
memory cost. \emph{ICML}.

\bibitem[Sioshansi et al.(2009)]{sioshansi2009}
Sioshansi, R., Denholm, P., Jenkin, T., and Weiss, J. (2009).
Estimating the value of electricity storage in PJM: arbitrage and some welfare effects.
\emph{Energy Economics}, 31(2):269--277.

\bibitem[Srikant and Ying(2019)]{srikant2019}
Srikant, R. and Ying, L. (2019).
Finite-time error bounds for linear stochastic approximation and TD learning.
\emph{COLT}.

\bibitem[Sun et al.(2022)]{sun2020adatd}
Sun, T., Shen, H., Chen, T., and Li, D. (2022). Adaptive temporal difference learning with
linear function approximation. \emph{IEEE Transactions on Pattern Analysis and Machine
Intelligence}, 44(12):8812--8824.

\bibitem[Sutton(1988)]{sutton1988}
Sutton, R. S. (1988). Learning to predict by the methods of temporal
differences. \emph{Machine Learning}, 3(1):9--44.

\bibitem[Talay and Tubaro(1990)]{talay1990}
Talay, D. and Tubaro, L. (1990). Expansion of the global error for numerical schemes solving
stochastic differential equations. \emph{Stochastic Analysis and Applications}, 8(4):483--509.

\bibitem[Tan and Frazier(2022)]{tan2022}
Tan, S. and Frazier, P. I. (2022). Regret bounds and experimental design for estimate-then-optimize.
arXiv:2210.15576.

\bibitem[Tang et al.(2019)]{tang2019}
Tang, X., Qin, Z., Zhang, F., Wang, Z., Xu, Z., Ma, Y., Zhu, H., and Ye, J. (2019). A deep
value-network based approach for multi-driver order dispatching. \emph{KDD}.

\bibitem[Taniguchi et al.(2024)]{taniguchi2024}
Taniguchi, S., Harada, K., Minegishi, G., Oshima, Y., Jeong, S. C., Nagahara, G.,
Iiyama, T., Suzuki, M., Iwasawa, Y., and Matsuo, Y. (2024).
ADOPT: Modified Adam can converge with any $\beta_2$ with the optimal rate.
\emph{NeurIPS}, 37.

\bibitem[TensorFlow Developers(2026)]{tfrmsprop}
TensorFlow Developers (2026). \texttt{tf.raw\_ops.ApplyRMSProp}. TensorFlow API documentation,
\url{https://www.tensorflow.org/api_docs/python/tf/raw_ops/ApplyRMSProp}, accessed 14 September 2026.

\bibitem[Tieleman and Hinton(2012)]{tieleman2012}
Tieleman, T. and Hinton, G. (2012). Lecture 6.5---RMSProp.
COURSERA: Neural Networks for Machine Learning.

\bibitem[Tsitsiklis and Van Roy(1997)]{tsitsiklis1997}
Tsitsiklis, J. N. and Van Roy, B. (1997).
An analysis of temporal-difference learning with function approximation.
\emph{IEEE Transactions on Automatic Control}, 42(5):674--690.

\bibitem[Xiong et al.(2021)]{xiong2021}
Xiong, H., Xu, T., Liang, Y., and Zhang, W. (2021). Non-asymptotic convergence of Adam-type
reinforcement learning algorithms under Markovian sampling. \emph{AAAI}, 35(12):10460--10468.

\bibitem[Zeng et al.(2026)]{zeng2026}
Zeng, M., Zhang, Y., and Shao, X. (2026). Self-normalized inference for constant-stepsize
temporal-difference learning under Markovian sampling. arXiv:2608.10896.

\bibitem[Zhang et al.(2018)]{zhang2018}
Zhang, Y., Laber, E. B., Davidian, M., and Tsiatis, A. A. (2018).
Interpretable dynamic treatment regimes.
\emph{Journal of the American Statistical Association}, 113(524):1541--1549.

\bibitem[Zhou et al.(2016)]{zhou2016}
Zhou, Y., Scheller-Wolf, A., Secomandi, N., and Smith, S. (2016).
Electricity trading and negative prices: storage vs.\ disposal.
\emph{Management Science}, 62(3):880--898.

\bibitem[Zhou et al.(2019)]{zhou2019}
Zhou, Z., Zhang, Q., Lu, G., Wang, H., Zhang, W., and Yu, Y. (2019).
AdaShift: Decorrelation and convergence of adaptive learning rate methods.
\emph{ICLR}.

\bibitem[Zhuang et al.(2021)]{zhuang2021}
Zhuang, J., Ding, Y., Tang, T., Dvornek, N., Tatikonda, S., and Duncan, J. (2021).
Momentum centering and asynchronous update for adaptive gradient methods.
\emph{NeurIPS}, 34.

\end{thebibliography}
\end{document}